%% file: Operations-Research-template.tex
\documentclass[mnsc,nonblindrev]{informs3} 
 
\OneAndAHalfSpacedXI 

\usepackage{endnotes}
\let\footnote=\endnote
\let\enotesize=\normalsize
\def\notesname{Endnotes}%
\def\makeenmark{$^{\theenmark}$}
\def\enoteformat{\rightskip0pt\leftskip0pt\parindent=1.75em
  \leavevmode\llap{\theenmark.\enskip}}

\usepackage{amsmath,amsfonts,amssymb,graphicx,bm}
\usepackage{algorithm}
\usepackage{hyperref}
\usepackage{xcolor}
\usepackage{subfig}

\newcommand{\cost}{c}
\newcommand{\vale}{v}
\newcommand{\omege}{{\bm{w}}} 
\newcommand{\ran}{{\bm{z}}}
\newcommand{\ctext}{{\bm{x}}}
\newcommand{\thete}{{\bm{\theta}}}

\newcommand{\omegeun}{{w}} 
\newcommand{\ranun}{{z}}

\newcommand{\theteun}{{\theta}}

\newcommand{\boldv}{{\bm{u}}}

\newcommand{\E}{{\mathbb{E}}}
\usepackage{algpseudocode}

\usepackage{booktabs}

\usepackage[natbib,style=authoryear]{biblatex}
\makeatletter
\newenvironment{subtheorem}[1]{%
  \def\subtheoremcounter{#1}%
  \refstepcounter{#1}%
  \protected@edef\theparentnumber{\csname the#1\endcsname}%
  \setcounter{parentnumber}{\value{#1}}%
  \setcounter{#1}{0}%
  \expandafter\def\csname the#1\endcsname{\theparentnumber.\Alph{#1}}%
  \ignorespaces
}{%
  \setcounter{\subtheoremcounter}{\value{parentnumber}}%
  \ignorespacesafterend
}
\makeatother
\newcounter{parentnumber}

\usepackage[title]{appendix}
\TheoremsNumberedThrough     
\ECRepeatTheorems

\EquationsNumberedThrough    

\begin{document}


\RUNAUTHOR{Elmachtoub, Lam, Zhang, and Zhao}

\RUNTITLE{ETO vs IEO vs SAA: A Stochastic Dominance Perspective}

\TITLE{Estimate-Then-Optimize versus Integrated-Estimation-Optimization versus Sample Average Approximation: A Stochastic Dominance Perspective}

\ARTICLEAUTHORS{%
\AUTHOR{Adam N. Elmachtoub}
\AFF{Department of Industrial Engineering and Operations Research \& Data Science Institute, Columbia University, New York, NY 10027, \EMAIL{adam@ieor.columbia.edu}} 
\AUTHOR{Henry Lam}
\AFF{Department of Industrial Engineering and Operations Research, Columbia University, New York, NY 10027, \EMAIL{henry.lam@columbia.edu}}
\AUTHOR{Haofeng Zhang}
\AFF{Department of Industrial Engineering and Operations Research, Columbia University, New York, NY 10027, \EMAIL{hz2553@columbia.edu}}
\AUTHOR{Yunfan Zhao}
\AFF{Department of Industrial Engineering and Operations Research, Columbia University, New York, NY 10027, \EMAIL{yz3685@columbia.edu}}
} 

\ABSTRACT{%
In data-driven stochastic optimization, model parameters of the underlying distribution need to be estimated from data in addition to the optimization task. Recent literature considers integrating the estimation and optimization processes by selecting model parameters that lead to the best empirical objective performance. This integrated approach, which we call integrated-estimation-optimization (IEO), can be readily shown to outperform simple estimate-then-optimize (ETO) when the model is misspecified. In this paper, we show that a reverse behavior appears when the model class is well-specified and there is sufficient data. Specifically, for a general class of nonlinear stochastic optimization problems, we show that simple ETO outperforms IEO asymptotically when the model class covers the ground truth, in the strong sense of stochastic dominance of the regret. Namely, the entire distribution of the regret, not only its mean or other moments, is always better for ETO compared to IEO. Our results also apply to constrained, contextual optimization problems where the decision depends on observed features. Whenever applicable, we also demonstrate how standard sample average approximation (SAA) performs the worst when the model class is well-specified in terms of regret, and best when it is misspecified. Finally, we provide experimental results to support our theoretical comparisons and illustrate when our insights hold in finite-sample regimes and under various degrees of misspecification. 
}%



\maketitle

%



\section{Introduction}

We consider data-driven stochastic optimization problems, where a decision maker aims to optimize an objective function in the form of an expectation that involves noisy or random outcomes. Moreover, the underlying distribution governing the randomness is unavailable and can only be observed from data. This problem arises in many real-life problems such as inventory management \citep{qi2022practical,ban2019big}, ship inspection \citep{yan2020semi}, revenue management \citep{chen2022statistical}, portfolio optimization \citep{butler2023integrating}, healthcare \citep{chung2022decision}, and ranking \citep{kotary2022end}. 
A distinguishing challenge in this problem, compared to classical stochastic optimization, lies in the efficient incorporation of the given data. In stochastic programming or machine learning, a natural approach is to use empirical optimization or sample average approximation (SAA), namely by replacing an unknown expectation with its empirical counterpart \citep{shapiro2021lectures}. While conceptually straightforward, such an approach cannot easily apply to more complex situations, such as constrained, contextual optimization where the decision is made conditional on features and guaranteeing feasibility is necessary. In these situations, or when parametric information is utilizable, an alternative model-based approach can be used to encode the underlying distribution via a parametric model.

Two approaches have been widely considered in model-based optimization. The classic approach is \textit{estimate-then-optimize (ETO)}, which first estimates the model parameters from observed data using standard statistical tools such as maximum likelihood estimation (MLE), and then optimizes the objective function calibrated by these estimated parameters. Second is \textit{integrated-estimation-optimization (IEO)} that lumps the estimation and optimization processes together by solving a ``meta-optimization" to select the model parameter values that give rise to the best empirical objective performance,
and then using these parameter values to drive the decision. Recent literature \citep{elmachtoub2022smart,wilder2019melding,donti2017task, elmachtoub2020decision,mandi2020smart,grigas2021integrated} suggests that IEO often results in better decisions than ETO when there is model misspecification, i.e., the model class does not contain the ground truth. This phenomenon is intuitive as the parameter selection process in IEO accounts for the downstream optimization, while in ETO the estimation and optimization are separated and hence could not achieve the combined meta-optimization objective. This outperformance of IEO has been a main driver of its growing literature. On the other hand, IEO is typically much harder to solve computationally than ETO due to its integrated objective, so many previous works propose approximation methods or heuristics to solve IEO \citep{kallus2022stochastic,grigas2021integrated,sadana2023survey,mandi2024decision}. This also raises the questions on what situations or problem configurations truly necessitate the use of IEO to offer significant gain over the cheaper ETO approach.

The main goal of this paper is to theoretically characterize and compare the performances of the three approaches, ETO, IEO and SAA, for general nonlinear stochastic optimization problems. Our main findings are that \emph{when the model class is well-specified and there is sufficient data,  ETO performs better than IEO and IEO performs better than SAA (whenever applicable), which is completely reversed from the misspecified setting.} Moreover, our comparisons are in a strong sense of \emph{stochastic dominance} \citep{shaked2007stochastic,mas1995microeconomic}. Our results thus support the utility of the conceptually simpler ETO in certain settings, in contrast to the typical belief in the literature. More concretely, we consider the regret, or equivalently optimality gap or excess risk, which refers to the ground-truth objective performance of the data-driven solution relative to the optimal solution. 
This criterion provides a natural measure of solution quality since a smaller regret directly implies a better generalization performance \citep{lam2021impossibility,grigas2021integrated,hu2022fast,estes2021slow}. Our main results entail that, when the model class covers the ground-truth distribution, the large-sample regret of ETO is stochastically dominated by that of IEO, which in turn is dominated by SAA. Moreover, the ordering is a complete reversal in the misspecified setting. The stochastic dominance that we harness here is a strong notion because it implies not only the mean or any moments, but the \emph{entire probability distribution} of the regret of ETO is better than that of IEO and in turn SAA.

Our insights described above apply to general nonlinear stochastic optimization problems under standard smoothness conditions. Under these conditions, the regrets of all three approaches vanish at a rate of $O\left( \frac{1}{n} \right)$, where $n$ is the number of samples, when the model is well-specified. This necessitates us to investigate the more detailed stochastic behaviors of the regrets, represented by the limiting random variables to which these regrets converge when scaled by $n$.
Our analysis reveals how the performance ordering among ETO, IEO and SAA happens -- with suitable first and second-order optimality conditions, the regrets of all considered methods are roughly equivalent to quadratic functions of the estimated parameters (which involve derivative information of the objective function and the distribution model). These structurally resemble the mean squared errors of these parameters and, in this regard, MLE provides the asymptotically best estimator according to the celebrated Cramer-Rao bound \citep{cramer1946mathematical,rao1945information} and hints at the superiority of ETO. Despite such an intuition, eliciting this phenomenon and the full comparisons among SAA, ETO and IEO require elaborate matrix manipulations and comparisons arising from the variances of the limiting regrets which are represented as squared sums of correlated Gaussian variables.

Our findings hold for two important generalizations. First is the presence of constraints. This calls for substantial additional technicalities arising both from the incorporation of orthogonal projection to reduce the asymptotic covariance matrix (in the sense of matrix inequality), and the need to  handle Lagrangian functions and more complicated optimality conditions \citep{duchi2021asymptotic}. For instance, in these settings, the second-order optimality conditions do not guarantee that the Hessian matrix is positive definite  which, along with the orthogonal projection, results in the presence of the Moore-Penrose pseudoinverse in the asymptotic covariance, and subsequently hinders the derivation of the stochastic dominance relation. These ultimately require careful calculations and connections among an array of matrix derivatives. Second, our findings apply to contextual stochastic optimization, both with and without constraints. While previous work on contextual optimization  considers discrete distributions  \citep{grigas2021integrated}
or linear objectives \citep{elmachtoub2022smart,hu2022fast}, here we consider general nonlinear objectives and general distributions as well as provide theoretical performance comparison among the different methodologies using stochastic dominance.




Finally, we conduct numerical experiments for a variety of newsvendor and portfolio optimization problems that can be formulated as unconstrained, constrained, and contextual stochastic optimization problems with various degrees of misspecification.
Our experimental findings corroborate our theory in verifying the performance ordering and stochastic dominance among the regrets under large samples, while observing similar trends under smaller sample regimes.

\section{Related Work} \label{sec:relatedwork}
There is a vast literature on the general topic of data-driven optimization. We roughly divide them into two areas: non-contextual optimization (Section \ref{sec:non-contextual}) and contextual optimization (Section \ref{sec:contextual}), where the latter is further divided into two subareas: contextual linear optimization (Section \ref{sec:contextual linear}) and contextual nonlinear optimization (Section \ref{sec:contextual nonlinear}). Furthermore, we position our work relative to some recent works that also theoretically compare different data-driven optimization approaches statistically (Section \ref{sec:comparison}).

\subsection{Non-Contextual Optimization}\label{sec:non-contextual}
In addition to SAA introduced earlier, another popular framework to handle non-contextual data-driven optimization is distributionally robust optimization (DRO) \citep{delage2010distributionally,goh2010distributionally,ben2013robust,wiesemann2014distributionally,mohajerin2018data,bertsimas2018robust} which finds solutions that optimize the worst-case scenario, where the worst case is defined over an ambiguity set or uncertainty set. Modifications to SAA such as regularization \citep{hastie2009elements} are also shown to be (approximately) equivalent to DRO \citep{lam2016robust,lam2018sensitivity,namkoong2017variance,blanchet2019robust,gao2022wasserstein,gotoh2018robust,gupta2019near}. These lines of literature focus on nonparametric instead of model-based settings that we consider in this paper. 

In the parametric settings considered in this paper, our ETO uses the commonly used MLE \citep{bickel2015mathematical,van2000asymptotic} in the estimation step. Our IEO, on the other hand, is related to operational data analytics (ODA) or operational statistics \citep{feng2023framework,lim2006model,liyanage2005practical} where the data-driven decision, called an operational statistic, is selected within a subspace of statistics that possesses some desired property inherent in the decision-making problem. A main assertion of this literature is that one can improve traditional estimators like MLE in finite sample. However, this relies on problem-specific structures. In contrast, our IEO derived from an oracle problem (see \eqref{equ:oracle} below) is not problem-specific and, moreover, we assert the optimality of ETO under well-specified model family in the large-sample regime. The latter is consistent with the ODA literature in that the solution from ODA typically reduces to the traditional solution as the sample size goes to infinity (e.g., in \citet{liyanage2005practical,feng2023framework}).


There are also works considering other optimization setups like data-pooling \citep{gupta2022data} and with small data  \citep{besbes2021big,gupta2021small,gupta2022debiasing}, which differ from the large-sample asymptotic regime that we focus on in this work.


\subsection{Contextual Optimization}\label{sec:contextual}
In contextual data-driven stochastic optimization, the random object depends on some feature or context information (see the survey papers \citet{qi2022integrating,sadana2023survey}). In linear contextual optimization, i.e., when the cost function is a bilinear function of the decision and the random object, the expected cost can also be written bilinearly in the decision and the (mean) regression function \citep{elmachtoub2022smart}. In this case, it suffices to handle the regression function instead of the entire conditional distributions, which is otherwise required for the nonlinear setting. In the following, we discuss both the linear and nonlinear contextual settings.

\subsubsection{Contextual Linear Optimization.} \label{sec:contextual linear}
In contextual linear optimization, the objective function is linear, and the problem could possess integer or convex constraints.
\citet{elmachtoub2022smart} proposes an integrated approach to estimate the regression function (expected cost vector) by minimizing a certain decision error, called the SPO loss, instead of the prediction error. Since the SPO loss
is nonconvex and discontinuous, \citet{elmachtoub2022smart} provides a convex
surrogate loss function that is consistent under some assumptions. \citet{elmachtoub2020decision} presents strategies for training decision trees using the SPO loss function directly.  \citet{donti2017task,wilder2019melding} provide methods to differentiate the loss function, which allows the training of models to approximately minimize the decision error. In terms of performance guarantees,
\citet{el2019generalization,liu2021risk,ho2022risk} establish generalization and risk bounds in the SPO loss framework.
\citet{hu2022fast} shows that the ETO approach can have much faster convergence rates than the integrated approach when the model family is well-specified. \citet{tang2022pyepo} provides a python package for integrated approaches for contextual linear optimization. In addition, we note that there are other studies on integrated approaches for combinatorial discrete optimization \citep{wilder2019melding,mandi2020smart,wilder2019end,poganvcic2020differentiation}.
However, all the above papers focus on contextual linear or discrete optimization, while our work focuses on contextual nonlinear continuous optimization.




\subsubsection{Contextual Nonlinear Optimization.}\label{sec:contextual nonlinear}
When the objective function is nonlinear, such as in the newsvendor problem, additional efforts are required to handle the estimation beyond the regression function. We first pinpoint that works focusing on computational or empirical performances without theoretical guarantees, e.g., \citet{donti2017task,wilder2019end,munoz2022bilevel}, are different from the statistical focus in this work.
In the following, we discuss works on SAA, ETO, and IEO for contextual nonlinear optimization with statistical guarantees.

\textbf{SAA.} The naive application of SAA in constrained, contextual optimization is infeasible because the decision is now regarded as a feature map that could be high or infinite-dimensional (see Section \ref{sec:cont}). In general, one needs to restrict the SAA minimization problem to a hypothesis class of the feature-to-decision maps. For instance, \citet{ban2019big} proposes to use a class of linear functions for the newsvendor problem. \citet{bertsimas2022data} proposes to use a reproducing kernel Hilbert space. These approaches bear performances that depend on the user's choice of the feature-to-decision hypothesis class. They are conceptually natural and procedurally attractive in tackling contextual optimization. However, when there are constraints in addition to the contextualization, enforcing the constraints via the feature-to-decision map only can become very challenging. On the other hand, approaches that model the underlying distribution, such as IEO and ETO, are more appealing because they structurally maintain both the objective cost and constraints in the downstream optimization.
Another related work is \citet{esteban2022distributionally} who incorporates DRO into contextual optimization via optimal transport from the empirical distribution to the target conditional distribution. All of these works, however, do not involve statistical comparisons among different approaches as we do.

\textbf{ETO.} The first step of ETO is to obtain an accurate estimate of the conditional response distribution given the feature. \citet{bertsimas2020predictive,ban2019big,bertsimas2019predictions,srivastava2021data} propose to use nonparametric regression methods, such as $k$-nearest-neighbor, kernel regression, or local linear methods, to estimate the conditional distribution, which are different from our parametric estimation of conditional distributions.
By assuming a specially structured relation between the random response and the feature, \citet{kannan2022data,kannan2020residuals,kannan2021heteroscedasticity} studies an ETO-type approach where the first step is to estimate the regression function and the conditional covariance function and the second step is an SAA-type or DRO-type optimization step. Their approach achieves asymptotic optimality and finite sample guarantees if the estimation step achieves so. However, they consider a restricted setting where the conditional distribution estimation is not required, and thus differ from our work using parametric estimation of conditional distributions.

\textbf{IEO.} To incorporate downstream optimization in the estimation step,
\citet{kallus2022stochastic} trains a forest to reweight the empirical distribution that is conscious of the optimization task, and show that asymptotically, their forest policy achieves the optimal risk. 
For problems where the randomness in the objective has a finite discrete probability distribution, \citet{grigas2021integrated} proposes to solve an empirical risk minimization problem with respect to the in-sample cost regularized by an oracle from the conditional probability vector. These works are different from our work using parametric estimation of possibly infinite-supported conditional distributions and involving statistical comparisons using stochastic dominance on the regrets.



\subsection{Theoretical Statistical Comparisons}\label{sec:comparison}
Few theoretical studies have conducted statistical comparisons between SAA, ETO, and IEO. Most theoretical work in contextual nonlinear optimization focuses on the performance guarantees of specific proposed methods, making direct comparisons across papers infeasible due to differences in problem settings and assumptions. Additionally, in the opposite direction, \citet{estes2021slow} examines the regularity conditions necessary for establishing non-trivial convergence guarantees. Their findings indicate that no approach can achieve fast convergence with zero regret unless certain regularity conditions exist between side information and the random response. To our knowledge, the two most relevant previous studies to our work are \cite{hu2022fast,lam2021impossibility}, which we discuss in detail below. Built upon our work, a subsequent study \citep{elmachtoub2025dissecting} extends the asymptotic analysis presented in this paper to a finite-sample setting and provides finite-sample regret comparisons by developing new materials
techniques, such as Berry–Esseen-type bounds.

\citet{hu2022fast} establish some findings similar to ours in spirit. They show that ETO can be better than IEO when the model family is well-specified for contextual linear optimization. However, our work differs from \citet{hu2022fast} in the following aspects: First is that we consider nonlinear optimization while \citet{hu2022fast} focuses on linear problems. Second, in \citet{hu2022fast}, ETO and IEO exhibit different convergence rates under noise-dependent assumptions, and these ``fast" rates distinguish their performances. In contrast, in our considered nonlinear settings, the estimated parameters and decisions all exhibit a ``slow" rate, i.e., the rates of the estimated parameters and decisions are the same $O(\frac{1}{\sqrt{n}})$ for all considered approaches. Coupled with smoothness and optimality conditions, this can be shown to lead to a $O(\frac{1}{n})$ convergence rate of the attained regret. As a result, and as our third distinction with \citet{hu2022fast}, to compare ETO and IEO (and also SAA), we need to derive more precise limiting distributions at the $O(\frac{1}{n})$ scale of the regret, and moreover use the notion of first-order stochastic dominance to differentiate their performances. 

In terms of techniques, our work uses concepts similar to  \citet{lam2021impossibility} which also statistically compares optimization formulations utilizing the notion of stochastic dominance on regrets. 
However, we have significant differences. First is that \citet{lam2021impossibility} focuses on a different problem of assessing the optimality of SAA relative to locally modified algorithms with regularization or distributional robustification, without considering model misspecification issues or contextual information. Second, \citet{lam2021impossibility} argues the superiority of SAA via second-order stochastic dominance, where our conclusion here is based on a stronger first-order stochastic dominance. Third, the route to our stronger conclusion requires detailed derivations on the exact forms of the asymptotic covariance matrices for all algorithms, while \citet{lam2021impossibility} does not need these derivations as their comparisons could be concluded based on general asymptotic normality.

\section{Methodology and Preliminaries} \label{sec:uncon}

Consider a standard stochastic optimization problem in the form 
\begin{equation} \label{equ:optimalsol}
\omege^* \in \argmin_{\omege\in \Omega} \left\{\vale_0(\omege):=\mathbb{E}_{P}[\cost(\omege,\ran)]     \right\}
\end{equation}
where $\omege\in \Omega\subset \mathbb{R}^p$ is the decision and $\Omega$ is an open set in $\mathbb{R}^p$, $\ran$ is a random vector distributed according to an unknown data generating distribution $P$, $c(\cdot,\cdot)$ is a known cost function, and $\vale_0 (\cdot)$ is the expected cost under $P$.  
Our goal is to find an optimal decision $\omege^*$. 
In data-driven stochastic optimization, the ground-truth $P$ is typically unknown and, instead, we have independent and identically distributed (i.i.d.) data $\ran_1,\ldots,\ran_n$ generated from $P$. 

To infer the distribution $P$, we use a parametric approach by constructing a family of distributions $\{P_\thete: \thete\in \Theta\}$ parameterized by $\thete$. We introduce the oracle problem
\begin{equation} \label{equ:oracle}
\omege_\thete\in\argmin_{\omege\in \Omega} \{ \vale(\omege,\thete):=\mathbb{E}_\thete[\cost(\omege,\ran)] \},
\end{equation}
where $\thete\in \Theta \subset\mathbb{R}^q$ is a parameter in the underlying distribution $P_\thete$ and $\Theta$ is an open set in $\mathbb{R}^q$, $\ran$ is a random vector (variable) distributed according to $P_\thete$, and $v(\cdot,\thete)$ is the expected cost under distribution $P_\thete$. Problem \eqref{equ:oracle} outputs the solution $ \omege_\thete$ that minimizes the expected cost when the true model is $P_\thete$. In this parametric modeling framework, depending on the choice of $\{P_\thete: \thete\in \Theta\}$, $P$ may or may not be in the parametric family $\{P_\thete: \thete\in \Theta\}$. We say that the parametric family $\{P_\thete: \thete\in \Theta\}$ is \textit{well-specified} if it covers the ground-truth distribution $P$ (but the true value of $\thete$ is unknown). In contrast, we say a family $\{P_\thete: \thete\in \Theta\}$ is \textit{misspecified} if it does not cover $P$.
More precisely, we define the following:

\begin{definition}[Well-Specified Model Family]\label{def:well-specidied}
We say that the parametric family $\{P_\thete: \thete\in \Theta\}$ is \textit{well-specified} if there exists a $\thete_0\in \Theta$ such that $P=P_{\thete_0}$ among the class $\{P_\thete: \thete\in \Theta\}$.     \Halmos
\end{definition}

\begin{definition}[Misspecified Model Family]\label{def:misspecidied}
We say that the parametric family $\{P_\thete: \thete\in \Theta\}$ is \textit{misspecified} if $P\notin\{P_\thete: \thete\in \Theta\}$.
\Halmos
\end{definition}



To evaluate the performance of a decision $\omege$, we use the notion of regret, which is also known as the optimality gap or excess risk. The regret $R(\omege)$ is the expected difference in performance using decision $\omege$ compared to the optimal decision $\omege^*$ in terms of the ground-truth objective value. We provide a formal definition below. 
\begin{definition}[Regret]
For any $\omege\in \Omega$, the \textit{regret} of $\omege$ is given by
$R(\omege) := \vale_0(\omege)-\vale_0(\omege^*)$,
where $\omege^*$ is an optimal solution to \eqref{equ:optimalsol}. \Halmos
\end{definition}

The regret $R(\omege)$ is clearly non-negative and decreases as the ground-truth objective value of $\omege$ decreases, making $R(\omege)$ a natural criterion to measure solution quality. Note that any data-driven algorithm, including the approaches we  introduce below, outputs a decision $\hat\omege$ that has randomness inherited from the data. Thus, its regret $R(\hat\omege)$ is a random variable. 

While one may consider the mean of $R(\hat\omege)$ with respect to the data distribution (e.g., in \cite{hu2022fast}), we show a stronger sense of comparisons in terms of stochastic dominance of the regret. More precisely, the concept of first-order stochastic dominance \citep{quirk1962admissibility} provides a form of stochastic ordering to rank two random variables, as defined below.


\begin{definition}[Stochastic Dominance]
For any two random variables $X$, $Y$, we say that $X$ is first-order stochastically dominated by $Y$, written as $X \preceq_{st} Y$, or $Y \succeq_{st} X$, if 
\begin{equation}\label{def:stochasticdominance}
\mathbb{P}[X>x]\le \mathbb{P}[Y>x] \ \text{ for all } x\in \mathbb{R}    .
\end{equation}
\end{definition}
In addition, we say $X =_{st} Y$ if $X \preceq_{st} Y$ and $Y \preceq_{st} X$. It is easy to see that $X =_{st} Y$ if and only if $\mathbb{P}[X>x]= \mathbb{P}[Y>x] \text{ for all } x\in \mathbb{R}$, and if and only if $X \overset{d}{=} Y$, i.e., $X$ has the same distribution as $Y$. \Halmos

Importantly, 
if $X$, $Y$ are both nonnegative random variables, then $X \preceq_{st} Y$ implies that $\mathbb{E}[X^k]\le \mathbb{E}[Y^k]$ for any $k>0$. Hence the first-order stochastic dominance relation implies that any $k$th-moment, including the mean, of $X$ is no bigger than the $k$th-moment of $Y$. 
We present further properties of first-order stochastic dominance in Lemma \ref{prop1} in Appendix \ref{sec:techdetails}. 


\subsection{Data-Driven Approaches} \label{sec:algo}

We consider three approaches to obtain a data-driven solution for \eqref{equ:optimalsol}.

\noindent\paragraph{Sample Average Approximation (SAA):} This is the most straightforward approach, where the unknown expectation $\mathbb{E}_{P}$ is replaced with the empirical mean. In SAA, we solve
\begin{equation}
\inf_{\omege\in \Omega} \left\{\hat{\vale}_0(\omege):=\frac{1}{n}\sum_{i=1}^n\cost(\omege,\ran_i)\right\},\label{EO def}
\end{equation}
where $\hat{\vale}_0(\cdot)$ is the empirical mean of the cost.
In practice, an exact solution may not be obtainable due to computational reasons, but we can obtain an \textit{approximate} solution to \eqref{EO def} denoted by $\hat{\omege}^{SAA}$. The exact meaning of this approximate solution will be defined in our results.
\\

\noindent\paragraph{Estimate-Then-Optimize (ETO):} We use maximum likelihood estimation (MLE) to estimate $\thete$, i.e.,  
$$\sup_{\thete\in\Theta}\frac{1}{n}\sum_{i=1}^n \log p_\thete (\ran_i),$$
where $P_\thete$ has probability density or mass function $p_\thete$. Like SAA, in practice, an exact solution may not be obtainable, and we call an approximate solution $\hat{\thete}^{ETO}$. 
Once $\hat{\thete}^{ETO}$ is obtained, we plug into the objective and obtain 
$$\hat{\omege}^{ETO}:=\omege_{\hat{\thete}^{ETO}}=\argmin_{\omege\in \Omega} \vale(\omege,\hat{\thete}^{ETO}).$$
where $v(\cdot,\cdot)$ is the oracle objective function in \eqref{equ:oracle}. In other words, this approach uses the standard  MLE tool to estimate the unknown parameter and then plugs in the parameter estimate $\hat{\thete}^{ETO}$ in the optimization problem \eqref{equ:oracle} to obtain $\hat{\omege}^{ETO}$.
\\

\noindent\paragraph{Integrated-Estimation-Optimization (IEO):} 
We estimate $\thete$ by solving
$$\inf_{\thete\in\Theta} \hat{\vale}_0(\omege_\thete)$$
where $\hat{\vale}_0(\cdot)$ is the SAA objective function defined in \eqref{EO def} and $\omege_\thete$ is the oracle solution defined in \eqref{equ:oracle}. Once again, an exact solution may not be obtainable, and we find an approximate solution $\hat{\thete}^{IEO}$. This approach integrates optimization with the estimation process in that the loss function used to ``train" $\thete$ is the decision-making optimization problem evaluated on $\omege_\thete$. In other words, when we make decisions from a model parameterized by $\thete$, $\hat{\thete}^{IEO}$ is the choice that leads to the lowest empirical risk. 
Once $\hat{\thete}^{IEO}$ is obtained, we plug it into the objective and obtain$$\hat{\omege}^{IEO}:=\omege_{\hat{\thete}^{IEO}}=\argmin_{\omege\in \Omega} \vale(\omege,\hat{\thete}^{IEO}).$$ 






Our paper primarily focuses on the statistical comparisons among the above three approaches for nonlinear problems. Computation and algorithmic study is a separate (yet important) focus that differs from this paper. Nonetheless, our results have some relevance to this latter aspect. First, our main results allow some computation errors in obtaining $\hat{\omege}^{SAA}$, $\hat{\thete}^{ETO}$ and $\hat{\thete}^{IEO}$.
Second, we note that regarding the computation cost, SAA tends to be easier to solve than ETO and IEO, at least in the simple setting without constraints and contexts where it is applicable. In fact, SAA can be solved directly using gradient-based approaches if the cost function is convex, while the tractability of ETO and IEO depends on the adopted parametric distribution family and the structure of the objective function. For instance, we may not even have a closed form for the parametrized expected cost $v(\omege, \hat{\thete})$, in which case we may need to resort to sampling  from the fitted parametric model $P_{\hat{\thete}}$.



\subsection{Notations}
In order to describe some preliminary results on the large-sample behaviors of the regrets of different considered methods, we introduce notations that will be used throughout the paper.

For a general distribution $\tilde{P}$, we write $\mathbb{E}_{\tilde{P}}[\cdot]$ and $Var_{\tilde{P}}(\cdot)$ as the expectation and (co)variance with respect to the distribution $\tilde{P}$. We sometimes write $\mathbb{E}_{\thete}[\cdot]$ as the shorthand for $\mathbb{E}_{P_\thete}[\cdot]$ in the case of parametric distribution $P_\thete$. We let $\xrightarrow{d}$ and $\xrightarrow{P}$ denote ``convergence in distribution" and ``convergence in probability" respectively. We also use the standard stochastic order notations
$o_P(\cdot)$ and $O_P(\cdot)$ for ``convergence in probability" and ``stochastic boundedness" respectively.

For any vector $\boldv$, unless otherwise specified, $\boldv$ is viewed as a column vector, and we write $\boldv^{(j)}$ as the $j$-th element in the vector $\boldv$, $\|\boldv\|_2:=\sqrt{\sum_{j}(\boldv^{(j)})^2}$ and $\|\boldv\|_\infty:=\max_{j}|\boldv^{(j)}|$.

When $\bm{y}(\boldv): \mathbb{R}^{d_1} \to \mathbb{R}^{d_2}$ is a differentiable map, we write $\nabla \bm{y}(\boldv)$ as the $d_2\times d_1$ first-order derivative (Jacobian) matrix $(\frac{\partial y^{(i)}}{\partial u^{(j)}})_{i=1,\cdots,d_2; j=1,\cdots,d_1}$. In particular, when $y(\boldv): \mathbb{R}^{d_1} \to \mathbb{R}$ is real-valued, $\nabla y(\boldv)$ is a row vector $1\times d_1$.
When $y(\boldv_1, \boldv_2): \mathbb{R}^{d_1} \times \mathbb{R}^{d_2} \to \mathbb{R}$ is a twice differentiable real-valued function, we write $\nabla_{\boldv_k,\boldv_l} y(\boldv_1,\boldv_2)$ ($k=1,2; l=1,2$) as the $d_k\times d_l$ second-order derivative matrix $(\frac{\partial^2 y}{\partial u_{k}^{(i)} \partial u_{l}^{(j)}})_{i=1,\cdots,d_k; j=1,\cdots,d_l}$.

For any matrix $Q$, we write $\text{rank}(Q)$ as the rank of $Q$, $\text{ker}(Q)$ as the kernel (null space) of $Q$, $Q^\top$ as the transpose of $Q$, $Q^{\dagger}$ as the Moore-Penrose pseudoinverse of $Q$ \citep{stanimirovic2017minimization}. When $Q$ is invertible, we write $Q^{-1}=Q^{\dagger}$ as the (standard) inverse of $Q$ instead. For any matrix $Q$, we write $\|Q\|_{op}$ as the standard $(2,2)$-operator norm of $Q$, that is, $\|Q\|_{op}=\sup_{\mathbf u\ne 0}\frac{\|A\mathbf u\|_2}{\|\mathbf u\|_2}$ where $\mathbf u$ is a column vector.

For any symmetric matrix $Q$, we write $Q\ge0$ if $Q$ is positive semi-definite and $Q> 0$ if $Q$ is positive definite. For two symmetric matrices $Q_1$ and $Q_2$, we write $Q_1\ge Q_2$ if $Q_1-Q_2\ge 0$, in other words, $Q_1-Q_2$ is positive semi-definite. Similarly, we write $Q_1> Q_2$ if $Q_1-Q_2> 0$, in other words, $Q_1-Q_2$ is positive definite.

\subsection{Basic Statistical Results on Consistency}\label{sec:consistency}

We list out standard conditions and consistency guarantees for SAA, IEO, and ETO, which are direct consequences of asymptotic statistical theory (e.g., Theorem 5.7 in \citet{van2000asymptotic}; See Appendix \ref{sec:techdetails}).
Let $d(\cdot,\cdot)$ denote the standard Euclidean distance in the corresponding parameter or decision space. 

\begin{subtheorem}{assumption}\label{consistency all}
\begin{assumption}[Consistency conditions for SAA]\label{EOconsistency:assm}
Suppose that:
\begin{enumerate}
\item $\sup_{\omege\in \Omega} |\hat{\vale}_0(\omege)-\vale_0(\omege)|\xrightarrow{P} 0$. 
\item For every $\epsilon> 0$,
$\inf_{\omege\in \Omega: d(\omege,\omege^*)\ge \epsilon} \vale_0(\omege) > \vale_0(\omege^*)$ where $\omege^*=\argmin_{\omege\in \Omega}\vale_0(\omege)$.
\item The SAA solution $\hat{\omege}^{SAA}$ is solved approximately in the sense that
$\hat{\vale}_0(\hat{\omege}^{SAA})\le \hat{\vale}_0(\omege^*)+o_{P}(1)$. 
\end{enumerate}
\end{assumption}


\begin{assumption}[Consistency conditions for ETO]\label{ETOconsistency:assm}
Suppose that:
\begin{enumerate}
\item $\sup_{\thete\in \Theta} |\frac{1}{n}\sum_{i=1}^n \log p_\thete (\ran_i)-\mathbb{E}_{P} [ \log p_\thete(\ran)] |\xrightarrow{P} 0$.
\item For every $\epsilon> 0$, $\sup_{\thete\in \Theta: d(\thete,\thete^{KL})\ge \epsilon} \mathbb{E}_{P} [ \log p_\thete(\ran)] < \mathbb{E}_{P} [ \log p_{\thete^{KL}}(\ran)]$
where 
$\thete^{KL}:= \argmin_{\thete\in \Theta} KL(P,P_\thete)= \argmax_{\thete\in \Theta} \mathbb{E}_{P} [ \log p_{\thete}(\ran)],$
with $KL$ denoting  Kullback-Leibler divergence.
\item The estimated model parameter $\hat{\thete}^{ETO}$ in ETO is solved approximately in the sense that
$\frac{1}{n}\sum_{i=1}^n \log p_{\hat{\thete}^{ETO}} (\ran_i)\ge \frac{1}{n}\sum_{i=1}^n \log p_{\thete^{KL}} (\ran_i)-o_{P}(1)$. 
\end{enumerate}
\end{assumption}

\begin{assumption}[Consistency conditions for IEO]\label{IEOconsistency:assm}
Suppose that:
\begin{enumerate}
\item $\sup_{\thete\in \Theta} |\hat{\vale}_0(\omege_\thete)-\vale_0(\omege_\thete)|\xrightarrow{P} 0$. 
\item For every $\epsilon> 0$,
$\inf_{\thete\in \Theta: d(\thete,\thete^*)\ge \epsilon} \vale_0(\omege_\thete) > \vale_0(\omege_{\thete^*})$ where $\thete^*$ is given by
$\thete^*:= \argmin_{\thete\in \Theta} \vale_0(\omege_\thete)$. 
\item The estimated model parameter $\hat{\thete}^{IEO}$ in IEO is solved approximately in the sense that
$\hat{\vale}_0(\omege_{\hat{\thete}^{IEO}})\le \hat{\vale}_0(\omege_{\thete^{*}})+o_{P}(1)$. 
\end{enumerate}
\end{assumption}
\end{subtheorem}

In each of Assumptions \ref{EOconsistency:assm}, \ref{ETOconsistency:assm} and \ref{IEOconsistency:assm}, the first part is a uniform law of large numbers that are satisfied via Glivenko-Cantelli conditions for the corresponding function class. The second part stipulates the uniqueness of the associated ``population-level" solution of SAA, ETO or IEO, which correspond to $\omege^*$, $\thete^{KL}$ or $\thete^*$ respectively.
The third part allows the data-driven optimization procedure to incur computation error, giving rise to $\hat{\omege}^{SAA}$, $\hat{\thete}^{ETO}$, or $\hat{\thete}^{IEO}$ that can differ from the optimal solution $\omege^{*}$, $\thete^{KL}$ or $\thete^{*}$ up to $o_P(1)$ error.  

Via an application of classical M-estimation theory (Lemma \ref{prop3} or Theorem 5.7 in \citet{van2000asymptotic}), 
we have that $\hat{\omege}^{SAA}$, $\hat{\thete}^{ETO}$ and $\hat{\thete}^{IEO}$ converge in probability to $\omege^{*}$, $\thete^{KL}$ and $\thete^{*}$ respectively.

\begin{subtheorem}{proposition}\label{consistency of all}
\begin{proposition} [Consistency of SAA] \label{EOconsistency}
Suppose Assumption \ref{EOconsistency:assm} holds. Then $\hat{\omege}^{SAA}\xrightarrow{P}\omege^*$.
\end{proposition}

\begin{proposition} [Consistency of ETO] \label{ETOconsistency}
Suppose Assumption \ref{ETOconsistency:assm} holds. Then $\hat{\thete}^{ETO}\xrightarrow{P}\thete^{KL}$.
\end{proposition}

\begin{proposition} [Consistency of IEO] \label{IEOconsistency}
Suppose Assumption \ref{IEOconsistency:assm} holds. Then $\hat{\thete}^{IEO}\xrightarrow{P}\thete^{*}$.
\end{proposition}
\end{subtheorem}

We provide further details for Assumption \ref{consistency all} and the proof of Proposition \ref{consistency of all} in Appendix \ref{sec:techdetails}.


\subsection{Basic Statistical Results on Asymptotic Normality}\label{sec:normality}

We list out standard conditions and asymptotic normality guarantees for SAA, IEO, and ETO which, like Section \ref{sec:consistency}, follow directly from established results in asymptotic statistical theory (e.g., Theorem 5.23 in \citet{van2000asymptotic}; see Appendix \ref{sec:techdetails}).

\begin{subtheorem}{assumption}\label{Regularity conditions for all}
\begin{assumption}[Regularity conditions for SAA] \label{RCforEO:assm}
Suppose that $\cost(\omege,\ran)$ is a measurable function of $\ran$ such that $\omege\mapsto \cost(\omege,\ran)$ is differentiable at $\omege^*$ for almost every $\ran$ with derivative $\nabla_{\omege}\cost(\omege^*,\ran)$. Moreover, for any $\omege_1$ and $\omege_2$ in a neighborhood of $\omege^*$, there exists a measurable function $K$ with $\mathbb{E}_{P}[K(\ran)]<\infty$ such that
$|\cost(\omege_1,\ran)-\cost(\omege_2,\ran)|\le K(\ran) \|\omege_1-\omege_2\|.$ Furthermore, the map $\omege\mapsto \vale_0(\omege)$ admits a second-order Taylor expansion at the point of minimum $\omege^*$ with nonsingular symmetric second derivative matrix $\nabla_{\omege\omege}\vale_0(\omege^*)$. Lastly, 
$\hat{\omege}^{SAA}$ is solved approximately in the sense that
$$\hat{\vale}_0(\hat{\omege}^{SAA})\le \inf_{\omege\in \Omega} \hat{\vale}_0(\omege)+o_{P}(n^{-1}).$$      
\end{assumption}

\begin{assumption}[Regularity conditions for ETO] \label{RCforETO:assm}
Suppose that $\log p_\thete(\ran)$ is a measurable function of $\ran$ such that $\thete\mapsto \log p_\thete(\ran)$ is differentiable at $\thete^{KL}$ for almost every $\ran$ with derivative $\nabla_\thete \log p_{\thete^{KL}}(\ran)$. Moreover, for any $\thete_1$ and $\thete_2$ in a neighborhood of $\thete^{KL}$, there exists a measurable function $K$ with $\mathbb{E}_{P}[K(\ran)]<\infty$ such that
$|\log p_{\thete_1}(\ran)-\log p_{\thete_2}(\ran)|\le K(\ran) \|\thete_1-\thete_2\|.$
Furthermore, the map $\thete\mapsto \mathbb{E}_{P}[\log p_\thete(\ran)]$ admits a second-order Taylor expansion at the point of maximum $\thete^{KL}$ with nonsingular symmetric second derivative $\nabla_{\thete\thete}\mathbb{E}_{P}[\log p_{\thete}]|_{\thete=\thete^{KL}}$. Lastly, 
$\hat{\thete}^{ETO}$ is solved approximately in the sense that
$$\frac{1}{n}\sum_{i=1}^n \log p_{\hat{\thete}^{ETO}} (\ran_i)\ge \sup_{\thete\in \Theta}\frac{1}{n}\sum_{i=1}^n \log p_{\thete} (\ran_i)-o_{P}(n^{-1}).$$
\end{assumption}

\begin{assumption}[Regularity conditions for IEO] \label{RCforIEO:assm}
Suppose that $\cost(\omege_\thete,\ran)$ is a measurable function of $\ran$ such that $\thete\mapsto \cost(\omege_\thete,\ran)$ is differentiable at $\thete^*$ for almost every $\ran$ with derivative $\nabla_\thete \cost(\omege_{\thete^*},\ran)$. Moreover, for any $\thete_1$ and $\thete_2$ in a neighborhood of $\thete^*$, there exists a measurable function $K$ with $\mathbb{E}_{P}[K(\ran)]<\infty$ such that
$|\cost(\omege_{\thete_1},\ran)-\cost(\omege_{\thete_2},\ran)|\le K(\ran) \|\thete_1-\thete_2\|.$ Furthermore, the map $\thete\mapsto \vale_0(\omege_{\thete})$ admits a second-order Taylor expansion at the point of minimum $\thete^*$ with nonsingular symmetric second derivative $\nabla_{\thete\thete}\vale_0(\omege_{\thete^*})$. Lastly, 
$\hat{\thete}^{IEO}$ is solved approximately in the sense that
$$\hat{\vale}_0(\omege_{\hat{\thete}^{IEO}})\le \inf_{\thete\in \Theta} \hat{\vale}_0(\omege_{\thete})+o_{P}(n^{-1}).$$    
\end{assumption}
\end{subtheorem}

Although Assumption \ref{Regularity conditions for all} is standard, we provide several remarks below to clarify and provide transparency.

\begin{enumerate}
\item Regarding the notations, $\nabla_{\omege\omege}\vale_0(\omege^*) = \nabla_{\omege\omege}\vale_0(\omege) |_{\omege=\omege^*}$. 
 $\nabla_{\thete\thete}\vale_0(\omege_{\thete})$ is the second-order derivative of the map $\thete \mapsto \vale_0(\omege_{\thete})$, so $\nabla_{\thete\thete}\vale_0(\omege_{\thete^*}):=\nabla_{\thete\thete}\vale_0(\omege_\thete)|_{\thete=\thete^*}$.  
Note that in this notation, $\nabla_{\thete\thete}\vale_0(\omege_{\thete})$ is different from $\nabla_{\thete\thete}\vale(\omege_{\thete}, \thete)$ in the well-specified setting as the $\thete_0$ appearing implicitly in the definition of $\vale_0(\omege_{\thete})$ is not a variable. We  never use the latter notation $\nabla_{\thete\thete}\vale(\omege_{\thete}
, \thete)$.

\item Assumption \ref{Regularity conditions for all} includes the \emph{first-order} and \emph{second-order} optimality conditions for SAA, ETO and IEO, since an interior minimum or maximum point with nonsingular symmetric second derivative matrix must satisfy the first-order and second-order optimality conditions. These optimality conditions are standard in nonlinear optimization \citep{bazaraa2013nonlinear,nocedal1999numerical}.

\item Assumption \ref{Regularity conditions for all} does not require exact solutions but allows approximate solutions with errors up to $o_P(n^{-1})$. This assumption could be satisfied for instance by proper subsampling or using stochastic gradient descent as the computation procedure (see \citet{johnson2013accelerating,defazio2014saga,duchi2021asymptotic} and references therein). Note that, compared with Assumption \ref{consistency all}, the computation errors in Assumption \ref{Regularity conditions for all} need to be more stringently enforced to be  $o_P(n^{-1})$ instead of $o_P(1)$. 
\end{enumerate}

Via an application of the classical M-estimation theory (Lemma \ref{prop2} or Theorem 5.23 in \citet{van2000asymptotic}), we have the asymptotic normality for SAA, IEO, and ETO as follows. 

\begin{subtheorem}{proposition}\label{Asymptotic normality for all}
\begin{proposition}[Asymptotic normality for SAA] \label{RCforEO}
Suppose that Assumptions \ref{EOconsistency:assm} and \ref{RCforEO:assm} hold. Then $\sqrt{n} (\hat{\omege}^{SAA}- \omege^*)$ is asymptotically normal with mean zero and covariance matrix 
$$\nabla_{\omege\omege}\vale_0(\omege^*)^{-1}Var_{P}(\nabla_\omege \cost(\omege^*,\ran))\nabla_{\omege\omege}\vale_0(\omege^*)^{-1}$$
where $Var_{P}(\nabla_\omege \cost(\omege^*,\ran))$ is the covariance matrix of the cost gradient $\nabla_\omege \cost(\omege^*,\ran)$ under $P$.
\end{proposition} 

\begin{proposition} [Asymptotic normality for ETO] \label{RCforETO}
Suppose that Assumptions \ref{ETOconsistency:assm} and \ref{RCforETO:assm} hold. Then
$\sqrt{n} (\hat{\thete}^{ETO}- \thete^{KL})$ is asymptotically normal with mean zero and covariance matrix 
\begin{equation}
(\nabla_{\thete\thete}\mathbb{E}_{P}[\log p_{\thete}(\ran)]|_{\thete=\thete^{KL}})^{-1}
Var_{P}(\nabla_\thete \log p_{\thete^{KL}}(\ran))(\nabla_{\thete\thete}\mathbb{E}_{P}[\log p_{\thete}(\ran)]|_{\thete=\thete^{KL}})^{-1}\label{cov matrix}
\end{equation}
where $Var_{P}(\nabla_\thete \log p_{\thete^{KL}}(\ran))$ is the covariance matrix of $\nabla_\thete \log p_{\thete^{KL}}(\ran)$ under $P$.
Moreover, when $\thete^{KL}$ corresponds to the ground-truth $P$, i.e., $P_{\thete^{KL}}=P$, the covariance matrix \eqref{cov matrix} is simplified to the inverse Fisher information $\mathcal{I}_{\thete^{KL}}^{-1}$, that is,
$$\eqref{cov matrix} = \mathcal{I}_{\thete^{KL}}^{-1}= (\mathbb{E}_{P}[ (\nabla_\thete \log p_{\thete^{KL}}(\ran))^\top \nabla_\thete \log p_{\thete^{KL}}(\ran)])^{-1}.$$ 
\end{proposition}

\begin{proposition} [Asymptotic normality for IEO] \label{RCforIEO}
Suppose that Assumptions \ref{IEOconsistency:assm} and \ref{RCforIEO:assm} hold. Then
$\sqrt{n} (\hat{\thete}^{IEO}- \thete^*)$ is asymptotically normal with mean zero and covariance matrix 
$$\nabla_{\thete\thete}\vale_0(\omege_{\thete^*})^{-1} Var_{P}(\nabla_\thete \cost(\omege_{\thete^*},\ran))\nabla_{\thete\thete}\vale_0(\omege_{\thete^*})^{-1}$$
where $Var_{P}(\nabla_\thete \cost(\omege_{\thete^*},\ran))$ is the covariance matrix of the cost gradient $\nabla_\thete \cost(\omege_{\thete^*},\ran)$ under $P$.
\end{proposition}
\end{subtheorem}
We provide further details for Assumption \ref{Regularity conditions for all} and the proof of Proposition \ref{Asymptotic normality for all} in Appendix \ref{sec:techdetails}. 

\section{Main Results} \label{sec:main}
We first consider the statistical comparisons among all methods in the case of the well-specified model family in Section \ref{sec:well-specified}, which is our main focus. Then we consider the misspecified model case in Section \ref{sec:misspecified}.

\subsection{Optimization under Well-Specified Model Family}\label{sec:well-specified}

When the model family is well-specified in the sense of Definition \ref{def:well-specidied} and the consistency assumption (Assumption \ref{consistency all}) holds, it is easy to see that the optimal parameters coincide with the ground-truth value: $\thete^{KL}=\thete^*=\thete_0$ and $\omege^*=\omege_{\thete^{KL}}=\omege_{\thete^*}=\omege_{\thete_0}$, as the optimal decision $\omege^*$ can be expressed as
$$\omege^* = \argmin_{\omege\in \Omega} \left\{ \vale(\omege,\thete_0)=\mathbb{E}_{P}[\cost(\omege,\ran)] \right\}=\omege_{\thete_0}.$$
In this case, our first observation, described in Theorem \ref{correctlyspecified}, is that the regrets of all methods vanish as the sample size grows large.

\begin{theorem}[Vanishing regrets]
Suppose the model family is well-specified, i.e., there exists $\thete_0\in \Theta$ such that $P=P_{\thete_0}$. Suppose Assumption \ref{consistency all} holds. Moreover, suppose that $\vale_0(\omege)$ is continuous with respect to $\omege$ at $\omege^*$ and $\omege_\thete$ is continuous with respect to $\thete$ at $\thete_0$. Then we have $R(\hat{\omege}^{SAA}) \xrightarrow{P} 0$, $R(\hat{\omege}^{IEO})\xrightarrow{P}0$, $R(\hat{\omege}^{ETO})\xrightarrow{P} 0$.
\label{correctlyspecified}
\end{theorem}


The proof of Theorem \ref{correctlyspecified} follows from Proposition \ref{consistency of all} and the continuous mapping theorem. The detailed proof of Theorem \ref{correctlyspecified}, and all the rest of our results, are given in Appendix \ref{sec:proofs}. Theorem \ref{correctlyspecified} shows that in the well-specified case, the regrets of all three approaches have the identical limit $0$, which thus cannot be used to distinguish them. This is intuitive as ETO and IEO are able to estimate $\thete_0$ asymptotically correctly, and SAA also possesses solution consistency under the imposed assumptions. In light of Theorem \ref{correctlyspecified}, 
we now compare the regrets of these methods in terms of their higher-order convergence behaviors. In the following, we first introduce some additional assumptions and standard optimality conditions.


\begin{assumption}[Smoothness and gradient-expectation interchangeability] \label{SCforh}
Suppose that:
\begin{enumerate}
    
\item $\vale(\omege, \thete)$ is twice differentiable with respect to $(\omege, \thete)$ at $(\omege^*, \thete_0)$. 

\item The optimal solution $\omege_{\thete}$ to the oracle problem \eqref{equ:oracle} satisfies that $\omege_{\thete}$ is twice differentiable with respect to $\thete$ at 
$\thete_0$.

\item Any involved operations of integration (expectation) and differentiation can be interchanged.
Specifically, for any $\thete\in\Theta$,
$$ 
\nabla_\thete \int \nabla_{\omege} \cost(\omege^*,\ran)^\top   p_\thete(\ran)  d\ran =  \int \nabla_{\omege} \cost(\omege^*,\ran)^\top \nabla_\thete p_\thete(\ran)  d\ran, $$
$$ 
\int \nabla_{\omege} \cost(\omege,\ran) p_\thete(\ran)  d\ran|_{\omege=\omege^*} =  \nabla_\omege \int  \cost(\omege,\ran) p_\thete(\ran) d\ran|_{\omege=\omege^*}$$
\end{enumerate}
\end{assumption}

The interchangeability condition in Assumption \ref{SCforh} is a standard assumption in the Cramer-Rao bound \citep{bickel2015mathematical}. A standard route to check the interchangeability condition is to use the dominated convergence theorem. For instance, we provide a way to check the first interchange equation. If $p_{\thete}(\ran)$ is continuously differentiable with respect to $\thete$, and there exists a real-valued function $q(\ran)$ such that $\int \nabla_{\omege} \cost(\omege^*,\ran)^\top q(\ran)  d\ran< +\infty$ and
$\|\nabla_\thete p_{\thete}(\ran)\|_{\infty}\le q(\ran)$, then we have $\nabla_\thete \int \nabla_{\omege} \cost(\omege^*,\ran)^\top   p_\thete(\ran)  d\ran =  \int \nabla_{\omege} \cost(\omege^*,\ran)^\top \nabla_\thete p_\thete(\ran)  d\ran$. Other sufficient conditions (more delicate but still based on the dominated convergence theorem) can be found in \cite{l1990unified,asmussen2007stochastic,glasserman2004monte}.

Assuming Assumptions \ref{Regularity conditions for all} and \ref{SCforh} simultaneously can lead to some non-trivial facts. 
Since $\nabla_{\thete\thete} \omege_{\thete_0}$ exists by Assumption \ref{SCforh}, the chain rule implies that
$$\nabla_{\thete\thete}\vale_0(\omege_{\thete_0})= \nabla_{\thete} (\nabla_{\omege} \vale_0(\omege^*) \nabla_\thete \omege_{\thete_0})=\nabla_\thete \omege_{\thete_0}^\top \nabla_{\omege\omege} \vale_0(\omege^*) \nabla_\thete \omege_{\thete_0}$$
where we use the fact that $\nabla_{\omege} \vale_0(\omege^*)=0$. Hence, we must have
$$\text{rank}(\nabla_{\thete\thete}\vale_0(\omege_{\thete^*}))\le \text{rank}(\nabla_{\omege\omege} \vale_0(\omege^*)).$$
Assumption \ref{Regularity conditions for all} requires that both $\nabla_{\thete\thete}\vale_0(\omege_{\thete^*})^{-1}$ and $\nabla_{\omege\omege} \vale_0(\omege^*)^{-1}$ exist, which  
implicitly implies that
$$q=\text{dim}(\thete)=\text{rank}(\nabla_{\thete\thete}\vale_0(\omege_{\thete^*})), \quad p=\text{dim}(\omege)=\text{rank}(\nabla_{\omege\omege} \vale_0(\omege^*)).$$
Therefore Assumptions \ref{Regularity conditions for all} and \ref{SCforh} imply that $q \le p$. 
This is also consistent with our intuition: If the parametric model of $\thete$ is ``over-parameterized", then the optimal decision $\omege^*=\omege_{\thete_0}$ may correspond to a set of multiple $\thete$ (not only $\thete_0$) and thus making the ground-truth $\thete$ non-identifiable.





We are now ready to state our main performance comparison result in this section:


\begin{theorem}[Stochastic ordering among SAA, ETO and IEO] \label{SD}
Suppose the model family is well-specified, i.e., there exists $\thete_0\in \Theta$ such that $P=P_{\thete_0}$. Suppose Assumptions \ref{consistency all}, \ref{Regularity conditions for all}, \ref{SCforh} hold. 
Then we have $nR(\hat\omege^{\cdot})\xrightarrow{d}\mathbb G^\cdot$
for some limiting distribution $\mathbb G^\cdot=\mathbb G^{ETO}$, $\mathbb G^{SAA}$, $\mathbb G^{IEO}$ when $\hat\omege^\cdot=\hat{\omege}^{ETO}$, $\hat{\omege}^{SAA}$, $\hat{\omege}^{IEO}$ respectively. Moreover,
$\mathbb G^{ETO}\preceq_{st} \mathbb G^{IEO}\preceq_{st}\mathbb G^{SAA}.$ Additionally, if $\nabla_\thete \omege_{\thete_0}$ is invertible, then 
$\mathbb G^{IEO}=_{st}\mathbb G^{SAA}.$
\end{theorem}

Theorem \ref{SD} stipulates that, in terms of the first-order asymptotic behavior (at the rate of $\frac{1}{n}$) of the regrets, ETO is preferable to IEO, which is in turn preferable to SAA, as long as the model is well-specified. This preference is attained using the strong notion of first-order stochastic dominance, namely
$P(\mathbb G^{ETO}\leq t)\geq P(\mathbb G^{IEO}\leq t) \geq P(\mathbb G^{SAA}\leq t)$
for all $t\ge 0$. By Lemma \ref{prop1}, this means the comparison holds not only for the mean of the regret, 
but also for any increasing function $\phi:[0, \infty)\to \mathbb{R}$:
$\mathbb{E}[\phi(\mathbb G^{ETO})]\le \mathbb{E}[\phi(\mathbb G^{IEO})]\le \mathbb{E}[\phi(\mathbb G^{SAA})]$. For instance,
$$\mathbb{E}[\mathbb G^{ETO}]\le \mathbb{E}[\mathbb G^{IEO}]\le \mathbb{E}[\mathbb G^{SAA}].$$
The property of first-order stochastic dominance generally does not extend to variance, as variance cannot be expressed as an increasing function of $\mathbb G^\cdot$. However, we can still establish the following corollary. Notably, this result is not directly implied by the first-order stochastic dominance relation but is instead derived from intermediate results in our proof of Theorem \ref{SD}.
\begin{corollary}\label{cor:largervariance}
Under the same assumption as in Theorem \ref{SD}, we have that
$$Var(\mathbb G^{ETO}) \le Var(\mathbb G^{IEO}) \le Var(\mathbb G^{SAA}).$$
\end{corollary}

Finally, although Theorem \ref{SD} applies for large $n$, we observe similar trends in the finite-sample regime in our experiments (see Section \ref{sec:exp}). In the remainder of this section, we provide some intuition on how we obtain Theorem \ref{SD}, and then discussions on the strategies in verifying our needed assumptions.

\textbf{Proof outline.}
First, 
the optimality of the solution $\omege^*$ implies $\nabla_{\omege}\vale_0(\omege^*)=0 $, and the induced optimality of parameter $\thete_0$ also implies $\nabla_{\thete}\vale_0(\omege_{\thete_0})=0$, so that
\begin{equation}
R(\omege)=\vale_0(\omege)-\vale_0(\omege^*)=\frac{1}{2}(\omege-\omege^*)^\top\nabla_{\omege\omege}\vale_0(\omege^*)(\omege-\omege^*)+ 
o(\|\omege-\omege^*\|^2)
\label{excess risk:outline}
\end{equation}
\begin{equation}
R(\omege_{\thete})=\vale_0(\omege_{\thete})-\vale_0(\omege^*)=\frac{1}{2}(\thete-\thete_0)^\top\nabla_{\thete\thete}\vale_0(\omege_{\thete_0})(\thete-\thete_0)+o(\|\thete-\thete_0\|^2).\label{excess risk1:outline}
\end{equation}
In particular, \eqref{excess risk:outline} holds for $\omege=\hat{\omege}^{ETO}$, $\hat{\omege}^{SAA}$, and $\hat{\omege}^{IEO}$ (with $o$ replaced by $o_P$), and \eqref{excess risk1:outline} holds for $\thete=\hat{\thete}^{ETO}$ and $\hat{\thete}^{IEO}$ (with $o$ replaced by $o_P$). 

Then, using the asymptotic normality in Proposition \ref{Asymptotic normality for all} and the ``second-order" delta method, the limiting distributions of $nR(\hat\omege^{\cdot})$, denoted by $\mathbb G^\cdot$, all behave roughly like a quadratic form of the estimated solution or parameter: $\mathbb G^{\cdot}=\frac{1}{2}{\mathcal N^{\cdot}}^\top \mathcal{H}^{\cdot}\mathcal N^{\cdot}$, which involves the Hessian information of the objective function $\mathcal{H}^{\cdot}$ and a Gaussian variable $\mathcal N^{\cdot}$. 
Specifically, to compare ETO and IEO, we plug the asymptotic normality in Proposition \ref{Asymptotic normality for all} into \eqref{excess risk1:outline} to obtain that
$$\mathbb G^{ETO}=\frac{1}{2}{\mathcal N_1^{ETO}}^\top\nabla_{\thete\thete}\vale_0(\omege_{\thete^*})\mathcal N_1^{ETO}, \quad
\mathcal N_1^{ETO}\sim N(0,\mathcal{I}_{\thete_0}^{-1}),$$
$$\mathbb G^{IEO}=\frac{1}{2}{\mathcal N_1^{IEO}}^\top\nabla_{\thete\thete}\vale_0(\omege_{\thete^*})\mathcal N_1^{IEO}, \quad \mathcal N_1^{IEO}\sim N(0,Cov(\mathcal N_1^{IEO})).$$ 

Note that the difference between $\mathbb G^{ETO}$ and $\mathbb G^{IEO}$ only lies in the two Gaussian variables $\mathcal N_1^{ETO}$ and $\mathcal N_1^{IEO}$. We establish the following lemma to assist our development.
\begin{lemma} \label{lemma1}
Let $Q_1$, $Q_2$, and $Q_3$ be any positive semi-definite matrices. Let $\bm{Y}_1$ and $\bm{Y}_2$ be multivariate Gaussian random vectors with distributions $N(0,Q_1)$ and $N(0,Q_2)$, respectively.  
If $Q_1\le Q_2$, then $\bm{Y}_1^\top Q_3 \bm{Y}_1 \preceq_{st} \bm{Y}_2^\top Q_3 \bm{Y}_2$.
\end{lemma}
Based on Lemma \ref{lemma1}, in order to compare $\mathbb G^{ETO}$ and $\mathbb G^{IEO}$, it suffices to compare $\mathcal{I}_{\thete_0}^{-1}$ in $\mathcal N_1^{ETO}$ versus $Cov(\mathcal N_1^{IEO})$ in $\mathcal N_1^{IEO}$.
The multivariate Cramer-Rao bound \citep{bickel2015mathematical} concludes that MLE provides the asymptotically best estimator in terms of the covariance, i.e., $\mathcal{I}_{\thete_0}^{-1}$, which hints at the superiority of ETO over IEO. 

Next, to see that IEO is better than SAA, we plug the asymptotic normality in Proposition \ref{Asymptotic normality for all} into \eqref{excess risk:outline} to obtain that
$$\mathbb G^{SAA}=\frac{1}{2}{\mathcal N_2^{SAA}}^\top\nabla_{\omege\omege}\vale_0(\omege^*)\mathcal N_2^{SAA}, \quad 
\mathcal N_2^{SAA}\sim N(0,Cov(\mathcal N_2^{SAA})),$$
$$\mathbb G^{IEO}=\frac{1}{2}{\mathcal N_2^{IEO}}^\top\nabla_{\omege\omege}\vale_0(\omege^*)\mathcal N_2^{IEO}, \quad
\mathcal N_2^{IEO}\sim N(0,Cov(\mathcal N_2^{IEO})).
$$ 
Note that since the SAA solution is obtained at the $\omege$ level instead of $\thete$ level, we now turn to expansion \eqref{excess risk:outline} instead of \eqref{excess risk1:outline} to compare SAA and IEO. In this case, $\mathcal N_2^{IEO}$, the Gaussian variable in the limit at the $\omege$ level, is different from $\mathcal N_1^{IEO}$, the counterpart at the $\thete$ level. To derive $\mathcal N_2^{IEO}$ from $\mathcal N_1^{IEO}$, we use the delta method: $\sqrt{n}(\omege_{\hat{\thete}^{IEO}}-\omege^*)=\sqrt{n}(\omege_{\hat{\thete}^{IEO}}-\omege_{\thete_0})=\nabla_{\thete} \omege_{\thete_0} \sqrt{n}(\hat{\thete}^{IEO}-\thete_0)+o_p(1)$.

Again by leveraging Lemma \ref{lemma1}, it is sufficient to compare the covariance matrices in $\mathcal N_2^{SAA}$ and $\mathcal N_2^{IEO}$. However, this requires additional technical efforts that cannot be addressed by the Cramer-Rao bound. To provide an intuition for our techniques, note that since IEO leverages the useful information that the model is well-specified, the covariance matrix in $\mathcal N_2^{IEO}$ behaves like the ``restriction" of the one in $\mathcal N_2^{SAA}$ to the correct subspace of $\omege$ induced by the parameter $\thete$ (the range of $\omege_\thete$), and this covariance is thus smaller (and hence better) than that of $\mathcal N_2^{SAA}$. The rigorous technical result to reflect this phenomenon is in Lemma \ref{lemma2} below. 
\begin{lemma} \label{lemma2}
Let $Q_1\in \mathcal{R}^{p\times p}$ be any invertible matrix, $Q_2\in \mathcal{R}^{p\times p}$ be any positive semi-definite matrix, and $Q_3\in \mathcal{R}^{p\times q}$ be any matrix (not necessarily a square matrix) such that $Q_3^\top Q_1 Q_3$ is a positive definite matrix. For any $\lambda\ge 0$, we have that
$$Q_3(Q_3^\top Q_1 Q_3 + \lambda I_q)^{-1}Q_3^\top Q_2 Q_3(Q_3^\top Q_1 Q_3+ \lambda I_q)^{-1}Q_3^\top\le Q_1^{-1} Q_2 Q_1^{-1}.$$
\end{lemma}
Finally, if $\nabla_\thete \omege_{\thete_0}$ is invertible (i.e., $\omege$ and $\thete$ is one-to-one in a small open neighborhood of $\thete_0$), then the ``restriction" is like the identity operator and therefore SAA and IEO behave the same. The full proof comparing SAA, ETO, and IEO requires elaborate matrix manipulations and establishing the connections and differences among multiple matrices, which are presented in Appendix \ref{sec:proofs}.

\textbf{Verifying assumptions.}
Our key assumptions to elicit the main Theorem \ref{correctlyspecified} in this section,  namely Assumptions \ref{consistency all}, \ref{Regularity conditions for all}, \ref{SCforh}, are all rather standard in the statistics and stochastic optimization literature. However, they do require case-by-case verifications by using some level of problem structure. To showcase the applicability of these assumptions, we provide a detailed verification of them in a newsvendor problem, which will be our main example to illustrate numerics in Section \ref{sec:exp}. More concretely, we have the following:
\begin{proposition} \label{prop: Verifying Assumptions}
Consider the newsvendor problem:
$$\min_{\bm \omege} 
\E_P\left[ \bm{h}^\top(\omege-\ran)^+ + \bm{b}^\top(\ran-\omege)^+\right].$$
We assume each product $j$ has demand distribution $\mathcal{N}(t_j\theteun,\sigma_j)$, where $\theteun\in\Theta$ is the unknown parameter that we want to learn and the ground truth is $\theteun_0$. 
$\bm{h}$, $\bm{b}$, $t_j$, $\sigma_j$ are all constants. Suppose that when this problem is solved by ETO, IEO, and SAA, there exists a compact set $\hat{\Omega} \subset \mathbb{R}^{p}$ and a compact set $\hat{\Theta} \subset \mathbb{R}^{q}$ where the two compact sets are allowed to be larger than $\Omega$ or $\Theta$ respectively, such that the estimates $\hat{\thete}^{ETO}$, $\hat{\thete}^{IEO}$, $\hat{\omege}^{SAA}$ satisfy that $\hat{\thete}^{IEO}\in \hat{\Theta}$, $\hat{\thete}^{IEO} \in \hat{\Theta}$, $\hat{\omege}^{SAA}\in \hat{\Omega}$ with probability 1.
For this setting, Assumptions \ref{consistency all} (including Assumptions \ref{EOconsistency:assm}, \ref{ETOconsistency:assm}, \ref{IEOconsistency:assm}), \ref{Regularity conditions for all} (including Assumptions \ref{RCforEO:assm}, \ref{RCforETO:assm}, \ref{RCforIEO:assm}), and \ref{SCforh} hold, and thus the result in Theorem \ref{SD} holds.
\end{proposition}


The proof of Proposition \ref{prop: Verifying Assumptions} is given in Appendix \ref{sec:proofs}. Our proof strategy, which is also generally applicable to other problems, is outlined as below:
\begin{itemize} 

\item First, we justify the interchange of differentiation and expectation of the cost, where the differentiation is up to the second order and with respect to both the distribution model parameter $\thete$ and the decision $\omege$. This can be done by directly applying the dominated convergence theorem, which is what we use in our proof, or other known results in the stochastic derivative literature, such as \cite{l1990unified,asmussen2007stochastic,glasserman2004monte} (though they all still use dominated convergence in certain ways). Along with this justification, we would also obtain expressions for the Hessian of $\vale(\omege,\thete)$ with respect to $(\omege,\thete)$. These correspond to Step 1 in our proof of Proposition \ref{prop: Verifying Assumptions}.


\item The above allows us to verify Assumption \ref{SCforh} Part 1 immediately. Then, we would need to obtain an expression for the solution map $\thete \mapsto \omege_\thete$ that allows us to verify twice differentiability of this map, or use tools such as the implicit function theorem and other structural knowledge. This allows us to verify Assumption \ref{SCforh} Part 2. For the newsvendor problem considered in our proof, we can obtain expression for this solution map that allows a ready check of twice differentiability. Moreover, we need to verify Assumption \ref{SCforh} Part 3, by direct computation which is what we do in the proof, or use similar techniques as the bullet point above to interchange differentiation and expectation. These correspond to Step 2 in our proof of Proposition \ref{prop: Verifying Assumptions}, and at this point we verify the entire Assumption \ref{SCforh}.

\item The remainder is to verify Assumptions \ref{IEOconsistency:assm} and \ref{RCforIEO:assm} (for IEO), Assumptions \ref{ETOconsistency:assm} and \ref{RCforETO:assm} (for ETO), and Assumptions \ref{EOconsistency:assm} and  \ref{RCforEO:assm} (for SAA). Note that we have grouped these assumptions for the three different methods as they are indeed verified most efficiently in these groupings. For Assumptions \ref{IEOconsistency:assm} and \ref{RCforIEO:assm}, we would need to use the expressions derived in the first two bullet points above (Steps 1 and 2 in our proof of Proposition \ref{prop: Verifying Assumptions}) and trace down the needed detailed properties such as the Lipschitz constant and nonsingularity of the Hessian. These correspond to Step 3 in our proof of Proposition \ref{prop: Verifying Assumptions}. For Assumptions \ref{ETOconsistency:assm} and \ref{RCforETO:assm}, they are purely about the likelihood function of the distribution model, and do not require the downstream optimization objective, and thus the verification strategy is the same as standard MLE. In Step 4 in our proof of Proposition \ref{prop: Verifying Assumptions} we tackle this task. For Assumptions \ref{EOconsistency:assm} and  \ref{RCforEO:assm}, they are purely about SAA and reduce to the standard verification machinery for M-estimation or SAA. This is Step 5 in our proof of Proposition \ref{prop: Verifying Assumptions}.





\end{itemize}

\subsection{Optimization under Misspecified Model Family}\label{sec:misspecified}

When the model family is misspecified in the sense of Definition \ref{def:misspecidied}, 
the regrets of the contending methods no longer all converge to zero, as stated below.

\begin{theorem}[Comparisons under model misspecification]  \label{misspecified}
Suppose Assumption \ref{consistency all} holds. Moreover, suppose that $\vale_0(\omege)$ is continuous with respect to $\omege$ at $\omege^*$, and $\omege_\thete$ is continuous with respect to $\thete$ at $\thete^*$ and $\thete^{KL}$. Then we have $R(\hat{\omege}^{SAA}) \xrightarrow{P} 0$, $R(\hat{\omege}^{ETO})\xrightarrow{P} \vale_0(\omege_{\thete^{KL}})-\vale_0(\omege^*):=\kappa^{ETO}$, and $R(\hat{\omege}^{IEO})\xrightarrow{P} \vale_0(\omege_{\thete^*})-\vale_0(\omege^*):=\kappa^{IEO}$. Moreover,
$\kappa^{ETO}\geq\kappa^{IEO}\geq 0.$
\end{theorem}

Theorem \ref{misspecified} concludes that the performance ordering of the three approaches completely reverses in the case of misspecified model family, compared to Theorem \ref{SD}. For instance, although $\hat{\omege}^{ETO}$ exhibits the best regret asymptotically when the model is well-specified, it becomes the worst in the misspecified situation.
The reason is that in the misspecified setting,  the regrets for IEO and ETO may not vanish as the sample size $n$ grows, and significant differences already arise in the zeroth-order behaviors of the three approaches.

Our findings (Theorems \ref{correctlyspecified}, \ref{SD} and \ref{misspecified}) hold for two important generalizations of our model. The first is constrained stochastic optimization (Section \ref{sec:cons}), where additional known constraints appear in the optimization task. The second is contextual stochastic optimization (Section \ref{sec:cont}), where the distribution of the randomness depends on some contextual information.

\section{Generalizations to Constrained Stochastic Optimization} \label{sec:cons}

We generalize our results to constrained stochastic optimization problems.
Consider the formulation
\begin{align} 
\omege^*\in     \argmin_{\omege\in \Omega} \ & \left\{\vale_0(\omege):=\mathbb{E}_{P}[\cost(\omege,\ran)]\right\}
 \label{equ:optimalsol:cons}\\
\text{s.t. }  & g_j(\omege) \le 0 \text{ for } j \in J_1  \nonumber\\
& g_j(\omege) = 0 \text{ for } j \in J_2\nonumber
\end{align}
where we have $|J_1|$ inequality constraints and $|J_2|$ equality constraints, with known constraint functions denoted by $g_j(\omege)$.
We let $J=J_1\cup J_2$ be the index set of all constraints. Let $\tilde{\Omega}$ denote the resulting feasible region of  \eqref{equ:optimalsol:cons}, i.e., $\tilde{\Omega}:=\{\omege\in \Omega:g_j(\omege) \le 0 \text{ for } j \in J_1, \  g_j(\omege) = 0 \text{ for } j \in J_2\nonumber\}$.

To address the constraints in \eqref{equ:optimalsol:cons}, we define its Lagrangian function:
\begin{equation}
\vale_0(\omege) 
+ \sum_{j\in J}\alpha_j g_j(\omege)   
\end{equation}
where $\bm{\alpha}=(\alpha_{j})_{j\in J}$ are Lagrange multipliers.
Let $\bm{\alpha}^*=(\alpha_{j}^*)_{j\in J}$ denote the Lagrange multipliers corresponding to the solution $\omege^*$.

Like the unconstrained stochastic optimization problem \eqref{equ:oracle}, the parametrized oracle problem with constraints is
\begin{align} 
\omege_{\thete}\in    \argmin_{\omege\in \Omega} \ & \left\{\vale(\omege,\thete):=\mathbb{E}_{P_{\thete}}[\cost(\omege,\ran)]\right\} \label{equ:oracle:cons}\\
\text{s.t. }  & g_j(\omege) \le 0 \text{ for } j \in J_1 \nonumber\\
& g_j(\omege) = 0 \text{ for } j \in J_2\nonumber
\end{align}
for any $\thete\in \Theta$. To address the constraints in \eqref{equ:oracle:cons}, we consider its Lagrangian function:
\begin{equation} \label{equ:oracleLagrangian}
\vale(\omege,\thete) 
+ \sum_{j\in J}\alpha_j g_j(\omege)   
\end{equation}
for any $\thete$ where $\bm{\alpha}=(\alpha_{j})_{j\in J}$ are Lagrange multipliers. Let $\bm{\alpha}(\thete)=(\alpha_{j}(\thete))_{j\in J}$ denote the Lagrange multipliers corresponding to the solution $\omege_\thete$ under the parameter $\thete$.

Depending on the choice of $\{P_\thete: \thete\in \Theta\}$, $P$ may or may not be in the parametric family $\{P_\thete: \thete\in \Theta\}$. Therefore, like in Section \ref{sec:main}, we shall study the well-specified and misspecified cases separately.

\subsection{Data-Driven Approaches for Constrained Stochastic Optimization} \label{sec:algo:cons}
The three approaches considered in Section \ref{sec:uncon} can be similarly applied for constrained stochastic optimization.  
\\

\noindent\paragraph{Sample Average Approximation (SAA):} We do not utilize parametric information, and we solve
\begin{align}
\inf_{\omege\in \Omega} & \left\{\hat{\vale}_0(\omege):=\frac{1}{n}\sum_{i=1}^n\cost(\omege,\ran_i)\right\}  \label{equ:EO:cons}  \\
\text{s.t. }  & g_j(\omege) \le 0 \text{ for } j \in J_1  \nonumber\\ & g_j(\omege) = 0 \text{ for } j \in J_2. \nonumber
\end{align}
In practice, an exact solution may not be obtainable, and we call an approximate solution $\hat{\omege}^{SAA}$. To address the constraints in this problem, we consider its Lagrangian function 
\begin{equation} 
\frac{1}{n}\sum_{i=1}^n\cost(\omege,\ran_i) + \sum_{j\in J}\alpha_j g_j(\omege),
\end{equation}
where $\bm{\alpha}=(\alpha_{j})_{j\in J}$ are Lagrange multipliers. Let $\hat{\bm{\alpha}}^{SAA}=(\hat{\alpha}_j^{SAA})_{j\in J}$ denote the Lagrange multipliers corresponding to the solution $\hat{\omege}^{SAA}$.\\

\noindent\paragraph{Estimate-Then-Optimize (ETO):} We use MLE to estimate $\thete$, i.e.,  
$$\sup_{\thete\in\Theta}\frac{1}{n}\sum_{i=1}^n \log p_\thete (\ran_i),$$
where $P_\thete$ has probability density or mass function $p_\thete$. Again, an exact solution may not be obtainable, and we call an approximate solution $\hat{\thete}^{ETO}$. 
Once $\hat{\thete}^{ETO}$ is obtained, we plug into the objective and obtain
$$\hat{\omege}^{ETO}:=\omege_{\hat{\thete}^{ETO}}=\argmin_{\omege\in\tilde{\Omega}} \vale(\omege,\hat{\thete}^{ETO}).$$
Let $\hat{\bm{\alpha}}^{ETO}=(\alpha_j(\hat{\thete}^{ETO}))_{j\in J}$ denote the Lagrange multipliers corresponding to the solution $\hat{\omege}^{ETO}=\omege_{\hat{\thete}^{ETO}}$ under the parameter $\hat{\thete}^{ETO}$ in \eqref{equ:oracleLagrangian}. 
\\

\noindent\paragraph{Integrated-Estimation-Optimization (IEO):} We estimate $\thete$ by solving
$$\inf_{\thete\in\Theta} \hat{\vale}_0(\omege_\thete).$$
where $\hat{\vale}_0(\cdot)$ is the SAA objective function defined in \eqref{equ:EO:cons} and $\omege_\thete$ is the oracle solution defined in \eqref{equ:oracle:cons}. 
An exact solution may not be obtainable, and we call an approximate solution $\hat{\thete}^{IEO}$.
Once $\hat{\thete}^{IEO}$ is obtained, we plug it into the objective and obtain$$\hat{\omege}^{IEO}:=\omege_{\hat{\thete}^{IEO}}=\argmin_{\omege\in \tilde{\Omega}} \vale(\omege,\hat{\thete}^{IEO}).$$ 
Let $\hat{\bm{\alpha}}^{IEO}=(\alpha_j(\hat{\thete}^{IEO}))_{j\in J}$ denote the Lagrange multipliers corresponding to the solution $\hat{\omege}^{IEO}$ under the parameter $\hat{\thete}^{IEO}$ in \eqref{equ:oracleLagrangian}. \\




\subsection{Optimization under Well-Specified Model Family}\label{sec:constrained well-specified}
Suppose the parametric family $\{P_\thete: \thete\in \Theta\}$ is well-specified in the sense of Definition \ref{def:well-specidied}. In this case, the optimal decision $\omege^*$ can be expressed as
$$\omege^* = \argmin_{\omege\in\tilde{\Omega}} \left\{ \vale(\omege, \thete_0):=\mathbb{E}_{P}[\cost(\omege,\ran)] \right\}=\omege_{\thete_0}$$
and the Lagrange multipliers $\bm{\alpha}^*$ corresponding to the solution $\omege^*$ can be expressed as
$$\bm{\alpha}^*=\bm{\alpha}(\thete_0)$$
where $\bm{\alpha}(\thete)$ is given right after \eqref{equ:oracleLagrangian}.

Our first observation is the consistency of the regret in Theorem \ref{correctlyspecified:cons}, which extends Theorem \ref{correctlyspecified} to constrained stochastic optimization.

\begin{theorem}[Vanishing regrets in constrained stochastic optimization]
Suppose the model family is well-specified, i.e., there exists $\thete_0\in \Theta$ such that $P=P_{\thete_0}$. Suppose Assumption \ref{consistency all} (with $\Omega$ replaced by $\tilde{\Omega}$) holds. Moreover, suppose that $\vale_0(\omege)$ is continuous with respect to $\omege$ at $\omege^*$, and $\omege_\thete$ is continuous with respect to $\thete$ at $\thete_0$. Then we have $R(\hat{\omege}^{SAA}) \xrightarrow{P} 0$, $R(\hat{\omege}^{IEO})\xrightarrow{P}0$, $R(\hat{\omege}^{ETO})\xrightarrow{P} 0$.
\label{correctlyspecified:cons}
\end{theorem}

Therefore, like the unconstrained case, to meaningfully compare regrets, we seek to characterize the first-order convergence behaviors of the regrets. To this end, we need new techniques to handle the constraints and Lagrangian functions.

The following assumption, given by \citet{duchi2021asymptotic,shapiro1989asymptotic}, is the extension of Assumption \ref{RCforEO:assm} from the unconstrained to the constrained case.

\begin{assumption} [Regularity conditions for SAA with constraints] \label{RCforEO:cons:assm}
Suppose that 
\begin{enumerate}

\item $\cost(\omege,\ran)$ is a measurable function of $\ran$ such that $\omege\mapsto \cost(\omege,\ran)$ is convex and continuously differentiable for almost every $\ran$ with derivative $\nabla_\omege \cost(\omege,\ran)$. $g_j(\omege)$  is convex and twice differentiable with respect to $\omege$ for all $j \in J$. 

\item There exists $C_1 \in (0,\infty)$ such that for all $\omege \in \tilde{\Omega}$,
$$\|\nabla_\omege \vale_0(\omege) - \nabla_\omege \vale_0(\omege^*)\| \le C_1
\|\omege-\omege^*\|.$$
There exist $C_2$, $\varepsilon \in (0,\infty)$ such that for all $\omege \in \tilde{\Omega} \cap \{\omege : 	\| \omege-\omege^*\|\le \varepsilon\}$,
$$\|\nabla_\omege \vale_0(\omege) - \nabla_\omege \vale_0(\omege^*)-\nabla_{\omege\omege} \vale_0(\omege^*)(\omege-\omege^*)\| \le C_2
\|\omege-\omege^*\|^2.$$
There exist $C_3\in (0,\infty)$ such that for all $\omege \in \tilde{\Omega}$,
$$\mathbb{E}_{P}[\|\nabla_\omege \cost(\omege, \ran) - \nabla_\omege \cost(\omege^*, \ran)\|^2] \le C_3
\|\omege-\omege^*\|^2.$$
\item The second-order optimality conditions hold for the target problem \eqref{equ:optimalsol:cons}. More specifically, we let $B$ be the set of active constraints, i.e., $B =
\{j \in J_1 \cup J_2 : g_j(\omege^*) = 0\}$ and define the
critical tangent set at $\omege^*$ by
\begin{equation}\label{tangent}
\mathcal{T}(\omege^*) := \{
\bm{\beta} \in \Omega : \nabla_{\omege} g_j(\omege^*) \bm{\beta} = 0 \text{ for all } j \in B\}.    
\end{equation}
We assume that there exists $\mu > 0$ such that for any $\bm{\bm{\beta}} \in \mathcal{T}(\omege^*)$,
$$\bm{\beta}^\top \left(\nabla_{\omege\omege} \vale_0(\omege^*)+ \sum_{j\in J} \alpha^*_j \nabla_{\omege\omege} g_j(\omege^*)\right)
\bm{\beta} \ge \mu \|\bm{\beta}\|^2.$$

\item The Linear Independence Constraint Qualification (LICQ) holds for the target problem \eqref{equ:optimalsol:cons}. More specifically, we assume that $-\vale_0(\omege^*)$ is a relative interior point of the set
$$\{\bm{\beta} \in \Omega : 
\langle \bm{\beta},\omege' - \omege^*\rangle \le 0 \text{ for all } \omege' \in \tilde{\Omega}\}$$
and the set $\{\nabla_{\omege} g_j(\omege^*): j\in B\}$ is linearly independent.

\item 
The SAA solution $\hat{\omege}^{SAA}$ is solved approximately in the sense that
$$\hat{\vale}_0(\hat{\omege}^{SAA})\le \inf_{\omege\in \tilde{\Omega}} \hat{\vale}_0(\omege)+o_{P}(n^{-1}).$$  
\end{enumerate}
\end{assumption}

It is known that the following asymptotic normality for SAA holds, which is more delicate than Proposition \ref{RCforEO} because of the constraints.
\begin{proposition} [Asymptotic normality for SAA under constraints] \label{RCforEO:cons}
Suppose that Assumptions \ref{EOconsistency:assm} (with $\Omega$ replaced by $\tilde{\Omega}$) and \ref{RCforEO:cons:assm} hold. Let $$\nabla_{\omege\omege}\bar{\vale}_0(\omege^*)=\nabla_{\omege\omege}\vale_0(\omege^*)+\sum_{j\in J} \alpha^*_j \nabla_{\omege\omege} g_j(\omege^*)=\nabla_{\omege\omege}\vale_0(\omege^*)+\sum_{j\in B} \alpha^*_j \nabla_{\omege\omege} g_j(\omege^*),$$
$$\Phi=I-A^T(AA^T)^{-1}A$$
where $B$ is the set of active constraints, i.e., $B =
\{j \in J_1 \cup J_2 : g_j(\omege^*) = 0\}$, $A=(\nabla_{\omege} g_j(\omege^*))_{j\in B}$, i.e., $A$ is the matrix whose rows consist of $\nabla_{\omege} g_j(\omege^*)$ only for the active constraints ($|B|$ rows in total), $\Phi$ is the orthogonal projection onto the tangent set $\mathcal{T}(\omege^*)$ in \eqref{tangent}, and we write $(\Phi\nabla_{\omege\omege}\bar{\vale}_0(\omege^*)\Phi)^{\dagger}$ as the Moore-Penrose pseudoinverse of $\Phi\nabla_{\omege\omege}\bar{\vale}_0(\omege^*)\Phi$.

Then we have that
$\sqrt{n} (\hat{\omege}^{SAA}- \omege^*)$ is asymptotically normal with mean zero and covariance matrix 
$$\Phi(\Phi\nabla_{\omege\omege}\bar{\vale}_0(\omege^*)\Phi)^{\dagger}\Phi Var_{P}(\nabla_\omege \cost(\omege^*,\ran))\Phi(\Phi\nabla_{\omege\omege}\bar{\vale}_0(\omege^*)\Phi)^{\dagger}\Phi.\footnotemark$$
\footnotetext{It seems that \citet{duchi2021asymptotic} has a typo in their Proposition 1 and Corollary 1. Their matrix $(\nabla_{\omege\omege}\bar{\vale}_0(\omege^*))^{\dagger}$ should be $(\Phi\nabla_{\omege\omege}\bar{\vale}_0(\omege^*)\Phi)^{\dagger}$. This typo originates from the solution to the quadratic programming problem with linear constraints in the final step of their proof, which should be given by, e.g.,
Proposition 2.1. in \cite{stanimirovic2017minimization}. We have confirmed this with the original authors of \citet{duchi2021asymptotic}. It is also worth mentioning that by the property of Moore-Penrose pseudoinverse on the orthogonal projection matrix, we have equivalently $\Phi(\Phi\nabla_{\omege\omege}\bar{\vale}_0(\omege^*)\Phi)^{\dagger}\Phi=\Phi(\Phi\nabla_{\omege\omege}\bar{\vale}_0(\omege^*)\Phi)^{\dagger}=(\Phi\nabla_{\omege\omege}\bar{\vale}_0(\omege^*)\Phi)^{\dagger}\Phi=(\Phi\nabla_{\omege\omege}\bar{\vale}_0(\omege^*)\Phi)^{\dagger}$.}
\end{proposition}

Note that Propositions \ref{RCforETO} and \ref{RCforIEO} are still valid under Assumptions \ref{RCforETO:assm} and \ref{RCforIEO:assm}, as the constraints on $\omege$ do not explicitly enter into the optimization problems on $\thete$ in ETO and IEO that are considered under Assumptions \ref{RCforETO:assm} and \ref{RCforIEO:assm}. On the other hand, the constraints on $\omege$ indeed impact the oracle problem \eqref{equ:oracle:cons} in ETO and IEO, which is not captured by Assumption \ref{RCforEO:cons:assm}. Therefore, in addition to Assumption \ref{SCforh}, we introduce the following assumption that the Karush–Kuhn–Tucker (KKT) conditions for the oracle problem hold, which is common in constrained optimization \citep{kallus2022stochastic,duchi2021asymptotic,wright1993identifiable,bazaraa2013nonlinear,nocedal1999numerical}.




\begin{assumption} [Conditions on constraints] \label{OCforZ:cons}

Suppose that:

\begin{enumerate}

\item $\alpha_j(\thete)$ is twice differentiable with respect to $\thete$ at $\thete_0$ for all $j \in J$.

\item The KKT conditions hold for problem \eqref{equ:oracle:cons} for all $\thete\in\Theta$. More specifically, 
$\omege_\thete$ is a function of $\thete$ that satisfies
$$\nabla_\omege \vale(\omege, \thete)+ \sum_{j\in J} \alpha_j(\thete) \nabla_\omege g_j(\omege)|_{\omege=\omege_\thete} = 0, \quad \forall \thete\in \Theta$$
and complementary slackness holds:
$$\alpha_j(\thete) g_j(\omege_\thete)=0, \quad \forall j\in J, \ \forall \thete\in \Theta.$$

\item We assume $\hat{\alpha}_j^{SAA}\xrightarrow{P} \alpha_j^*$ for all $j\in J_1 \cap B \cap \{j\in J: \alpha_j^*\ne 0\}$ and complementary slackness holds for problem \eqref{equ:EO:cons}:
$$\hat{\alpha}_j^{SAA} g_j(\hat{\omege}^{SAA})=0, \quad \forall j\in J.$$




\end{enumerate}

\end{assumption}

It is worth mentioning that complementary slackness in the second part of Assumption \ref{OCforZ:cons} implies that
$$\hat{\alpha}_j^{ETO} g_j(\hat{\omege}^{ETO})=0, \quad \forall j\in J,$$
$$\hat{\alpha}_j^{IEO} g_j(\hat{\omege}^{IEO})=0, \quad \forall j\in J,$$
$$\alpha_j^* g_j(\omege^*)=0, \quad \forall j\in J,$$
by setting $\thete=\hat{\thete}^{ETO}$, $\hat{\thete}^{IEO}$, or $\thete_0$. The third part of Assumption \ref{OCforZ:cons} implies that the active constraints are ``preserved" when solving SAA, in the sense that the SAA solution is also active on the constraints with index $j\in J_1 \cap B \cap \{j\in J: \alpha_j^*\ne 0\}$ (a subset of active inequality constraints from the optimal solution) with high probability. A more rigorous statement about this implication can be found in our proof.

Now we are ready to present our main result for constrained stochastic optimization in Theorem \ref{SD2} below.

\begin{theorem}[Stochastic ordering in constrained stochastic optimization] \label{SD2}
Suppose the model family is well-specified, i.e., there exists $\thete_0\in \Theta$ such that $P=P_{\thete_0}$. Suppose Assumptions \ref{consistency all} (with $\Omega$ replaced by $\tilde{\Omega}$), \ref{Regularity conditions for all} (with Assumption \ref{RCforEO:assm} replaced by Assumption \ref{RCforEO:cons:assm}), \ref{SCforh},  \ref{OCforZ:cons} hold. Then we have $nR(\hat\omege^{\cdot})\xrightarrow{d}\mathbb G^\cdot$
for some limiting distribution $\mathbb G^\cdot=\mathbb G^{ETO}$, $\mathbb G^{SAA}$, $\mathbb G^{IEO}$ when $\hat\omege^\cdot=\hat{\omege}^{ETO}$, $\hat{\omege}^{SAA}$, $\hat{\omege}^{IEO}$ respectively. Moreover,
$\mathbb G^{ETO}\preceq_{st} \mathbb G^{IEO}\preceq_{st}\mathbb G^{SAA}$. 
\end{theorem}

Theorem \ref{SD2} shows the same conclusion as Theorem \ref{SD} but for constrained optimization. Despite the similarity of the conclusion, the proof of Theorem \ref{SD2} is substantially more complex than Theorem \ref{SD}, due to the following additional technical difficulties.
\begin{enumerate}

\item The conditions in Assumptions \ref{RCforEO:cons:assm} and \ref{OCforZ:cons}, including the KKT conditions, the second-order optimality conditions, and the linear independence constraint qualification, are all new ingredients in the constrained problem. More specifically, we rely on Lagrangian function and Lagrange multipliers in our analysis.  Moreover, the second-order optimality conditions do not guarantee that the Hessian matrix $\nabla_{\omege\omege}\bar{\vale}_0(\omege^*)$ is positive definite and subsequently hinders the derivation of stochastic dominance relations by leveraging Lemma \ref{lemma1}. To resolve these difficulties, we  establish new connections of multiple derivative matrices in our proof which allow us to leverage Lemma \ref{lemma1} again but in a modified way.

\item SAA exhibits different asymptotic normality in the constrained case (Proposition \ref{RCforEO:cons}) than the unconstrained counterpart (Proposition \ref{RCforEO}). In particular, the emergence of the orthogonal projection matrix $\Phi$ reduces the covariance matrix in the asymptotic Gaussian variable, i.e., the covariance matrix in constrained SAA is smaller than the one in unconstrained SAA. Therefore, even if we know the covariance matrix in IEO is smaller than the one in unconstrained SAA, we cannot directly claim that it is smaller than the one in constrained SAA.


\item Since the asymptotic normality in Proposition \ref{RCforEO:cons} involves the Moore-Penrose pseudoinverse of $\Phi\nabla_{\omege\omege}\bar{\vale}_0(\omege^*)\Phi$, which is not invertible (even $\nabla_{\omege\omege}\bar{\vale}_0(\omege^*)$ is not necessarily invertible), 
Lemma \ref{lemma2} needs to be generalized to handle the Moore-Penrose pseudoinverse, which is stated in Lemma \ref{lemma3} below.

\begin{lemma} \label{lemma3}
Let $Q_0\in \mathcal{R}^{p\times p}$ be any orthogonal projection matrix, $Q_1\in \mathcal{R}^{p\times p}$ be any matrix such that $Q_0 Q_1 Q_0$ is positive semi-definite with $\text{rank}(Q_0 Q_1 Q_0)=\text{rank}(Q_0)$, $Q_2\in \mathcal{R}^{p\times p}$ be any positive semi-definite matrix, and $Q_3\in \mathcal{R}^{p\times q}$ be any matrix (not necessarily a square matrix) such that $Q_3^\top Q_0 Q_1 Q_0 Q_3$ is a positive definite matrix. For any $\lambda\ge 0$,
\begin{align*}
& Q_0Q_3(Q_3^\top Q_0Q_1Q_0 Q_3 + \lambda I_q)^{-1}Q_3^\top Q_2 Q_3(Q_3^\top Q_0Q_1Q_0 Q_3 + \lambda I_q)^{-1}Q_3^\top Q_0   \\
& \le  Q_0(Q_0 Q_1 Q_0)^{\dagger} Q_2 (Q_0 Q_1 Q_0)^{\dagger}Q_0.
\end{align*}
\end{lemma}

\item The Moore-Penrose pseudoinverse does not have the ``continuity" property that the standard inverse possesses. That is, if we have a result for $(Q+\gamma I_p)^{-1}$ for any $\gamma>0$, we \textit{cannot} say that this result holds for $Q^{\dagger}$ in general. Thus Lemma \ref{lemma3} is not a simple generalization of Lemma \ref{lemma2}, as it requires additional conditions (such as the rank condition) and new technical details. This also increases the challenge in using Lemma \ref{lemma3} for our main theorem, as all the conditions in Lemma \ref{lemma3} must be verified in the proof of our main Theorem \ref{SD2}.

\end{enumerate}

Therefore, all the above challenges, including Lemma \ref{lemma3}, require substantial new proof ideas and techniques compared to the unconstrained case. The proofs are provided in Section \ref{sec:proofs}.


Finally, regarding assumption verification, we note that the additional assumptions that are required to handle the constrained case, namely Assumptions \ref{RCforEO:cons:assm} and \ref{OCforZ:cons}, are not unique to our analysis, but have also appeared in similar forms in prior works, including \citet{shapiro1989asymptotic,duchi2021asymptotic,kallus2022stochastic}. To our best knowledge, none of these existing studies have explicitly verified these assumptions, even for canonical or simple examples. This omission can be attributed to the substantial technical challenges in ensuring the underlying second-order conditions and differentiability of projected mappings, which in turn requires elaborate efforts in handling the associated Lagrangian systems. Such a rigorous study on assumption verification for constrained problems warrants a separate full-length work. Nonetheless, in our experiments in Section \ref{sec:exp}, we will show that the asymptotic solution behaviors in the considered constrained settings align with our Theorem \ref{SD2}, thus giving confidence to the applicability of our theory to the constrained case.








\subsection{Optimization under Misspecified Model Family}

Suppose now the parametric family $\{P_\thete: \thete\in \Theta\}$ is misspecified in the sense of Definition \ref{def:misspecidied}. 
Theorem \ref{misspecified:cons} below shows that the result in Theorem \ref{misspecified} also holds for constrained stochastic optimization.

\begin{theorem}[Comparisons under model misspecification with constraints]  \label{misspecified:cons}
Suppose Assumption \ref{consistency all} (with $\Omega$ replaced by $\tilde{\Omega}$) holds. Moreover, suppose that $\vale_0(\omege)$ is continuous with respect to $\omege$ at $\omege^*$, and $\omege_\thete$ is continuous with respect to $\thete$ at $\thete^*$ and $\thete^{KL}$. Then we have $R(\hat{\omege}^{SAA}) \xrightarrow{P} 0$, $R(\hat{\omege}^{ETO})\xrightarrow{P} \vale_0(\omege_{\thete^{KL}})-\vale_0(\omege^*):=\kappa^{ETO}$, and $R(\hat{\omege}^{IEO})\xrightarrow{P} \vale_0(\omege_{\thete^*})-\vale_0(\omege^*):=\kappa^{IEO}$. Moreover,
$\kappa^{ETO}\geq\kappa^{IEO}\geq 0.$
\end{theorem}

\section{Generalizations to Contextual Stochastic Optimization} \label{sec:cont}

We generalize our previous discussions to contextual stochastic optimization problems. We focus on the setting where the class of conditional distributions is parameterized by a vector $\thete$, as a natural extension of the presented non-contextual optimization problems. This is in constrast to  nonparametric approaches considered in previous work \citep{kallus2022stochastic,grigas2021integrated}.


Consider a contextual stochastic optimization problem in the form (either with constraints or without constraints)
\begin{align}
\omege^*(\ctext) \in \quad \quad  \argmin_{\omege\in \Omega} & \left\{ \vale_0(\omege|\ctext) :=\mathbb{E}_{P(\ran|\ctext)}[\cost(\omege,\ran)|\ctext]\right\} \label{contextualproblem}\\
\text{s.t. }  & g_j(\omege) \le 0 \text{ for } j \in J_1 , \nonumber\\
& g_j(\omege) = 0 \text{ for } j \in J_2\nonumber   
\end{align}
where 1) $\omege(\ctext)\in \Omega\subset \mathbb{R}^p$ is the decision and $\Omega$ is an open set in $\mathbb{R}^p$; 2) $\ctext \in \mathcal{X}$ is the associated contextual information that affects
the distribution of $\ran$ and given $\ctext$, $\ran$ is a random response distributed according to an unknown ground-truth data generating distribution $P(\ran|\ctext)$; 3) $c(\cdot,\cdot)$ is a known cost function; 4) We have $|J_1|$ inequality constraints and $|J_2|$ equality constraints, with known constraint functions denoted $g_j(\omege)$. We allow $J_1=J_2=\emptyset$ to represent the case without any constraints.

Our goal is to find the optimal decision function $\omege^*(\ctext)$ under the ground-truth conditional distribution $P(\ran|\ctext)$ in \eqref{contextualproblem}. 
In our considered data-driven settings, the ground-truth $P(\ran|\ctext)$ is unknown. Instead, we have i.i.d. data $(\ctext_1,\ran_1),\ldots,(\ctext_n,\ran_n)$ generated from the joint distribution of $(\ctext,\ran)$ denoted $P:=P(\ctext, \ran)=P(\ran|\ctext)P(\ctext)$, where
$P(\ctext)$ is the ground-truth data generating marginal distribution of $\ctext$. 
Let $J=J_1\cup J_2$. Let $\tilde{\Omega}$ denote the feasible region of this problem $\tilde{\Omega}:=\{\omege\in \Omega:g_j(\omege) \le 0 \text{ for } j \in J_1, g_j(\omege) = 0 \text{ for } j \in J_2\nonumber\}$. If $J_1=J_2=\emptyset$, then $\tilde{\Omega}=\Omega$.

To address the constraints in \eqref{equ:oracle:cont}, we define its Lagrangian function:
\begin{equation}
\vale_0(\omege|\ctext) 
+ \sum_{j\in J}\alpha_j g_j(\omege)   
\end{equation}
where $\bm{\alpha}=(\alpha_{j})_{j\in J}$ are the Lagrange multipliers. Let $\bm{\alpha}^*(\ctext)=(\alpha^*_{j}(\ctext))_{j\in J}$ denote the Lagrange multipliers corresponding to the solution $\omege^*(\ctext)$.



To infer the distribution $P(\ran|\ctext)$, we can use a parametric approach by constructing a family of distributions $\{P_\thete(\ran|\ctext): \thete\in \Theta\}$ parameterized by $\thete$ as in Section \ref{sec:uncon}. In this case, we define a class of oracle problems:
\begin{align}
\omege_\thete(\ctext) \in \quad \quad  \argmin_{\omege\in \Omega} & \left\{ \vale(\omege, \thete|\ctext):=\mathbb{E}_{P_\thete(\ran|\ctext)}[\cost(\omege,\ran)|\ctext]\right\} \label{equ:oracle:cont}\\
\text{s.t. }  & g_j(\omege) \le 0 \text{ for } j \in J_1 , \nonumber\\
& g_j(\omege) = 0 \text{ for } j \in J_2\nonumber   
\end{align}
where 1) $\thete\in \Theta \subset\mathbb{R}^q$ is a parameter in the underlying distribution $P_\thete(\ran|\ctext)$ and $\Theta$ is an open set in $\mathbb{R}^q$; 2) Given $\ctext$, $\ran$ is a random response distributed according to $P_\thete(\ran|\ctext)$.



To address the constraints in \eqref{equ:oracle:cont}, we consider its Lagrangian function:
\begin{equation} \label{equ:oracleLagrangian:cont}
\vale(\omege,\thete|\ctext) 
+ \sum_{j\in J}\alpha_j g_j(\omege)   
\end{equation}
where $\bm{\alpha}=(\alpha_{j})_{j\in J}$ are the Lagrange multipliers. Let $\bm{\alpha}(\thete,\ctext)=(\alpha_{j}(\thete,\ctext))_{j\in J}$ denote the Lagrange multipliers corresponding to the solution $\omege_\thete(\ctext)$ under the parameter $\thete$.

Depending on the choice of $\{P_\thete(\ran|\ctext): \thete\in \Theta\}$, $P(\ran|\ctext)$ may or may not be in the parametric family $\{P_\thete(\ran|\ctext): \thete\in \Theta\}$. We say that the parametric family $\{P_\thete(\ran|\ctext): \thete\in \Theta\}$ is \textit{well-specified} if it covers the ground-truth distribution $P(\ran|\ctext)$ (but the true value of $\thete$ is unknown). More precisely, we define the following:
\begin{definition}[Well-Specified Model]\label{def:well-specified:cont}
We say that the parametric family $\{P_\thete(\ran|\ctext): \thete\in \Theta\}$ is
\textit{well-specified} if there exists $\thete_0\in \Theta$ such that $P(\ran|\ctext)=P_{\thete_0}(\ran|\ctext)$ for any $\ctext$ among the class $\{P_\thete(\ran|\ctext): \thete\in \Theta\}$.\Halmos
\end{definition}

\begin{definition}[Misspecified Model Family]\label{def:misspecified:cont}
We say that the parametric family $\{P_\thete(\ran|\ctext): \thete\in \Theta\}$ is \textit{misspecified} if for any $\thete\in \Theta$,  $P(\ran|\ctext)\ne P_{\thete}(\ran|\ctext)$ for some $\ctext$.
\Halmos
\end{definition}


Throughout Section \ref{sec:cont}, we write
$$P_\thete(\ctext,\ran):=P_\thete(\ran|\ctext)P(\ctext)$$ 
as the parametric joint distribution of $(\ctext,\ran)$ under the conditional distribution $P_\thete(\ran|\ctext)$ and write
$$\vale(\omege,\thete):=\mathbb{E}_{P(\ctext)}[\vale(\omege(\ctext), \thete|\ctext)]=\mathbb{E}_{P_\thete(\ran|\ctext)P(\ctext)}[\cost(\omege(\ctext),\ran)]$$
as the mean of $\vale(\omege(\ctext), \thete|\ctext)$ with respect to $P(\ctext)$, i.e.,
$\vale(\omege,\thete)$ measures the \textit{average expected cost} from the decision function $\omege(\ctext)$ under the parametric joint distribution $P_\thete(\ctext,\ran)$ where ``average" is in the sense of averaging over the $\ctext$ space. Recall that $\vale_0(\omege(\ctext)|\ctext)=\mathbb{E}_{P(\ran|\ctext)}[\cost(\omege(\ctext),\ran)|\ctext]$ and we also write $\vale_0(\omege):=\mathbb{E}_{P(\ctext)}[\vale_0(\omege(\ctext)|\ctext)]$ as the the ground-truth average expected cost from the decision function $\omege(\ctext)$.
Remark that in the notation $\vale_0(\omege(\ctext)|\ctext)$ or $\vale(\omege(\ctext), \thete|\ctext)$, we explicitly write ``$|\ctext$" to emphasize it is derived from the expectation conditioning on $\ctext$.

As in Section \ref{sec:uncon}, we introduce the average regret as our evaluation criterion of a decision function $\omege(\ctext)$. 

\begin{definition} [Average regret]
For any $\omege(\ctext): \mathcal{X}\to \Omega$, define the average regret of $\omege(\ctext)$ as
$$R(\omege) := \int_{\mathcal{X}} \Big(\vale_0(\omege(\ctext)|\ctext)-\vale_0(\omege^*(\ctext)|\ctext)\Big) P(d\ctext)=\vale_0(\omege)-\vale_0(\omege^*)$$
where $\omege^*(\ctext)$ is a ground-truth optimal solution. \Halmos
\end{definition}


\subsection{Data-Driven Approaches for Contextual Stochastic Optimization} \label{sec:algo:cont}
For contextual stochastic optimization, a straightforward application of SAA is not a viable approach since it allows to choose \textit{any} map from context to decision, which clearly overfits the finite-sample problem. To be specific, SAA considers
$$\inf_{\omege(\ctext)\in \tilde{\Omega}} \left\{\hat{\vale}_0(\omege)=\frac{1}{n}\sum_{i=1}^n\cost(\omege(\ctext_i),\ran_i)\right\}.$$
To obtain a solution to this problem, it suffices to optimize the value of $\omege(\ctext_i)$ only at $\ctext_1,\cdots,\ctext_n$, while
$\omege(\ctext)$ for any $\ctext\ne \ctext_1,\cdots,\ctext_n$ is irrelevant to the optimization problem and hence can be defined as any values. This obtained SAA solution thus cannot generalize properly to any $\ctext$ that is not previously observed and, in this sense, it overfits the problem for any finite sample.


Hence, it is common to restrict SAA to a certain hypothesis class of feature-to-decision maps, such as the class of functions induced by IEO or any user’s choice (see Section \ref{sec:contextual nonlinear}).
To add to the complication, SAA is also difficult to implement when there are constraints in contextual optimization, as guaranteeing the feasibility of the feature-to-decision map requires an intermediate step that again leads to IEO-like approaches. 
For these reasons, we consider only IEO and ETO in this section.
These two approaches, originally considered in Section \ref{sec:uncon}, can be naturally extended to contextual stochastic optimization.  

\noindent\paragraph{Estimate-Then-Optimize (ETO):} We use the data to obtain an estimate $\hat{\thete}^{ETO}$, via MLE. More precisely, 
$$\sup_{\thete\in\Theta}\frac{1}{n}\sum_{i=1}^n \log p_\thete (\ran_i|\ctext_i).$$
where $P_\thete(\ran|\ctext)$ has the conditional density or mass function $p_\thete(\ran|\ctext)$. Note that $p_\thete(\ctext, \ran)$, the joint density or mass function of $P_\thete(\ctext, \ran)$, can be written as $p_\thete(\ctext, \ran)=p_\thete(\ran|\ctext)p(\ctext)$ where $p(\ctext)$ is the density or mass function of $P(\ctext)$, independent of $\thete$. Thus this problem is equivalent to  
$$\sup_{\thete\in\Theta}\frac{1}{n}\sum_{i=1}^n \log p_\thete (\ctext_i, \ran_i).$$
In practice, the exact solutions may not be obtainable, and we call the approximate solution $\hat{\thete}^{ETO}$.
Plug it into the objective to obtain 
$$\hat{\omege}^{ETO}(\ctext):=\omege_{\hat{\thete}^{ETO}}(\ctext)=\argmin_{\omege\in\tilde{\Omega}} \vale(\omege,\hat{\thete}^{ETO}|\ctext).$$
Let $\hat{\bm{\alpha}}^{ETO}(\ctext)=(\alpha_j(\hat{\thete}^{ETO}, \ctext))_{j\in J}$ denote the Lagrange multipliers corresponding to the solution $\hat{\omege}^{ETO}(\ctext)$ under the (random) parameter $\hat{\thete}^{ETO}$ in \eqref{equ:oracleLagrangian:cont}. 
\\

\noindent\paragraph{Integrated-Estimation-Optimization (IEO):} 
We estimate $\thete$ by solving
$$\inf_{\thete\in\Theta} \left\{\hat{\vale}_0(\omege_\thete)=\frac{1}{n}\sum_{i=1}^n\cost(\omege_\thete(\ctext_i),\ran_i)\right\}$$
where $\hat{\vale}_0(\cdot)$ is the SAA objective function and $\omege_\thete$ is the oracle solution defined in \eqref{equ:oracle:cont}. The exact solutions may not be obtainable, and we call the approximate solution $\hat{\thete}^{IEO}$.
This approach presents an optimization-aware way to estimate $\thete$, and a similar idea for discrete distributions has been studied in \cite{grigas2021integrated}. Note that by definition, for any $\omege(\ctext): \mathcal{X}\to \Omega$,
$$\mathbb{E}_{P}[\hat{\vale}_0(\omege)]=\vale_0(\omege),$$
i.e., $\hat{\vale}_0(\omege)$ is the empirical counterpart of $\vale_0(\omege)$. Therefore, $\hat{\thete}^{IEO}$ is obtained at the level of ``the joint distribution $P$". 
Once $\hat{\thete}^{IEO}$ is obtained, plug it into the objective to obtain 
$$\hat{\omege}^{IEO}(\ctext):=\omege_{\hat{\thete}^{IEO}}(\ctext)=\argmin_{\omege\in\tilde{\Omega}} \vale(\omege,\hat{\thete}^{IEO}|\ctext).$$
Let $\hat{\bm{\alpha}}^{IEO}(\ctext)=(\alpha_j(\hat{\thete}^{IEO}, \ctext))_{j\in J}$ denote the Lagrange multipliers corresponding to the solution $\hat{\omege}^{IEO}(\ctext)$ under the (random) parameter $\hat{\thete}^{IEO}$ in \eqref{equ:oracleLagrangian:cont}. \\

We pinpoint that the standard conditions to guarantee consistency and asymptotic normality for IEO and ETO that we introduced in Sections \ref{sec:consistency} and \ref{sec:normality} can be naturally extended to contextual optimization without difficulties. See Assumption \ref{consistency all:cont} and Proposition \ref{consistency of all:cont} for consistency results in Appendix \ref{sec:cont:more}. See Assumption \ref{Regularity conditions for all:cont} and Proposition \ref{Asymptotic normality for all:cont} for regularity results in Appendix \ref{sec:cont:more}.

\subsection{Optimization under Well-Specified Model}
Suppose the parametric family $\{P_\thete(\ran|\ctext): \thete\in \Theta\}$ is well-specified in the sense of Definition \ref{def:well-specified:cont}. In this case, the optimal decision $\omege^*(\ctext)$ can be expressed as
$$\omege^*(\ctext) = \argmin_{\omege\in\tilde{\Omega}}  \left\{ \vale(\omege,\thete_0|\ctext):=\mathbb{E}_{P_{\thete_0}(\ran|\ctext)}[\cost(\omege,\ran)|\ctext]\right\}=\omege_{\thete_0}(\ctext)$$
for any $\ctext$ and the Lagrange multipliers $\bm{\alpha}^*(\ctext)$ corresponding to the solution $\omege^*(\ctext)$ can be expressed as
$$\bm{\alpha}^*(\ctext)=\bm{\alpha}(\thete_0,\ctext)$$
where $\bm{\alpha}(\thete,\ctext)$ is given in \eqref{equ:oracleLagrangian:cont}.

Our first observation is the consistency of the regret, which shows that the result in Theorem \ref{correctlyspecified} also holds in the contextual stochastic optimization.

\begin{theorem} [Vanishing regrets in contextual stochastic optimization] \label{correctlyspecified:cont}
Suppose that there exists a $\thete_0\in \Theta$ such that $P(\ran|\ctext)=P_{\thete_0}(\ran|\ctext)$ for any $\ctext$.
Suppose Assumption \ref{consistency all:cont} hold. Moreover, suppose that $\vale_0(\omege_\thete)$ is continuous with respect to $\thete$ at $\thete_0$. Then we have $R(\hat{\omege}^{IEO})\xrightarrow{P}0$, $R(\hat{\omege}^{ETO})\xrightarrow{P} 0$.
\end{theorem}

As in the non-contextual stochastic optimization, to meaningfully compare regrets, we seek to characterize the first-order convergence behaviors. 
We adapt Assumption \ref{SCforh} in the non-contextual case to the following assumption:

\begin{assumption}[Smoothness and gradient-expectation interchangeability] \label{SCforh:cont}
Suppose that:
\begin{enumerate}
    
\item For any fixed $\ctext\in\mathcal{X}$, $\vale(\omege, \thete|\ctext)$ is twice differentiable with respect to $(\omege, \thete)$ at $(\omege^*(\ctext), \thete_0)$, $g_j(\omege)$ ($\forall j\in J$) is twice differentiable with respect to $\omege$ at $\omege^*(\ctext)$, and $\alpha_j(\thete, \ctext)$ ($\forall j\in J$) is twice differentiable with respect to $\thete$ at $\thete_0$.

\item 
The optimal solution $\omege_{\thete}(\ctext)$ to the oracle problem \eqref{equ:oracle:cont} satisfies that for any fixed $\ctext\in\mathcal{X}$, $\omege_{\thete}(\ctext)$ is twice differentiable with respect to $\thete$ at $\thete_0$.

\item Any involved operations of integration (expectation) and differentiation can be interchanged.
Specifically, for any $\thete\in\Theta$ and any $\ctext\in\mathcal{X}$,
$$ 
\nabla_\thete \int  \nabla_{\omege} \cost(\omege^*(\ctext),\ran)^\top p_\thete(\ran|\ctext) d\ran =  \int  \nabla_{\omege} \cost(\omege^*(\ctext),\ran)^\top \nabla_\thete p_\thete(\ran|\ctext)  d\ran, $$
$$ 
\int  \nabla_{\omege} \cost(\omege,\ran) p_\thete(\ran|\ctext) 
d\ran|_{\omege=\omege^*(\ctext)} =  \nabla_{\omege} \int  \cost(\omege,\ran) p_\thete(\ran|\ctext) d\ran|_{\omege=\omege^*(\ctext)}$$
where 
$\nabla_{\omege} \cost$ represents the gradient of $\cost$ over the first component.
\end{enumerate}
\end{assumption}

We adapt Assumption \ref{OCforZ:cons} in the non-contextual case to the following assumption:

\begin{assumption} [Conditions on constraints in contextual optimization] \label{OCforZ:cont}
Suppose that
the KKT conditions hold for the oracle problems \eqref{equ:oracle:cont} for all $\thete\in\Theta$ and $\ctext\in\mathcal{X}$. More specifically, for any fixed $\ctext\in\mathcal{X}$,
$\omege_\thete(\ctext)$ is a function of $\thete$ that satisfies
$$\nabla_{\omege} \vale(\omege, \thete|\ctext)+ \sum_{j\in J} \alpha_j(\thete,\ctext) \nabla_{\omege} g_j(\omege)|_{\omege=\omege_\thete(\ctext)} = 0, \quad \forall \thete\in\Theta$$
and complimentary slackness holds:
$$\alpha_j(\thete,\ctext) g_j(\omege_\thete(\ctext))=0, \quad \forall j\in J, \quad \forall \thete\in\Theta, \quad \forall \ctext\in\mathcal{X}.$$




\end{assumption}

It is worth mentioning that the complimentary slackness in Assumption \ref{OCforZ:cont} implies that
$$\hat{\alpha}_j^{ETO}(\ctext) g_j(\hat{\omege}^{ETO}(\ctext))=0, \quad \forall j\in J, \quad \forall \ctext\in\mathcal{X}$$
$$\hat{\alpha}_j^{IEO}(\ctext) g_j(\hat{\omege}^{IEO}(\ctext))=0, \quad \forall j\in J, \quad \forall \ctext\in\mathcal{X}$$
$$\alpha_j^*(\ctext) g_j(\omege^*(\ctext))=0, \quad \forall j\in J, \quad \forall \ctext\in\mathcal{X}.$$

We are now ready to state our main performance comparison result in this section:

\begin{theorem} [Stochastic ordering in contextual stochastic optimization] \label{SD3}
Suppose that there exists a $\thete_0\in \Theta$ such that $P(\ran|\ctext)=P_{\thete_0}(\ran|\ctext)$ for any $\ctext$.
Suppose Assumptions \ref{SCforh:cont}, \ref{OCforZ:cont}, \ref{consistency all:cont}, \ref{Regularity conditions for all:cont}  hold. Then we have $nR(\hat\omege^{\cdot})\xrightarrow{d}\mathbb G^\cdot$
for some limiting distribution $\mathbb G^\cdot=\mathbb G^{ETO}$, $\mathbb G^{IEO}$ when $\hat\omege^\cdot=\hat{\omege}^{ETO}$, $\hat{\omege}^{IEO}$ respectively. Moreover,
$\mathbb G^{ETO}\preceq_{st} \mathbb G^{IEO}$.
\end{theorem}

Theorem \ref{SD3} shows that the result in Theorem \ref{SD} also holds in the contextual stochastic optimization. The proof consists of the following steps. First, we derive and compare the conditional covariance matrices given a fixed context $\ctext$ appearing in the regret of two approaches, which is similar to what we did in Theorem \ref{SD2}. 
The main inequalities we leverage here are again the Cramer-Rao bound and Lemma \ref{lemma1}.
Then, we compare the average regret by taking the expectation over $P(\ctext)$ and using the matrix extension of the Cauchy-Schwarz inequality \citep{lavergne2008cauchy,tripathi1999matrix} to conclude the result.

Regarding assumption verification, the contextual case studied in this subsection largely follows the discussions for the non-contextual case in Sections \ref{sec:main} and \ref{sec:cons}. For unconstrained contextual problems, the verification strategy is similar to the unconstrained non-contextual case mentioned in Section \ref{sec:main}, with an additional layer on the distribution of the covariate $\ctext$ when passing the derivatives into expectations. Proposition \ref{prop: Verifying Assumptions} and its proof (in Appendix \ref{sec:proofs}) continue to offer guidance in this contextual setting. For constrained contextual problems, the challenges discussed in Section \ref{sec:cons} still remain and, as mentioned there, a thorough verification strategy of assumptions constitutes a substantial technical undertaking that merits a separate full-length study.


\subsection{Optimization under Misspecified Model}

Suppose now the parametric family  $\{P_\thete(\ran|\ctext): \thete\in \Theta\}$ is misspecified in the sense of Definition \ref{def:misspecified:cont}. Theorem \ref{misspecified:cont} shows that the result in Theorem \ref{misspecified} also holds in contextual stochastic optimization when comparing the two approaches from Section \ref{sec:algo:cont}.

\begin{theorem} [Contextual stochastic optimization under model misspecification] \label{misspecified:cont}
Suppose Assumption \ref{consistency all:cont} hold. Moreover, suppose that $\vale_0(\omege_\thete)$ is continuous with respect to $\thete$ at $\thete^*$ and $\thete^{KL}$. Then we have $R(\hat{\omege}^{IEO})\xrightarrow{P} \vale_0(\omege_{\thete^*})-\vale_0(\omege^*):=\kappa^{IEO}$, $R(\hat{\omege}^{ETO})\xrightarrow{P} \vale_0(\omege_{\thete^{KL}})-\vale_0(\omege^*):=\kappa^{ETO}$, and
$\kappa^{ETO}\geq\kappa^{IEO}\geq 0.$
\end{theorem}


\input{experiments}

\section{Conclusions and Discussions} \label{sec:conclusions}

In this paper, we theoretically compare the common data-driven optimization approaches of ETO, IEO and, whenever applicable, SAA. Our results show that, when the model is well-specified, ETO outperforms IEO, which in turn outperforms SAA. Conversely, when the model is misspecified, the performance ordering is completely reversed. These rankings are evaluated using first-order stochastic dominance, a strong criterion that considers the entire distribution of the regret rather than isolated metrics. Besides the methodological novelty in devising such a strong probabilistic comparison criterion, our results send a key message that is arguably contrary to common belief: While IEO is intuited to perform better than ETO, as the former utilizes the downstream optimization objective in obtaining the decision while the latter separates estimation from optimization, ETO can outperform IEO as long as the model is well-specified and the sample size is large. The key intuition in obtaining this insight is that, in the nonlinear objective setting, the regret typically has a quadratic approximation in terms of decisions where a Cramer-Rao-type argument can apply to conclude the superiority of ETO. Nonetheless, rigorously materializing this insight requires delicate analysis, especially in the constrained or contextual cases where model-based approaches like ETO and IEO are the most advantageous against SAA.

Our paper constitutes a first step towards the largely open question of statistical comparisons among data-driven optimization approaches, which ultimately has strong practical relevance. As discussed in the Introduction, while IEO is often believed to be superior, the reality is that it is much harder to solve computationally than ETO, as IEO can easily distort the tractability of the original optimization problem. This gives a key motivation for us to understand whether IEO is really necessary and outperforms ETO. To this end, we invent the analysis framework (comparison of regret at the distribution level), the right mathematical tool (first-order stochastic dominance), and obtain the first set of instance-specific comparison results. Even though we focus on the idealized asymptotic setting, and thus bear a gap with reality (which is finite-sample and exhibits at least some amount of model misspecification), our main results shed light that when the model is close to well-specified through e.g., good modeling or use of data, IEO may not outperform ETO and, given its computational demand, ETO becomes preferable. With this work as the starting point, our immediate next steps would be to study what happens if the model is nearly but not perfectly well-specified, which points to the derivation of finite-sample bounds that account for data size limitation and the amount of model misspecification, and at the same time sharp enough to allow for statistical comparisons. 

Besides the above, there are several related and natural directions to undertake. One is the effect of dimensionality that ties to finite-sample performances. This would involve the relative size of data versus model parameters as well as learning-theory model complexity. In particular, to build a (nearly) well-specified model, we often need to use a large number of parameters, which correspondingly inflates the estimation variance. That is, there is a tradeoff between this variance inflation from a large number of parameters versus the bias reduction coming from a more correct model specification. Second, as we have discussed, methods such as IEO and ETO are the most advantageous against SAA when considering contextual, constrained problems. The constrained setting in particular has required nontrivial extra assumptions (as well as extra techniques to handle the Lagrangian systems, the linear independence constraint qualification, and the new asymptotic normality of constrained SAA). A next direction would be to investigate the verification of these constrained-case assumptions, which appear quite open despite the line of literature on constrained stochastic optimization. Moreover, non-deterministic constraints, such as chance constraints or expected-value constraints, warrant further study, with new challenges arising from the need to ensure feasibility in addition to optimality. Lastly, we have considered parametric model-based approaches in this paper, and one direction is to study the nonparametric counterparts. To carry out asymptotic analysis like in this paper, nonparametric approaches introduce additional complications due to the different rates elicited by different methods, which typically also depend on the underlying tuning parameters (such as the bandwidth) in the nonparametric model (e.g., \cite{iyengar2024cross}). If we consider SAA as a special example of nonparametric method, then when compared to other nonparametric approaches such as kernel density estimation, SAA would have a better rate as it directly uses the empirical distribution. So, in terms of the first-order regret, SAA would likely beat these other nonparametric approaches. However, SAA fails to generalize in the contextual setting (as discussed in Section \ref{sec:algo:cont}), and thus it is important to consider other nonparametric approaches. When the rates of these approaches differ, then it is likely that one of them has a better first-order regret, whereas when they share the same rate, then it would necessitate the study of asymptotic variances that follows the analysis route in this paper. These studies would likely be intricate, case-by-case, and deserve an elaborate future work.

\section*{Acknowledgement}
We gratefully acknowledge support from the National Science Foundation under grants CMMI-1763000, CAREER CMMI-1834710, IIS-1849280, and the Cheung-Kong Innovation Doctoral Fellowship. The authors thank Luofeng Liao for helpful discussions on the paper \citep{duchi2021asymptotic}. The authors thank the anonymous reviewers for their constructive comments, which have greatly improved the quality of our paper.

\printbibliography



\newpage 
\begin{appendices}

\input{appendix}

\input{appendix_experiments}

\end{appendices}


\end{document}

%% file: experiments.tex
\section{Experiments} \label{sec:exp}
In this section, we conduct numerical experiments to support our findings, and provide insights for both small and large-sample regimes. Specifically, we compare the performances of data-driven stochastic optimization algorithms on the newsvendor problem across multiple problem settings, including unconstrained, constrained and contextual cases under well-specification (Section \ref{sec:all cases}), a spectrum of well-specified to misspecified cases (Section \ref{sec:well to mis}), problems with different dimensions (Section \ref{sec:dim}), and an example using real-world data (Section \ref{sec:realworld}). We conduct further experiments on another portfolio optimization problem (Section \ref{sec:exp:Portfolio}). We also record and briefly discuss computational runtimes in Appendix~\ref{sec:runtime}. All missing experimental details are given in Section \ref{sec:appendix_experiment}.\footnotemark \footnotetext{Please see our
GitHub repo for code and documentation: \url{https://github.com/yzhao3685/StocContextualOpti}}

\subsection{The Newsvendor Problem} \label{sec:exp:Newsvendor}
The multi-product newsvendor problem can be described as 
$$\min_{\bm \omege} 
\E_P\left[ \bm{h}^\top(\omege-\ran)^+ + \bm{b}^\top(\ran-\omege)^+\right],$$
where $\omegeun^{(j)}$ is the order quantity for product $j$, $h^{(j)}$ is the holding cost for product $j$, $b^{(j)}$ is the backlogging cost for product $j$, and $\ranun^{(j)}$ is the random variable representing the demand of product $j$. 
We define the multi-product newsvendor problem with a capacity constraint to include the constraint $\textbf{1}^\top\omege \le C$,
where $C$ represents a given upper bound on budget or capacity. We define the contextual newsvendor problem as 
$$
\min_{\omege(\cdot)} \E_{P} \left[(\bm{h}^\top (\omege(\bm\ctext)-\bm\ran)^+ + \bm{b}^\top(\bm\ran-\omege(\bm\ctext))^+\right],
$$
where $\omege(\cdot)$ maps a feature $\bm\ctext$ to a decision. 
We describe how the SAA, ETO, and IEO solutions are computed in Appendix~\ref{sec:appendix_experiment}.

\subsubsection{Unconstrained, constrained, and contextual settings.}\label{sec:all cases}
We assume throughout the experiments that the demand for the $p$ products are independent, and the backordering and holding costs for the $p$ products are equal. Specifically, we set the backordering costs to be $b^{(j)}=5$ and the holding cost to be $h^{(j)}=1$ for all $j\in[p]$.


{\bf Multi-product newsvendor problem. }
We generate a dataset $\{\ranun_i^{(j)}\}_{i=1}^n$ for each $j\in[p]$, where each product $j$ has demand distribution $\mathcal{N}(3j,1)$. For the well-specified setting, we assume each product $j$ has demand distribution  $\mathcal{N}(j\theteun,1)$, where $\theteun$ is the unknown parameter that we want to learn. Notice the ground truth $\theteun_0=3$. For the misspecified setting, we assume each product $j$ has demand distribution  $\mathcal{N}(j\theteun,1+0.9j)$., i.e., we use the wrong standard deviation. Note that the required technical assumptions to invoke Theorem \ref{correctlyspecified} (and Theorem \ref{misspecified} as well) hold for this problem setting, thanks to Proposition \ref{prop: Verifying Assumptions}. 

{\bf Multi-product newsvendor problem with a single capacity constraint.}  
We set the capacity constraint to be  $\sum_{j=1}^p \omegeun^{(j)} \leq 40$. 
The rest of the set up is the same as that of the unconstrained problem. Notice the sum of optimal ordering quantities exceed 40 in the unconstrained problem, and therefore the constraint will be active. 

{\bf Contextual newsvendor problem. }
We generate a dataset $\{\ctext_i,\ran_i\}_{i=1}^n$, where feature $\ctext\in\mathbb{R}^2$ is uniformly sampled from $[0,1]^2$, and the demand distribution is Gaussian with mean $(1,\ctext^\top)\thete= 2+(0.5,0.5)^\top \ctext$ and fixed variance $\sigma^2=1$. Notice the ground truth is $\theteun_0=(2, 0.5, 0.5)$. For the well-specified setting, we assume the demand distribution is $\mathcal{N}((1,\ctext^\top)\thete, 1)$, where $\thete$ are unknown parameters that we want to learn. For the misspecified setting, we assume the  demand distribution is Uniform$\sim(0,(1,\ctext^\top)\thete)$, where $\thete$ are unknown parameters that we want to learn. 



\begin{figure}[ht]
     \begin{tabular}{ccc} 
        \includegraphics[width=0.3\textwidth]{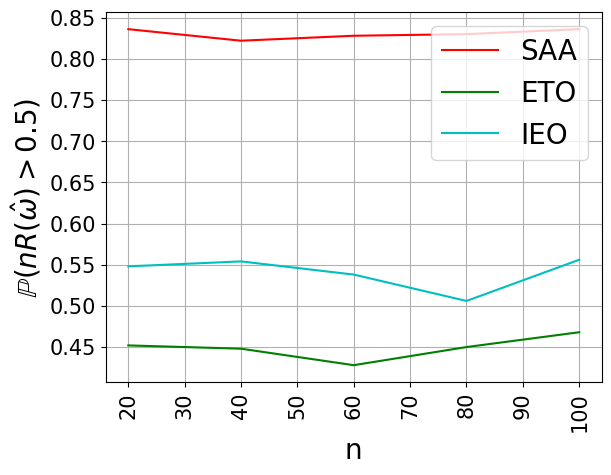}&
         \includegraphics[width=0.3\textwidth]{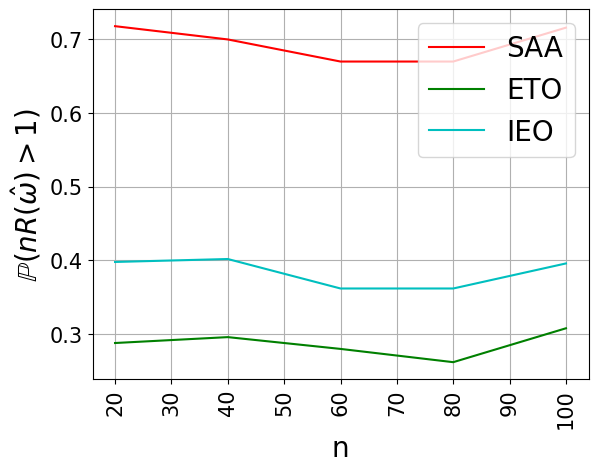}&
        \includegraphics[width=0.3\textwidth]{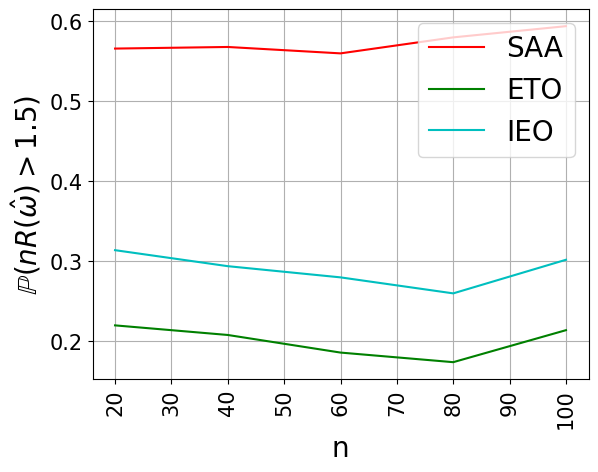}\\ 
        \includegraphics[width=0.3\textwidth]{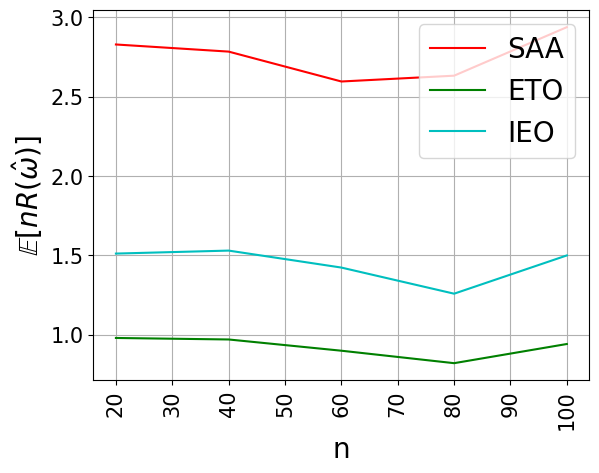}&      
        \includegraphics[width=0.3\textwidth]{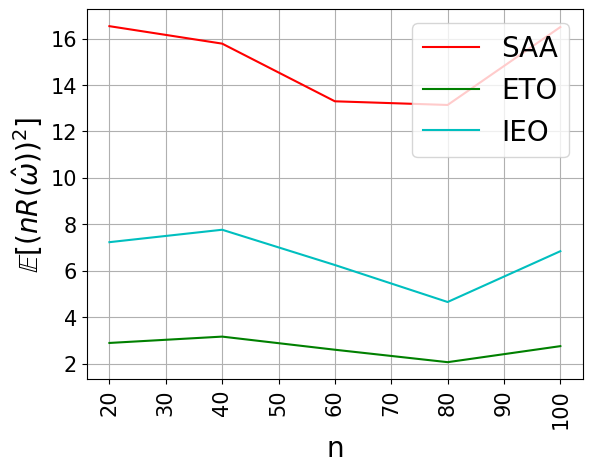}&
        \includegraphics[width=0.3\textwidth]{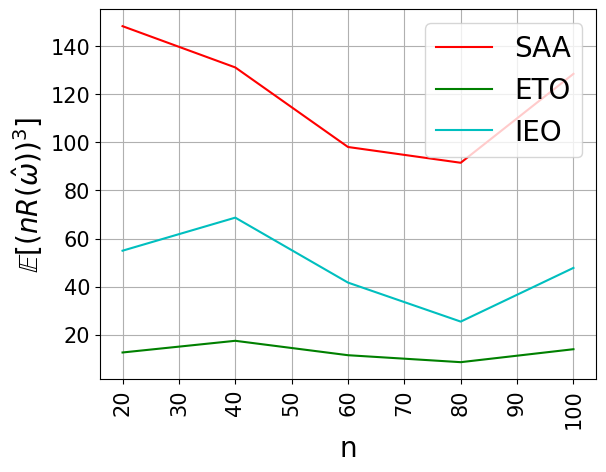}
     \end{tabular}
\caption{A multi-product newsvendor problem in the well-specified setting. The tail probability and moments are calculated over 500 random seeds. For this set of experiments, the number of products is $p=2$.}
\label{fig:verify_ast_experiment}
\end{figure}

In Figure~\ref{fig:verify_ast_experiment}, we present experimental results in the well-specified case to complement our theoretical results on stochastic dominance. Recall that our theoretical results (Theorems \ref{SD}, \ref{SD2},and \ref{SD3}) suggest that for any given $C_1>0$, we have the convergence in distribution $\mathbb{P}(n R(\hat{\omege}^{\cdot}) > C_1)\rightarrow C_2^{\cdot}(C_1)$ for some constant $C_2^{\cdot}(C_1)$ depending on $C_1$ where $\cdot=SAA,ETO,IEO$. 
Moreover, for any $C_1\in \mathbb{R}^+$, the stochastic dominance relation \eqref{def:stochasticdominance} implies that $C_2^{ETO}(C_1)\le C_2^{IEO}(C_1) \le C_2^{SAA}(C_1)$. Figure~\ref{fig:verify_ast_experiment} corroborates this finding. In this experiment, we set $C_1=0.5, 1, 1.5$, and the tail probability is estimated by the sample tail probability of the regret distribution over 500 simulations. 
In addition, we also evaluate the relation of the first, second, and third moments of $n R(\hat{\omege}^{\cdot})$. This is done by computing the first, second, and third sample moments of the $n R(\hat{\omege}^{\cdot})$ in 500 simulations.  
From Figure~\ref{fig:verify_ast_experiment}, we observe that although convergence in moments is not clear (as our theory does not indicate this), 
ETO has the smallest first, second, and third moments, while SAA has the largest first, second, and third moments, even in a small sample regime. 



\begin{figure}[ht]
  \begin{subfloat}
  \centering
     \begin{tabular}{ccc} 
     &\hspace{10mm}\textbf{Well-specified setting}&\\
        \includegraphics[width=0.3\textwidth]{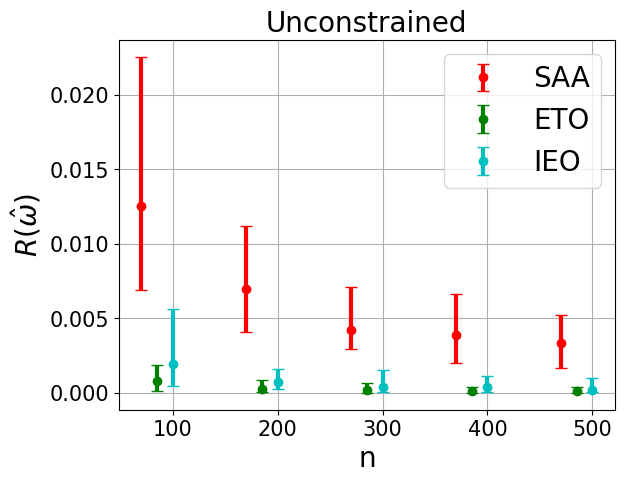}&
         \includegraphics[width=0.3\textwidth]{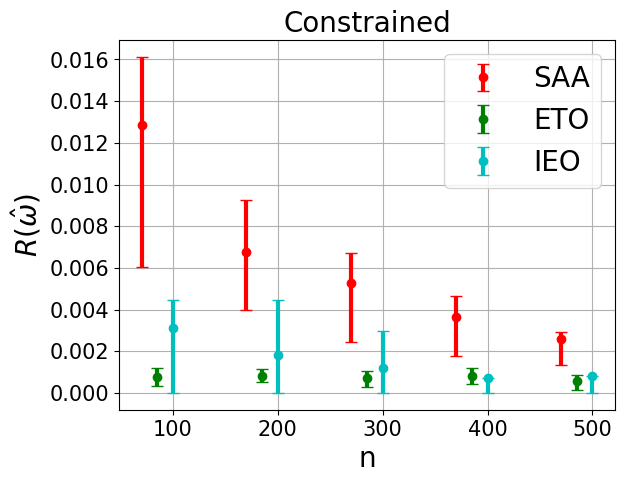}&
        \includegraphics[width=0.3\textwidth]{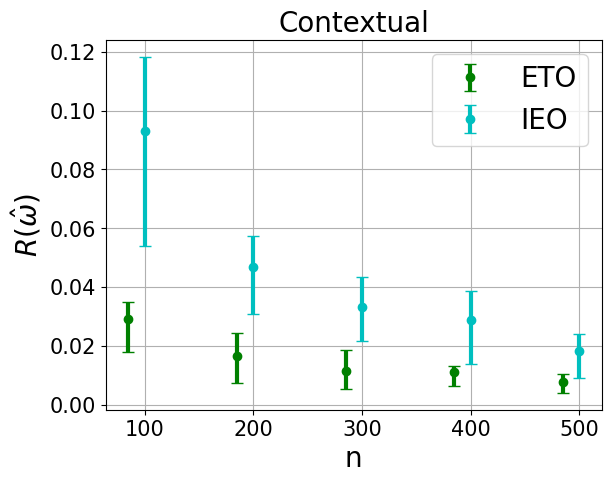}
     \end{tabular}
\label{fig:2a}
  \end{subfloat}%
  \begin{subfloat}
  \centering
     \begin{tabular}{ccc} 
     &\hspace{10mm}\textbf{Misspecified setting}&\\
        \includegraphics[width=0.3\textwidth]{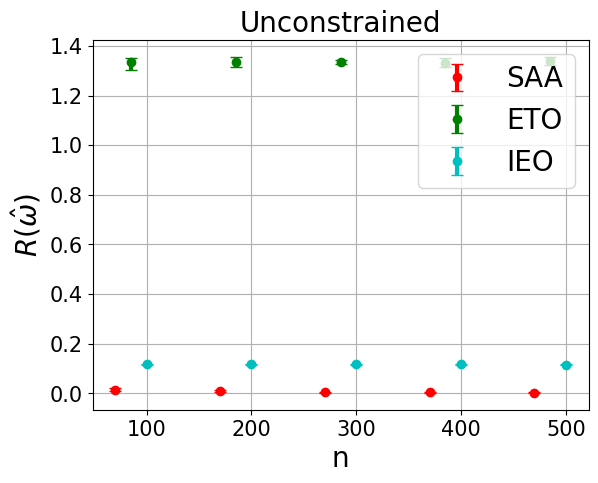}&      
        \includegraphics[width=0.31\textwidth]{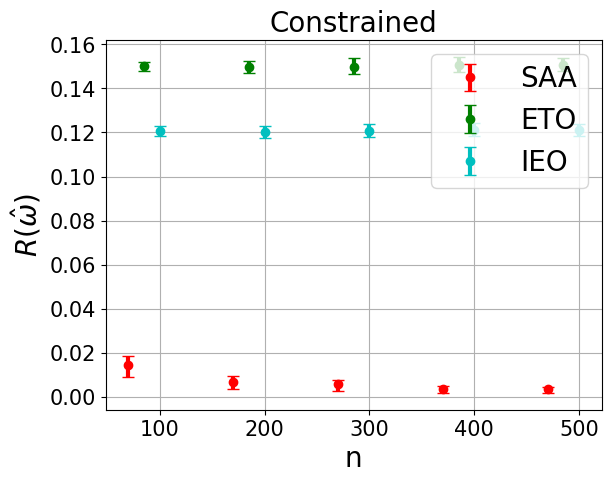}&
        \includegraphics[width=0.3\textwidth]{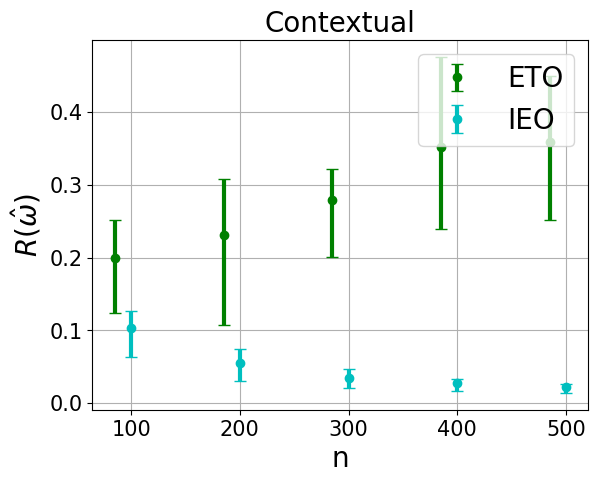}
     \end{tabular}
     \label{fig:2b}
  \end{subfloat}%
\caption{The regret plots show median, 25$^{th}$ quantile, and 75$^{th}$ quantile over 50 random seeds. For the unconstrained case and the constrained case, the number of products is $p=5$. For the contextual case, the number of products is $p=1$, similar to the setting in \citet{ban2019big}.}
\label{fig:news_vendor_finite_sample}
\end{figure}

Moreover, in Figure~\ref{fig:news_vendor_finite_sample}, we present experimental results for both well-specified and misspecified settings across multiple problem configurations for the newsvendor problem. We use sample size 10-50 for the unconstrained multi-product newsvendor problem and for the constrained multi-product newsvendor problem. 
The quantiles of regret in Figure~\ref{fig:news_vendor_finite_sample} show that regardless of the sample size, we observe the same trend that SAA is the worst approach and ETO is the best approach in the well-specified  setting, and the performance ordering is reversed in the misspecified setting. The same observation is made consistently for the unconstrained problem, the constrained problem, and the contextual problem. These results further support our findings on the performance of the three approaches.  

\subsubsection{From well-specified to misspecifed.}\label{sec:well to mis}

In this section, we conduct experiments to study how the degree of misspecification impacts the unconstrained multi-product newsvendor problem. We compare multiple model that get increasingly close to the well-specified setting.


Specifically, we generate a dataset $\{\ranun_i^{(j)}\}_{i=1}^n$ for each $j\in[5]$ where $p=5$, where each product $j$ has demand distribution $\mathcal{N}(3j,1)$. We assume each product $j$ has demand distribution  $\mathcal{N}(j\theteun, \gamma + (1-\gamma)(6-j))$ for some hyperparameter $\gamma$ and some unknown parameter $\theteun$ that we want to predict. Notice when $\gamma=1$, we are in the well-specified setting.

\begin{figure}[ht]
     \begin{tabular}{ccc} 
        \includegraphics[width=0.3\textwidth]{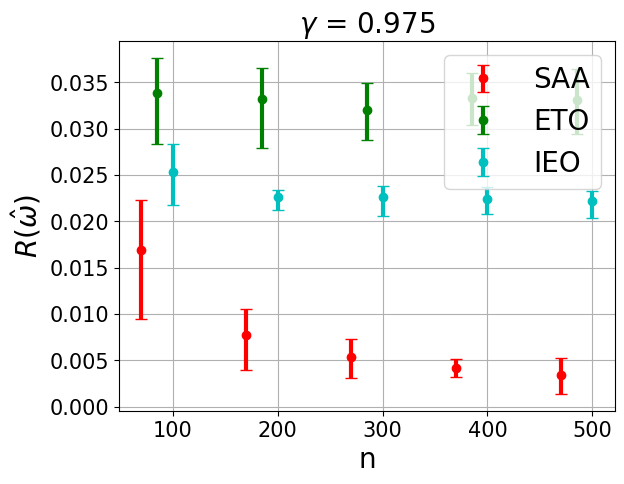}&
        \includegraphics[width=0.3\textwidth]{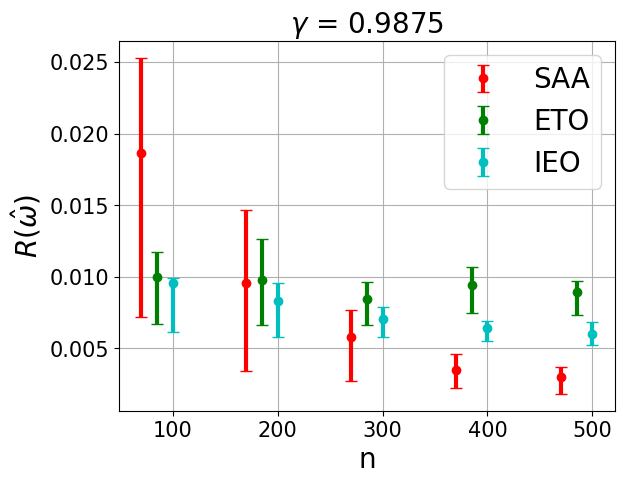}   
        \includegraphics[width=0.3\textwidth]{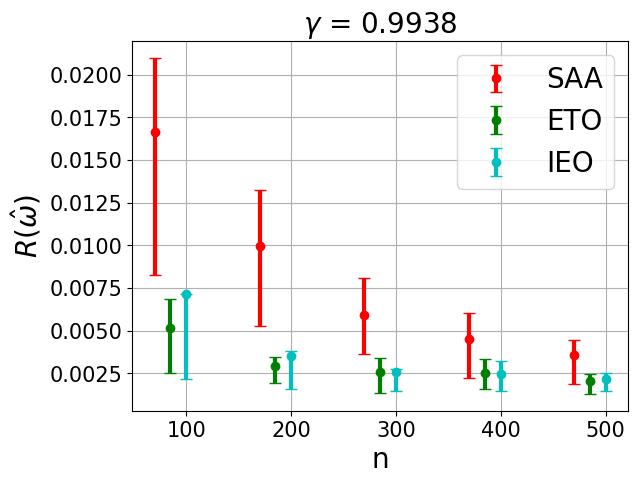}
     \end{tabular}
\caption{Results from well-specified to misspecified model. The regret plots show median, 25$^{th}$ quantile, and 75$^{th}$ quantile over 50 random seeds.}
\label{fig:between_correct_and_misspecifed}
\end{figure}

In Figure~\ref{fig:between_correct_and_misspecifed}, we present results for different values of $\gamma$. We observe that when $\gamma$ is close to 1 (i.e., the model is nearly well-specified), ETO has the best performance, and the regret of all three algorithms decreases as the sample size increases. On the other hand, as $\gamma$ moves away from 1 (i.e., the model deviates more from a well-specified model), the performance ordering of the three algorithms gradually reverses, and moreover, the regret of ETO and IEO no longer decreases as the sample size increases, as indicated by our Theorem \ref{misspecified}.

Lastly, we note that in the above experiments, SAA seems to consistently produce wider confidence intervals for the regret than IEO and in turn ETO, especially in the well-specified settings. This phenomenon appears coherent with Corollary \ref{cor:largervariance}.

\subsubsection{Dimensionality and finite-sample behaviors.}\label{sec:dim}
Our theoretical results on regret performances are asymptotic that most suit for a large sample size relative to problem dimension. To obtain a sense on whether our results apply to finite-sample situations, here we investigate the comparisons of different methods under varying dimensions of the decision variable $\omege$ and the parameter $\thete$. 

\begin{figure}[ht]
\centering
        \includegraphics[width=0.4\textwidth]{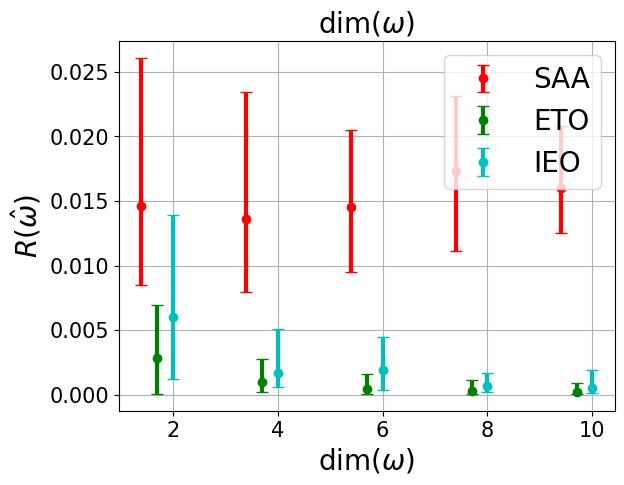}
\caption{Results for varying the relative dimensions of the decision and the parameter. The regret plots show median, 25$^{th}$ quantile, and 75$^{th}$ quantile over 50 random seeds. Results are for the unconstrained case, where the parameter dimension is fixed. Sample size is $n=100$.}
\label{fig:dimension_omege_theta}
\end{figure}

Our experiment focuses on the multi-product newsvendor problem in a well-specified model. In this setup, we fix the parameter dimension $q=1$ and vary the decision dimension $p$. This choice is motivated by our theoretical assumption that $p \ge q$ in unconstrained problems. Figure~\ref{fig:dimension_omege_theta} presents the results. We observe that, first of all, the performance ordering in the preference of ETO over IEO, and in turn over SAA, matches our theoretical result in Theorem~\ref{SD}. Second, we also see that the performance advantages of ETO and IEO over SAA become more pronounced as the decision dimension $p$ increases. This observation also aligns with the intuition of Theorem~\ref{SD}, which arises from the argument that in the well-specified setting, ETO and IEO effectively localize $\theta$ within the correct structured subset $\{\omege_\thete : \thete \in \Theta\}$, rather than optimizing over the full decision space $\{\omege : \omege \in \Omega\}$. This localization can be interpreted as a form of projection or regularization that constrains the search space to more meaningful decisions. As the dimensionality gap between $p$ and $q$ widens, this implicit regularization becomes more impactful, resulting in larger performance gains over SAA that does not exploit this structure.

In Appendix~\ref{sec:more dim}, we complement this analysis by exploring the opposite regime: we fix the decision dimension $p=1$ and increase the parameter dimension $q$, using the contextual newsvendor problem. There we discuss how these different relative dimensions affect performance comparisons and again connect to our theoretical findings.


\subsubsection{Real-world data.} \label{sec:realworld}

We continue to investigate the newsvendor problem, but now using real-world basket data from a retailer in 1997 and 1998 \citep{zhang2017assessing, oroojlooyjadid2020applying}. This dataset contains 13,170 observations made at different times of the year. Features available are categorical, representing day of the week (7 features), month of the year (12 features), and department (24 features). 
Note that, since we do not know the ground truth model for real-data situations, it necessitates some considerations as to how we can demonstrate alignments to our theoretical findings in this paper. We consider two initial approaches to motivate our ultimate approach, which will be a mix of using the data and additional assumptions.


Our initial attempt is to use the ``bootstrap", i.e., pretend that the observed data is the ground-truth, and that a ``data set" is uniformly generated from the observed data's support. In this way, the resampled data acts as our training set, which we can repeatedly draw for many times as in synthetic examples. To assess alignments with our theory, we can use the empirical loss as the ``ground-truth", since our theory holds for discrete distributions (in this case the empirical distribution). Then, the rest follows as in the previous subsections on synthetic examples. This is a natural attempt, and would work for the non-contextual case. However, in the contextual case, this approach will make the ``ground truth" feature values finite, and the ``ground truth" loss trivially 0 as the ``ground truth" optimal solution is simply the SAA solution that, as we have discussed in the beginning of Section \ref{sec:algo:cont}, fails to generalize. Thus, a more realistic ground-truth distribution assumption than merely using the bootstrap appears necessary.

We then consider another approach, which arguably remedies the generalization failure via the bootstrap in the contextual case, and moreover gives rise to a more ``honest" performance evaluation. This approach splits the real-world dataset into training and testing sets. We utilize the training set to run IEO and ETO, and evaluate their performance on the test set. This comparison, however, is restrictive as it only gives rise to one instance of the realized IEO and ETO solutions, and thus falls far off from reflecting distributional information which our main theorems aim to inform. Moreover, even considering this one instance as some estimate of the attained cost, this estimate is unbiased only for the mean (i.e., first moment of the) performance and hence the expected regret, but is biased for any higher-order moments of the regret, and thus cannot reliably reflect any rankings among the methods.


Given the above considerations, we adopt a hybrid approach that utilizes the data while making assumptions similar to synthetic examples. More precisely, we assume the ground truth is a Gaussian model and fix its parameters via an imputation from the real-world data. This allows us to know the ``ground truth" regret and consequently validate our theory in the contextual case. Following \citet{oroojlooyjadid2020applying}, we use $b/h$ = 5. We assume the demand follows $\mathcal{N}((1,\ctext^\top)\thete, \sigma)$, where the values of $\theta$ and $\sigma$ are data-imputed. For the well-specified setting, we assume a Gaussian demand model $\mathcal{N}((1,\ctext^\top)\thete, \sigma)$, where $\sigma$ is known (i.e., from the real-world data) and $\thete$ comprises the unknown parameters that we want to learn. For the misspecified setting, we assume the demand distribution is Uniform$(0,(1,\ctext^\top)\thete)$, where $\thete$ comprises the unknown parameters. 

\begin{figure}[ht]
\centering
     \begin{tabular}{ccc}    
     \hspace{3mm}\textbf{Correct Model} & \hspace{5mm}\textbf{Incorrect Model}\\
        \includegraphics[width=0.4\textwidth]{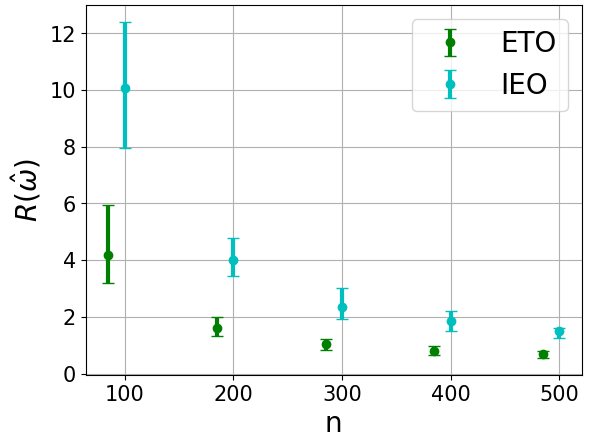}&
         \includegraphics[width=0.4\textwidth]{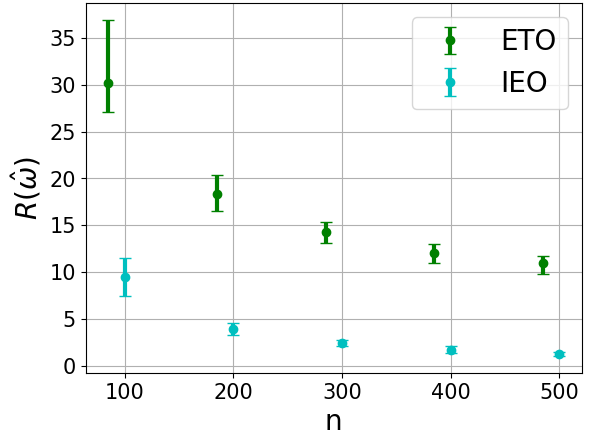}
     \end{tabular}
\caption{Real world data experiments with all available features. The regret plots show median, 25$^{th}$ quantile, and 75$^{th}$ quantile over 50 random seeds.}
\label{fig:real-world-all-features}
\end{figure}
\begin{figure}[ht]
\centering
     \begin{tabular}{ccc}    
     \hspace{3mm}\textbf{Correct Model} & \hspace{5mm}\textbf{Incorrect Model}\\
        \includegraphics[width=0.4\textwidth]{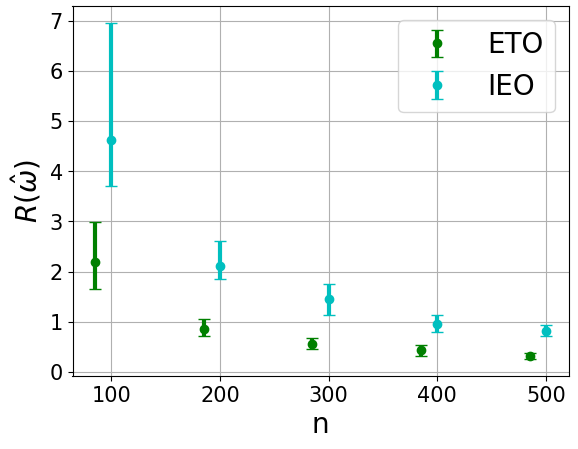}&      
        \includegraphics[width=0.4\textwidth]{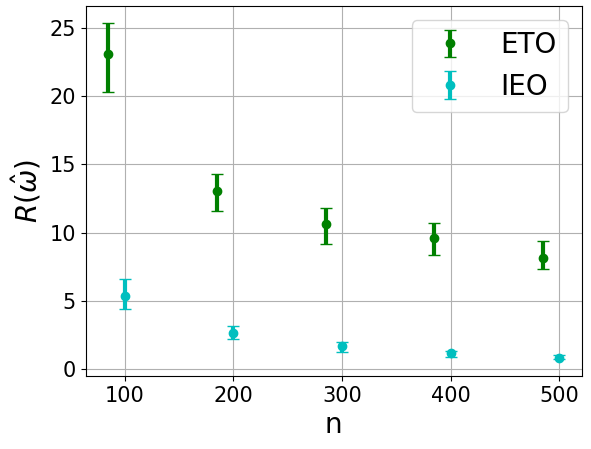}
     \end{tabular}
\caption{Real world data experiments with department features only. The regret plots show median, 25$^{th}$ quantile, and 75$^{th}$ quantile over 50 random seeds.}
\label{fig:real-world-department-features-only}
\end{figure}

Figures~\ref{fig:real-world-all-features} and ~\ref{fig:real-world-department-features-only} illustrate the results using our last adopted approach. In Figure~\ref{fig:real-world-all-features}, all 43 features available are used. In Figure~\ref{fig:real-world-department-features-only}, only the department features (24 features) are used. We observe that ETO has stronger performance than IEO in the well-specified setting, and IEO has stronger performance than ETO in the misspecified setting. Thus, similar to the previous subsections, our results support the theoretical findings on the performance comparisons between the two approaches.





\subsection{Portfolio Optimization} \label{sec:exp:Portfolio}
We consider the following objective as in \citet{kallus2022stochastic} and \citet{grigas2021integrated}:
$
c(\omege, \ran) := \alpha \left(\omege^\top (\ran_j,-1)\right)^2 -\omege^\top (\ran_j,0),
$
where the decisions $(\omegeun^{(1)},...,\omegeun^{(p-1)})\in\Delta^{p-1}$ represent the fraction of investments in asset $j$, and $\omegeun^{(p)}$ is an auxiliary variable. In the objective, the first term corresponds to the portfolio variance and the second term $\omege^\top (\ran, 0)$ represents the return. The expectation of the first term is $\E_P\left[\alpha \left(\omege^\top(\ran, -1)\right)^2\right] = 
\alpha\text{Var}\left(\omege^\top (\ran,0)\right)$, when $\omegeun^{(p)}$ is chosen optimally as $\mathbb{E}\left[\sum_{j=1}^{p-1}\omegeun^{(j)}\ranun^{(j)}\right]$ \citep{kallus2022stochastic,grigas2021integrated}. We describe how SAA, IEO, and ETO solutions are computed in Appendix~\ref{sec:appendix_experiment}. We assume the decisions $(\omegeun^{(1)},...,\omegeun^{(p-1)})\in\Delta^{p-1}$.

We generate a dataset $\{\ranun_i^{(j)}\}_{i=1}^n$ for each $j\in[p]$, where the return of each asset $j$ has distribution $\mathcal{N}(9+3j,3j)$. The $p$ assets are independent. For the well-specified setting, we assume the return of each asset $j$ has distribution $\mathcal{N}(\theteun^{(j)},3j)$, where $\theteun^{(j)}$ is the unknown parameter that we want to learn. For the misspecified setting, we assume the return of each asset $j$ has distribution $\mathcal{N}(\theteun^{(j)},3(p-j+1))$, i.e., we use the wrong standard deviation. We choose $\alpha=0.7$.

\begin{figure}[ht]
    \centering
     \begin{tabular}{cccc}
        \includegraphics[width=0.4\textwidth]{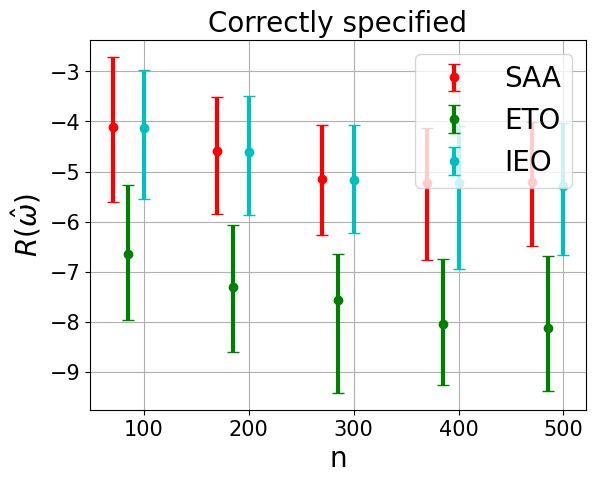}
        \includegraphics[width=0.4\textwidth]{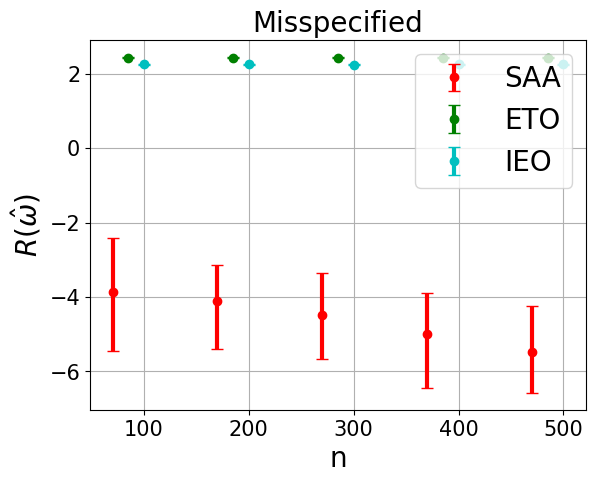}     
     \end{tabular}
\caption{A portfolio optimization problem. The regret plots show median, 25$^{th}$ quantile, and 75$^{th}$ quantile over 100 random seeds. For the unconstrained case and the constrained case, the number of assets is $p=2$. Notice we use log scale in the left figure to better illustrate the difference in performance among the algorithms.}
\label{fig:portfolio_opti}
\end{figure}

In Figure~\ref{fig:portfolio_opti}, we present experimental results for both well-specified and misspecified settings for the portfolio optimization problem. We choose sample size range from 10-50, which is the same range as in newsvendor problems. We observe the same trend as in newsvendor problems, specifically SAA is the worst approach and ETO is the best approach in the well-specified setting, and the performance ordering is reversed in the misspecified setting.

%% file: appendix.tex

\section{Technical Details for Section \ref{sec:uncon}} \label{sec:techdetails}


\textbf{Details on first-order stochastic dominance.}
The first-order stochastic dominance has the following important properties described in Lemma \ref{prop1}. This result is adopted from \citet{mas1995microeconomic,shaked2007stochastic}, but because of its importance in this paper, we provide a self-contained proof here. 

\begin{lemma} \label{prop1}
The following statements are equivalent:\\
(a) $X \preceq_{st} Y$.\\
(b) There exists a random vector $Z$ and two functions $\psi_1$, $\psi_2$ such that $X \overset{d}{=} \psi_1(Z), Y \overset{d}{=} \psi_2(Z)$ and $\psi_1\le \psi_2$.\\
(c) For any increasing functions $\phi$, $\mathbb{E}[\phi(X)]\le \mathbb{E}[\phi(Y)]$.
\end{lemma}

\proof{Proof of Lemma \ref{prop1}}

We show $(a) \implies (b)$. Let $F_X$ and $F_Y$ be the cumulative distribution function of $X$ and $Y$ respectively. Let $F_{X}^{-1}$ and $F_{Y}^{-1}$ be the generalized inverse of the cumulative distribution function $F_X$ and $F_Y$ respectively.
Let $Z$ be a uniformly distributed random variable
$Z\sim Uniform(0,1)$. Then the inverse transform sampling implies that $F_{X}^{-1}(Z)\overset{d}{=}X$, and $F_{Y}^{-1}(Z)\overset{d}{=}Y$. For any $x\in \mathbb{R}$, stochastic dominance \eqref{def:stochasticdominance} implies that $1-F_X(x)\le 1-F_Y(x)$, thus $F_X(x)\ge F_Y(x)$, and thus $F^{-1}_X(x)\le F^{-1}_Y(x)$. This shows that $\psi_1:= F^{-1}_X$ and $\psi_2:= F^{-1}_Y$ are as desired.

We show $(b) \implies (c)$. Note that by (b), we have
\begin{align*}
\mathbb{E}[\phi(X)]&= \mathbb{E}[\phi(\psi_1(Z))]\le \mathbb{E}[\phi(\psi_2(Z))] = \mathbb{E}[\phi(Y)]
\end{align*}
since $\psi_1(Z) \le \psi_2(Z)$ holds point-wise and thus $\phi(\psi_1(Z))\le \phi(\psi_2(Z))$ holds point-wise.

We show $(c) \implies (a)$. For any $x\in \mathbb{R}$, we let $\phi(t):=\bm{1}_{(x,+\infty)}(t)$ which is an increasing function. Then 
$$\mathbb{E}[\phi(X)]= \mathbb{E}[\bm{1}_{(x,+\infty)}(X)]= \mathbb{P}[X>x],$$
$$\mathbb{E}[\phi(Y)]= \mathbb{E}[\bm{1}_{(x,+\infty)}(Y)]= \mathbb{P}[Y>x],$$
Hence (c) implies that
$\mathbb{P}[X>x]=\mathbb{E}[\phi(X)]\le \mathbb{E}[\phi(Y)]=\mathbb{P}[Y>x]$, as desired.
\Halmos
\endproof

When $X$, $Y$ are both nonnegative random variables and $X \preceq_{st} Y$, then part (c) in Lemma \ref{prop1} implies that $\mathbb{E}[X^k]\le \mathbb{E}[Y^k]$ for any $k>0$. However, it is worth mentioning that the first-order stochastic dominance relation is even stronger than the relation on any moments of the distributions. Consider the following example: Suppose $X$ is 1 with probability $1$ and $Y$ is distributed as $\mathbb{P}[Y=2]=\mathbb{P}[Y=\frac{1}{2}]=\frac{1}{2}$. Then $\mathbb{E}[X^k]=1\le \frac{1}{2}(2^{k}+2^{-k})=\mathbb{E}[Y^k]$ for any $k>0$. However, $X$
is not first-order stochastically dominated by $Y$ since $\mathbb{P}[X>\frac{1}{2}]=1>\frac{1}{2}=\mathbb{P}[Y>\frac{1}{2}]$.


\textbf{Proving Proposition \ref{consistency of all}.} The following result is adopted from Theorem 5.7 in \citet{van2000asymptotic}.

\begin{lemma}[Consistency of M-estimation]\label{prop3}
Suppose the random variable $\ran$ follows the distribution $P$. Suppose that $m(\zeta,\ran)$ is a measurable function of $\ran$ such that
\begin{enumerate}
\item $\sup_{\zeta} |\frac{1}{n}\sum_{i=1}^n m(\zeta,\ran_i)-\mathbb{E}_{P}[m(\zeta,\ran)]|\xrightarrow{P} 0$. 
\item For every $\epsilon> 0$,
$\sup_{\zeta: d(\zeta,\zeta^*)\ge \epsilon} \mathbb{E}_{P}[m(\zeta,\ran)] < \mathbb{E}_{P}[m(\zeta^*,\ran)]$ where $\zeta^*=\argmax_{\zeta}\mathbb{E}_{P}[m(\zeta,\ran)]$.

\item The random sequence $\hat\zeta_n$ satisfies that $$\frac{1}{n}\sum_{i=1}^n m(\hat\zeta_n, \ran_i)\ge \frac{1}{n}\sum_{i=1}^n m(\zeta^*, \ran_i)-o_{P}(1),$$
\end{enumerate}
Then $\hat\zeta_n\xrightarrow{P}\zeta^*$.
\end{lemma}

\proof{Proof of Proposition \ref{consistency of all}}

For SAA, consider $m(\zeta,\ran)=-\cost(\omege,\ran)$ with the parameter $\zeta=\omege$ and apply Lemma \ref{prop3}.

For ETO, consider $m(\zeta,\ran)=\log p_\thete(\ran)$ with the parameter $\zeta=\thete$  and apply Lemma \ref{prop3}.

For IEO, consider $m(\zeta,\ran)=-\cost(\omege_\thete,\ran)$ with the parameter $\zeta=\thete$ and apply Lemma \ref{prop3}.
\Halmos
\endproof

\textbf{Proving Proposition \ref{Regularity conditions for all}.} 
The following result is adopted from Theorem 5.23 in \citet{van2000asymptotic}.

\begin{lemma}[Asymptotic normality of M-estimation]\label{prop2}
Suppose the random variable $\ran$ follows the distribution $P$. Suppose that $m(\zeta,\ran)$ is a measurable function of $\ran$ such that $\zeta\mapsto m(\zeta,\ran)$ is differentiable at $\zeta^*$ for almost every $\ran$ (with respect to $P$) with derivative $\nabla_\zeta m(\zeta^*,\ran)$. Moreover, for any $\zeta_1$ and $\zeta_2$ in a neighborhood of $\zeta^*$, there exists a measurable function $K$ with $\mathbb{E}_{P}[K(\ran)]<\infty$ such that
$|m(\zeta_1,\ran)-m(\zeta_2,\ran)|\le K(\ran) \|\zeta_1-\zeta_2\|.$ Furthermore, the map $\zeta\mapsto \mathbb{E}_{P}[m(\zeta,\ran)]$ admits a second-order Taylor expansion at a point of maximum $\zeta^*$ with nonsingular symmetric second derivative matrix $V_{\zeta^*}$.
If the random sequence $\hat\zeta_n$ satisfies  $\hat\zeta_n\xrightarrow{P}\zeta^*$ and 
$$\frac{1}{n}\sum_{i=1}^n m(\hat\zeta_n, \ran_i)\ge \sup_{\zeta} \frac{1}{n}\sum_{i=1}^n m(\zeta, \ran_i)-o_{P}(n^{-1}),$$    
then $\sqrt{n} (\hat\zeta_n- \zeta^*)$ is asymptotically normal with mean zero and covariance matrix 
$$V_{\zeta^*}^{-1}Var_{P}(\nabla_\zeta m(\zeta^*,\ran))V_{\zeta^*}^{-1}$$
where $Var_{P}(\nabla_\zeta m(\zeta^*,\ran))$ is the covariance matrix of the cost gradient $\nabla_\zeta m(\zeta^*,\ran)$ under $P$.
\end{lemma}

\proof{Proof of Proposition \ref{Regularity conditions for all}}

For SAA, consider $m(\zeta,\ran)=-\cost(\omege,\ran)$ with the parameter $\zeta=\omege$ and apply Lemma \ref{prop2}.

For ETO, consider $m(\zeta,\ran)=\log p_\thete(\ran)$ with the parameter $\zeta=\thete$  and apply Lemma \ref{prop2}.

For IEO, consider $m(\zeta,\ran)=-\cost(\omege_\thete,\ran)$ with the parameter $\zeta=\thete$ and apply Lemma \ref{prop2}.
\Halmos
\endproof

\section{Technical Details for Section \ref{sec:cont}} \label{sec:cont:more}

\subsection{Basic Statistical Results on Consistency}
The following assumptions are in parallel to Assumptions \ref{ETOconsistency}, \ref{IEOconsistency}.

\begin{subtheorem}{assumption}\label{consistency all:cont}
\begin{assumption} [Consistency conditions for ETO] \label{ETOconsistency:cont:assm}
Suppose that
\begin{enumerate}
\item $$\sup_{\thete\in \Theta} |\frac{1}{n}\sum_{i=1}^n \log p_\thete (\ran_i|\ctext_i)-\mathbb{E}_{P} [ \log p_\thete(\ran|\ctext)] |\xrightarrow{P} 0.$$ 
\item For every $\epsilon> 0$,
$$\sup_{\thete\in \Theta: d(\thete,\thete^{KL})\ge \epsilon} \mathbb{E}_{P} [ \log p_\thete(\ran|\ctext)] < \mathbb{E}_{P} [ \log p_{\thete^{KL}}(\ran|\ctext)]$$
and $\thete^{KL}$ is
$$\thete^{KL}:= \argmax_{\thete\in \Theta} \mathbb{E}_{P} [ \log p_{\thete}(\ran|\ctext)] =\argmax_{\thete\in \Theta} \mathbb{E}_{P} [ \log p_{\thete}(\ctext,\ran)]= \argmin_{\thete\in \Theta} KL(P, P_\thete(\ctext,\ran)),$$
where $KL$ represents the Kullback-Leibler divergence. The second equality is because $p_\thete(\ctext,\ran)=p_\thete(\ran|\ctext)p(\ctext)$ where $p(\ctext)$ is independent of $\thete$. 
\item The estimated model parameter $\hat{\thete}^{ETO}$ in ETO is solved approximately in the sense that
$$\frac{1}{n}\sum_{i=1}^n \log p_{\hat{\thete}^{ETO}} (\ran_i|\ctext_i)\ge \frac{1}{n}\sum_{i=1}^n \log p_{\thete^{KL}} (\ran_i|\ctext_i)-o_{P}(1).$$ 
\end{enumerate}
\end{assumption}

\begin{assumption} [Consistency conditions for IEO] \label{IEOconsistency:cont:assm}
Suppose that
\begin{enumerate}
\item $$\sup_{\thete\in \Theta} |\hat{\vale}_0(\omege_\thete)-\vale_0(\omege_\thete)|\xrightarrow{P} 0.$$
\item For every $\epsilon> 0$,
$$\inf_{\thete\in \Theta: d(\thete,\thete^*)\ge \epsilon} \vale_0(\omege_\thete) > \vale_0(\omege_{\thete^*})$$
where $\thete^*$ is given by
$\thete^*:= \argmin_{\thete\in \Theta} \vale_0(\omege_\thete).$
\item The estimated model parameter $\hat{\thete}^{IEO}$ in IEO is solved approximately in the sense that
$$\hat{\vale}_0(\omege_{\hat{\thete}^{IEO}})\le \hat{\vale}_0(\omege_{\thete^*})+o_{P}(1).$$ 
\end{enumerate}
\end{assumption}
\end{subtheorem}

Using the same argument as in Proposition \ref{consistency of all}, we have the consistency for IEO and ETO.

\begin{subtheorem}{proposition}\label{consistency of all:cont}
\begin{proposition} [Consistency of ETO] \label{ETOconsistency:cont}
Suppose Assumption \ref{ETOconsistency:cont:assm} holds. We have $\hat{\thete}^{ETO}\xrightarrow{P}\thete^{KL}$.    
\end{proposition}

\begin{proposition} [Consistency of IEO]
\label{IEOconsistency:cont}
Suppose Assumption \ref{IEOconsistency:cont:assm} holds. We have $\hat{\thete}^{IEO}\xrightarrow{P}\thete^*$.    
\end{proposition}

\end{subtheorem}

\subsection{Basic Statistical Results on Asymptotic Normality}


We adapt Assumption \ref{RCforETO:assm}, \ref{RCforIEO:assm} in the non-contextual case to the following assumptions. 

\begin{subtheorem}{assumption}\label{Regularity conditions for all:cont}
\begin{assumption} [Regularity conditions for ETO] \label{RCforETO:cont:assm}
Suppose that $\log p_\thete(\ran|\ctext)$ is a measurable function of $(\ran,\ctext)$ such that $\thete\mapsto \log p_\thete(\ran|\ctext)$ is differentiable at $\thete^{KL}$ for almost every $(\ran,\ctext)$ with derivative $\nabla_\thete \log p_{\thete^{KL}}(\ran|\ctext)$ and such that for any $\thete_1$ and $\thete_2$ in a neighborhood of $\thete^{KL}$, there exists a measurable function $K$ with $\mathbb{E}_{P}[K(\ran,\ctext)]<\infty$ such that
$|\log p_{\thete_1}(\ran|\ctext)-\log p_{\thete_2}(\ran|\ctext)|\le K(\ran, \ctext) \|\thete_1-\thete_2\|.$
Furthermore, the map $\thete\mapsto \mathbb{E}_{P}[\log p_\thete(\ran|\ctext)]$ admits a second-order Taylor expansion at the point of maximum $\thete^{KL}$ with nonsingular symmetric second derivative $\nabla_{\thete\thete}\mathbb{E}_{P}[\log p_{\thete}(\ran|\ctext)]|_{\thete=\thete^{KL}}$. Moreover, 
$\hat{\thete}^{ETO}$ is obtained approximately in the sense that
$$\frac{1}{n}\sum_{i=1}^n \log p_{\hat{\thete}^{ETO}} (\ran_i|\ctext_i)\ge \sup_{\thete\in \Theta}\frac{1}{n}\sum_{i=1}^n \log p_{\thete} (\ran_i|\ctext_i)-o_{P}(n^{-1}).$$
\end{assumption}

\begin{assumption} [Regularity conditions for IEO] \label{RCforIEO:cont:assm}
Suppose that $\cost(\omege_\thete(\ctext),\ran)$ is a measurable function of $(\ran,\ctext)$ such that $\thete\mapsto \cost(\omege_\thete(\ctext),\ran)$ is differentiable at $\thete^*$ for almost every $(\ran,\ctext)$ with derivative $\nabla_\thete \cost(\omege_\thete(\ctext),\ran)$ and such that for any $\thete_1$ and $\thete_2$ in a neighborhood of $\thete^*$, there exists a measurable function $K$ with $\mathbb{E}_{P}[K(\ran,\ctext)]<\infty$ such that
$|\cost(\omege_{\thete_1}(\ctext),\ran)-\cost(\omege_{\thete_2}(\ctext),\ran)|\le K(\ran,\ctext) \|\thete_1-\thete_2\|.$ Furthermore, the map $\thete\mapsto \vale_0(\omege_{\thete})$ admits a second-order Taylor expansion at the point of minimum $\thete^*$ with nonsingular symmetric second derivative $\nabla_{\thete\thete}\vale_0(\omege_{\thete^*})$. Moreover, 
$\hat{\thete}^{IEO}$ is solved approximately in the sense that
$$\hat{\vale}_0(\omege_{\hat{\thete}^{IEO}})\le \inf_{\thete\in \Theta} \hat{\vale}_0(\omege_{\thete})+o_{P}(n^{-1}).$$
\end{assumption}
\end{subtheorem}

Using the same argument as in Proposition \ref{Asymptotic normality for all}, we have the asymptotic normality for IEO and ETO.

\begin{subtheorem}{proposition}\label{Asymptotic normality for all:cont}
\begin{proposition} [Asymptotic normality for ETO] \label{RCforETO:cont}
Suppose that Assumptions  \ref{ETOconsistency:cont:assm} and \ref{RCforETO:cont:assm} hold. Then
$\sqrt{n} (\hat{\thete}^{ETO}- \thete^{KL})$ is asymptotically normal with mean zero and covariance matrix 
\begin{equation}\label{cov matrix:cont}
(\nabla_{\thete\thete}\mathbb{E}_{P}[\log p_{\thete}(\ran|\ctext)]|_{\thete=\thete^{KL}})^{-1}
Var_{P}(\nabla_\thete \log p_{\thete^{KL}}(\ran|\ctext))(\nabla_{\thete\thete}\mathbb{E}_{P}[\log p_{\thete}(\ran|\ctext)]|_{\thete=\thete^{KL}})^{-1}    
\end{equation}
where $Var_{P}(\nabla_\thete \log p_{\thete^{KL}}(\ran|\ctext))$ is the covariance matrix of $\nabla_\thete \log p_{\thete^{KL}}(\ran|\ctext)$ under the joint distribution $P$.
Moreover, when $\thete^{KL}$ corresponds to the ground-truth $P$, i.e., $P_{\thete^{KL}}=P$, the covariance matrix \eqref{cov matrix:cont} is simplified to the inverse Fisher information $\mathcal{I}_{\thete^{KL}}^{-1}$ under the joint distribution, that is,
$$\eqref{cov matrix:cont}= \mathcal{I}_{\thete^{KL}}^{-1}=(\mathbb{E}_{P}[(\nabla_\thete \log p_{\thete^{KL}}(\ran|\ctext))^\top \nabla_\thete \log p_{\thete^{KL}}(\ran|\ctext)])^{-1}.$$
\end{proposition}
Note that $p_{\thete}(\ctext,\ran)=p_{\thete}(\ran|\ctext)p(\ctext)$ where $p(\ctext)$ is independent of $\thete$ so we can equivalently write $\mathcal{I}_{\thete^{KL}}$ as
$$\mathcal{I}_{\thete^{KL}}=\mathbb{E}_{P}[(\nabla_\thete \log p_{\thete^{KL}}(\ctext,\ran))^\top \nabla_\thete \log p_{\thete^{KL}}(\ctext,\ran)].$$
In addition, we introduce the Fisher information $\mathcal{I}_{\thete^{KL}}(\ctext)$ under the conditional distribution:
$$\mathcal{I}_{\thete^{KL}}(\ctext)=\mathbb{E}_{P(\ran|\ctext)}[(\nabla_\thete \log p_{\thete^{KL}}(\ran|\ctext))^\top \nabla_\thete \log p_{\thete^{KL}}(\ran|\ctext)].$$ 
It is clear that we have 
$\mathbb{E}_{P(\ctext)}[\mathcal{I}_{\thete^{KL}}(\ctext)]=\mathcal{I}_{\thete^{KL}}$.

\begin{proposition} [Asymptotic normality for IEO] \label{RCforIEO:cont}
Suppose that Assumptions  \ref{IEOconsistency:cont:assm} and \ref{RCforIEO:cont:assm} hold. Then
$\sqrt{n} (\hat{\thete}^{IEO}- \thete^*)$ is asymptotically normal with mean zero and covariance matrix 
$$\nabla_{\thete\thete}\vale_0(\omege_{\thete^*})^{-1} Var_{P}(\nabla_\thete \cost(\omege_{\thete^*}(\ctext),\ran))\nabla_{\thete\thete}\vale_0(\omege_{\thete^*})^{-1}$$
where $Var_{P}(\nabla_\thete \cost(\omege_{\thete^*}(\ctext),\ran))$ is the covariance matrix of the cost gradient $\nabla_\thete \cost(\omege_{\thete^*}(\ctext),\ran)$ under $P_{\thete_0}(\ctext, \ran)$.
\end{proposition}
\end{subtheorem}

\section{The Matrix Extension of the Cauchy-Schwarz Inequality and the Multivariate Cramer-Rao Bound} \label{sec:multivariateCramerRao}

In this section, we state existing results on the matrix extension of the Cauchy-Schwarz inequality and the multivariate Cramer-Rao bound that are leveraged in our proof. We provide self-contained proofs for these results for completeness.

The following is the matrix extension of the Cauchy-Schwarz inequality.

\begin{lemma} [Cauchy-Schwarz inequality, Lemma 2 in \citep{lavergne2008cauchy}] \label{lemma:Cauchy-Schwarz}
Let $Q_1 \in \mathbb{R}^{t_1\times t_2}$ and $Q_2 \in \mathbb{R}^{t_1\times t_3}$ be random matrices such that
$\mathbb{E}[\|Q_1\|^2_F]< +\infty$, $\mathbb{E}[\|Q_2\|^2_F]< +\infty$ where $\|\cdot\|_F$ is the Frobenius norm of the matrix, and $\mathbb{E}[Q_1^\top Q_1]$ is nonsingular.
Then 
$$\mathbb{E}[Q_2^\top Q_2] - \mathbb{E}[Q_2^\top Q_1] (\mathbb{E}[Q_1^\top Q_1])^{-1} \mathbb{E}[Q_1^\top Q_2]\ge 0$$ 
with
equality if and only if $Q_2 = Q_1 (\mathbb{E}[Q_1^\top Q_1])^{-1} \mathbb{E}[Q_1^\top Q_2].$   
\end{lemma} 

\proof{Proof of Lemma \ref{lemma:Cauchy-Schwarz}} 
Consider $\Lambda = (\mathbb{E}[Q_1^\top Q_1])^{-1} \mathbb{E}[Q_1^\top Q_2] \in \mathbb{R}^{t_2\times t_3}$. Then
$$\mathbb{E}[(Q_2 - Q_1 \Lambda)^\top (Q_2 - Q_1 \Lambda)]
= \mathbb{E}[Q_2^\top Q_2] - \mathbb{E}[Q_2^\top Q_1] (\mathbb{E}[Q_1^\top Q_1])^{-1} \mathbb{E}[Q_1^\top Q_2]$$
is always positive semidefinite as it is the expectation of a matrix product $(Q_2 - Q_1 \Lambda)^\top (Q_2 - Q_1 \Lambda)\ge 0$, and is zero if and only if $Q_2 = Q_1 \Lambda$.
\Halmos
\endproof

The following is the well-known multivariate Cramer-Rao bound, which is slightly more general than the classical form \citep{cramer1946mathematical,rao1945information} where we allow the dimension of the estimator to be different from the parameter. Interestingly, we can use Lemma \ref{lemma:Cauchy-Schwarz} to provide a concise proof.

\begin{lemma} [Multivariate Cramer-Rao bound] \label{lemma:Cramer-Rao}
Let $\thete\in \mathbb{R}^{t_1}$ be a parameter and $\bm{T}(\bm{X})\in\mathbb{R}^{t_2}$ (viewed as a column vector in $\mathbb{R}^{t_2\times 1}$) be an estimator where it is possible that $t_1\ne t_2$. Let $\bm{\psi}$ be the expectation of $\bm{T}(\bm{X})$, that is, $\bm{\psi}: \mathbb{R}^{t_1}\to \mathbb{R}^{t_2}, \thete\mapsto \mathbb{E}_{\thete}[\bm{T}(\bm{X})]:=\int \bm{T}(\bm{x}) p_\thete(\bm{x})  d\bm{x}$. Assume that the interchangeable condition holds:
$$ 
\nabla_\thete \int \bm{T}(\bm{x}) p_\thete(\bm{x})  d\bm{x} =  \int \bm{T}(\bm{x}) \nabla_\thete p_\thete(\bm{x})  d\bm{x},
$$
for any $\thete$. If $\bm{\psi}(\thete)$ is differentiable, then 
$$Var_{\thete}(\bm{T}(\bm{X}))\ge \nabla_{\thete}\bm{\psi}(\thete) \mathcal{I}_{\thete}^{-1} \nabla_{\thete}\bm{\psi}(\thete)^\top$$
for any $\thete$.   
\end{lemma} 

\proof{Proof of Lemma \ref{lemma:Cramer-Rao}}
Let $\bm{Y}_{\thete}(\bm{x}):=\nabla_\thete \log p_\thete(\bm{x})^\top\in \mathbb{R}^{t_1\times 1}$.
Recall that by definition,
$$\mathcal{I}_{\thete}=\mathbb{E}_{\thete}[\bm{Y}_{\thete}(\bm{X})\bm{Y}_{\thete}(\bm{X})^\top]$$
and
\begin{align*}
&\mathbb{E}_{\thete}[(\bm{T}(\bm{X})-\bm{\psi}(\thete)) \bm{Y}_{\thete}(\bm{X})^\top]\\
=&\mathbb{E}_{\thete}[\bm{T}(\bm{X})\bm{Y}_{\thete}(\bm{X})^\top] \quad \text{since } \mathbb{E}_{\thete}[\bm{Y}_{\thete}(\bm{X})^\top]=\bm{0}\\
=& \int \bm{T}(\bm{x}) \bm{Y}_{\thete}(\bm{x}) p_\thete(\bm{x})  d\bm{x}\\
=&  \int \bm{T}(\bm{x}) \nabla_\thete p_\thete(\bm{x})  d\bm{x}\\
=& \nabla_\thete \int \bm{T}(\bm{x}) p_\thete(\bm{x})  d\bm{x} \quad \text{by the interchangeable condition }\\
=& \nabla_{\thete}\bm{\psi}(\thete).
\end{align*}
Hence using Lemma \ref{lemma:Cauchy-Schwarz}, we have that
\begin{align*}
Var_{\thete}(\bm{T}(\bm{X}))=&\mathbb{E}_{\thete}[(\bm{T}(\bm{X})-\bm{\psi}(\thete)) (\bm{T}(\bm{X})-\bm{\psi}(\thete))^\top]\\
\ge & \mathbb{E}_{\thete}[(\bm{T}(\bm{X})-\bm{\psi}(\thete)) \bm{Y}_{\thete}(\bm{X})^\top] (\mathbb{E}_{\thete}[\bm{Y}_{\thete}(\bm{X}) \bm{Y}_{\thete}(\bm{X})^\top])^{-1} \mathbb{E}_{\thete}[\bm{Y}_{\thete}(\bm{X}) (\bm{T}(\bm{X})-\bm{\psi}(\thete))^\top]\\
=&\nabla_{\thete}\bm{\psi}(\thete) \mathcal{I}_{\thete}^{-1} \nabla_{\thete}\bm{\psi}(\thete)^\top
\end{align*}
as desired.
\Halmos
\endproof

\section{Proofs} \label{sec:proofs}

In this section, we provide the technical proofs of the results in the main paper.

\subsection{Proofs of Results in Section \ref{sec:main}}



\proof{Proof of Theorem \ref{correctlyspecified}}

In the well-specified case, it is easy to see that $\thete^*=\thete_0$ since $\thete_0$ indeed minimizes $\vale_0(\omege_{\thete})$. Therefore 
$\omege^*=\omege_{\thete_0}=\omege_{\thete^*}$. In addition,
$\thete^{KL}=\thete_0$ since $\thete_0$ indeed minimizes $KL(P_\thete,P)$. Therefore 
$\omege^*=\omege_{\thete_0}=\omege_{\thete^{KL}}$. 

For SAA, note that $\hat\omege^{SAA}\xrightarrow{P}\omege^*$ by Proposition \ref{EOconsistency}. Therefore, $R(\hat \omege^{SAA})= \vale_0(\hat \omege^{SAA})-\vale_0(\omege^*) \xrightarrow{P} \vale_0(\omege^*)-\vale_0(\omege^*)=0$ by the continuity of $\vale_0(\omege)$ and the continuous mapping theorem. That is, the regret of SAA is asymptotically 0.

Similarly, for ETO, note that $\hat\thete^{ETO}\xrightarrow{P}\thete^{KL}$ by Proposition \ref{ETOconsistency} where $\thete^{KL}$ is defined in Assumption \ref{ETOconsistency:assm}. Hence, we have that
$R(\hat \omege^{ETO})=R( \omege_{\hat \thete^{ETO}})=\vale_0(\omege_{\hat \thete^{ETO}})-\vale_0(\omege^*)\xrightarrow{P} \vale_0(\omege_{\thete^{KL}})-\vale_0(\omege^*)$ by the continuity of $\vale_0(\omege)$ and $\omege_\thete$.

Similarly, for IEO, note that $\hat\thete^{IEO}\xrightarrow{P}\thete^*$ by Proposition \ref{IEOconsistency} where $\thete^*$ is defined in Assumption \ref{IEOconsistency:assm}. Hence, we have that
$R(\hat \omege^{IEO})=R( \omege_{\hat \thete^{IEO}})=\vale_0(\omege_{\hat \thete^{IEO}})-\vale_0(\omege^*)\xrightarrow{P} \vale_0(\omege_{\thete^*})-\vale_0(\omege^*)$ by the continuity of $\vale_0(\omege)$ and $\omege_\thete$.

Since $\omege^*=\omege_{\thete_0}=\omege_{\thete^*}=\omege_{\thete^{KL}}$, the conclusion of the theorem follows.
\Halmos
\endproof

\proof{Proof of Theorem \ref{SD}}

With the optimality of the solution $\omege^*$ (Assumption \ref{RCforEO:assm}), we have that
\begin{equation}
R(\omege)=\vale_0(\omege)-\vale_0(\omege^*)=\frac{1}{2}(\omege-\omege^*)^\top\nabla_{\omege\omege}\vale_0(\omege^*)(\omege-\omege^*)+ 
o(\|\omege-\omege^*\|_2^2)
\label{excess risk}
\end{equation}
where $\nabla_{\omege\omege}\vale_0(\omege^*)$ is the positive-definite Hessian of $\vale_0$. In particular, this equation holds for $\omege=\hat\omege^{ETO}$, $\hat\omege^{SAA}$ and $\hat\omege^{IEO}$ (with $o$ replaced by $o_P$). Similarly, as $\omege_\thete$ is a twice differentiable function of $\thete$, with the optimality of the solution $\thete_0$ (Assumption \ref{RCforIEO:assm}), we have that
\begin{equation}
R(\omege_{\thete})=\vale_0(\omege_{\thete})-\vale_0(\omege^*)=\frac{1}{2}(\thete-\thete_0)^\top\nabla_{\thete\thete}\vale_0(\omege_{\thete_0})(\thete-\thete_0)+o(\|\thete-\thete_0\|_2^2).\label{excess risk1}
\end{equation}
In particular, this equation holds for $\thete=\hat{\thete}^{ETO}$ and $\hat{\thete}^{IEO}$ (with $o$ replaced by $o_P$). We shall use the asymptotic normality in Proposition \ref{Asymptotic normality for all} and the second-order delta method to obtain the limiting distributions of the asymptotic regret $nR(\hat\omege^{\cdot})$. Before going into the comparison of the three approaches, we point out two facts. 

\textbf{Claim 1.} We have that  
\begin{equation}
\nabla_{\omege\omege}\vale(\omege^*,\thete_0) \nabla_\thete \omege_{\thete_0}+\nabla_{\omege\thete}\vale(\omege^*,\thete_0)=0\label{interim}
\end{equation}
where $\nabla_{\omege\omege}\vale(\omege,\thete)$ denotes the Hessian with respect to $\omege$ only, $\nabla_\thete \omege_{\thete_0}$ is the gradient of $\omege_{\thete}$ at $\thete_0$, and $\nabla_{\omege\thete}\vale(\omege,\thete)$ the matrix containing the second-order cross-derivatives with respect to $\omege$ and $\thete$. To see this, note that 
$\nabla_\omege \vale(\omege_\thete,\thete)= \nabla_\omege \vale(\omege,\thete)|_{\omege=\omege_\thete}=0$ for all $\thete\in\Theta$ as $\omege_\thete$ is a minimum point of \eqref{equ:oracle}. Using the first two points in Assumption \ref{SCforh} and the chain rule, we can take the derivative of $\nabla_\omege \vale(\omege_\thete,\thete)$ with respect to $\thete$ at $\thete_0$ to get 
$$0= \nabla_{\thete} [\nabla_\omege \vale(\omege_\thete,\thete)]|_{\thete=\thete_0}=[\nabla_{\omege\omege}\vale(\omege_\thete,\thete) \nabla_\thete \omege_{\thete}+\nabla_{\omege\thete}\vale(\omege_\thete,\thete)]|_{ \thete=\thete_0}$$
which gives \eqref{interim} by noting that $\omege_{\thete_0}=\omege^*$.

\textbf{Claim 2.} We have that 
\begin{equation} \label{equ:Hessian}
\nabla_{\thete\thete}\vale_0(\omege_{\thete_0})=\nabla_\thete \omege_{\thete_0}^\top \nabla_{\omege\omege}\vale_0(\omege^*)\nabla_\thete \omege_{\thete_0}.    
\end{equation}
Equation \eqref{equ:Hessian} follows since 
\begin{align*}
\nabla_{\thete\thete}\vale_0(\omege_{\thete})|_{\thete=\thete_0}&= \nabla_{\thete} (\nabla_{\omege} \vale_0(\omege_\thete) \nabla_\thete \omege_{\thete})|_{\thete=\thete_0}\nonumber\\
&=\nabla_\thete \omege_{\thete}^\top \nabla_{\omege\omege} \vale_0(\omege_\thete) \nabla_\thete \omege_{\thete}|_{\thete=\thete_0}\nonumber\\
&=\nabla_\thete \omege_{\thete_0}^\top \nabla_{\omege\omege} \vale_0(\omege^*) \nabla_\thete \omege_{\thete_0}   
\end{align*}
where we have used the fact that $\nabla_{\omege} \vale_0(\omege_{\thete})|_{\thete=\thete_0}=\nabla_{\omege} \vale_0(\omege^*)=0$ by the optimality of $\omege^*$. Here the second-order derivative is only taken with respect to $\thete$ in $\omege_\thete$ while the distribution $P_{\thete_0}$ is fixed (see the remark below Assumption \ref{Regularity conditions for all}). 


\textbf{Step 1:} We shall prove that $$\mathbb G^{ETO}\preceq_{st}\mathbb G^{IEO}.$$

To show this, we compare the performance of ETO and IEO at the level of $\thete$ and then leverage Equation \eqref{excess risk1}. 
Note that ETO can be equivalently written as
$$\hat\omege^{ETO} =\min_{\omege\in\Omega} \vale(\omege,\hat{\thete}^{ETO})=\omege_{\hat{\thete}^{ETO}}$$
using the oracle problem \eqref{equ:oracle} by plugging in the MLE estimate $\hat{\thete}^{ETO}$. For ETO, Proposition \ref{RCforETO} gives that  
\begin{equation} \label{equ:MLE}
\sqrt n(\hat{\thete}^{ETO}-\thete_0)\xrightarrow{d} N(0,\mathcal{I}_{\thete_0}^{-1}).
\end{equation}
where
$\mathcal{I}_{\thete_0}^{-1}= (\nabla_{\thete\thete}\mathbb{E}_{P}[\log p_{\thete}(\ran)]|_{\thete=\thete_0})^{-1}$ is the inverse Fisher information.
Plugging \eqref{equ:MLE} into \eqref{excess risk1}, we have
$$n(\vale_0(\hat\omege^{ETO})-\vale_0(\omege^*))\xrightarrow{d}\mathbb G^{ETO}$$
where $\mathbb G^{ETO}=\frac{1}{2}{\mathcal N_1^{ETO}}^\top\nabla_{\thete\thete}\vale_0(\omege_{\thete_0})\mathcal N_1^{ETO}$ and 
$\mathcal N_1^{ETO}\sim N(0,\mathcal{I}_{\thete_0}^{-1})    .$


For IEO, Proposition \ref{RCforIEO} gives that
\begin{equation} \label{Mestimator}
\sqrt{n}(\hat{\thete}^{IEO}-\thete_0)\xrightarrow{d} N(0,\nabla_{\thete\thete}\vale_0(\omege_{\thete_0})^{-1} Var_{P}(\nabla_\thete \cost(\omege_{\thete_0},\ran))\nabla_{\thete\thete}\vale_0(\omege_{\thete_0})^{-1}).    
\end{equation}
Plugging \eqref{Mestimator} into \eqref{excess risk1}, we have
$$n(\vale_0(\hat\omege^{IEO})-\vale_0(\omege^*))\xrightarrow{d}\mathbb G^{IEO}$$
where $\mathbb G^{IEO}=\frac{1}{2}{\mathcal N_1^{IEO}}^\top\nabla_{\thete\thete}\vale_0(\omege_{\thete_0})\mathcal N_1^{IEO}$ and 
$$
\mathcal N_1^{IEO}\sim N(0,\nabla_{\thete\thete}\vale_0(\omege_{\thete_0})^{-1} Var_{P}(\nabla_\thete \cost(\omege_{\thete_0},\ran))\nabla_{\thete\thete}\vale_0(\omege_{\thete_0})^{-1}).   $$

From Lemma \ref{lemma1}, we can complete Step 1 by showing that 
\begin{equation}\label{CR2}
\nabla_{\thete\thete}\vale_0(\omege_{\thete_0})^{-1} Var_{P}(\nabla_\thete \cost(\omege_{\thete_0},\ran))\nabla_{\thete\thete}\vale_0(\omege_{\thete_0})^{-1}\ge \mathcal{I}_{\thete_0}^{-1}.    
\end{equation}
We shall use the multivariate Cramer-Rao bound \citep{cramer1946mathematical,rao1945information,bickel2015mathematical}, which is also stated in Lemma \ref{lemma:Cramer-Rao} in Section \ref{sec:multivariateCramerRao}. 
Recall that $P=P_{\thete_0}$ and $\omege^*=\omege_{\thete_0}$.
Consider a random vector $\ran\sim P_\thete$ and an estimator with the following form $\nabla_\omege \cost(\omege^*,\ran)$, which has the expectation $\mathbb{E}_\thete[\nabla_\omege \cost(\omege^*,\ran)]$. It follows from Assumption \ref{SCforh} that
$\mathbb{E}_\thete[\nabla_\omege \cost(\omege^*,\ran)] = \nabla_\omege \mathbb{E}_\thete [\cost(\omege,\ran)]|_{\omege=\omege^*} = \nabla_\omege \vale(\omege^*, \thete)$. Therefore we have that 
$$\nabla_\thete (\mathbb{E}_\thete[\nabla_\omege \cost(\omege^*,\ran)])^\top |_{\thete=\thete_0} = \nabla_\thete (\nabla_\omege \vale(\omege^*, \thete))^\top |_{\thete=\thete_0}= \nabla_{\omege\thete} \vale(\omege^*, \thete_0).$$
Applying Cramer-Rao bound on the estimator $ \nabla_\omege \cost(\omege^*,\ran)$ (assured by Assumption \ref{SCforh}), we have that
\begin{equation}\label{CR}
Var_{P}\Big(\nabla_\omege \cost(\omege^*,\ran)\Big) \geq  \nabla_{\omege\thete}\vale(\omege^*, \thete_0) \mathcal{I}_{\thete_0}^{-1}\nabla_{\thete\omege}\vale(\omege^*, \thete_0).    
\end{equation}
We can then show that
\begin{align*}
&Var_{P}(\nabla_\thete \cost(\omege_{\thete_0},\ran)) \\
=&Var_{P}(\nabla_\omege \cost(\omege^*,\ran)\nabla_\thete \omege_{\thete_0}) \\
=&\nabla_\thete \omege_{\thete_0}^\top Var_{P}(\nabla_\omege \cost(\omege^*,\ran)) \nabla_\thete \omege_{\thete_0} \quad\text{ since $\nabla_\thete \omege_{\thete_0}$ is deterministic}\\
\geq & \nabla_\thete \omege_{\thete_0}^\top\nabla_{\omege\thete}\vale(\omege^*, \thete_0) \mathcal{I}_{\thete_0}^{-1}\nabla_{\thete\omege}\vale(\omege^*, \thete_0)\nabla_\thete \omege_{\thete_0} \quad\text{ by } \eqref{CR} \\
=  &\nabla_\thete \omege_{\thete_0}^\top\nabla_{\omege\omege}\vale(\omege^*,\thete_0)\nabla_\thete \omege_{\thete_0} \mathcal{I}_{\thete_0}^{-1} \nabla_\thete \omege_{\thete_0}^\top\nabla_{\omege\omege}\vale(\omege^*,\thete_0)\nabla_\thete \omege_{\thete_0}  \quad\text{ by } \eqref{interim} \\
= & \nabla_{\thete\thete}\vale_0(\omege_{\thete_0}) \mathcal{I}_{\thete_0}^{-1} \nabla_{\thete\thete}\vale_0(\omege_{\thete_0}) \quad\text{ by } \eqref{equ:Hessian}.
\end{align*}
Since $\nabla_{\thete\thete}\vale_0(\omege_{\thete_0})^{-1}$ exists due to Assumption \ref{RCforIEO:assm},  then multiplying by the inverse twice gives \eqref{CR2}.




\textbf{Step 2:} We shall prove that $$\mathbb G^{IEO}\preceq_{st}\mathbb G^{SAA}.$$


To show this, we compare the performance of IEO and SAA at the level of $\omege$ and then leverage Equation \eqref{excess risk}. 
For IEO, we reuse the fact \eqref{Mestimator} to obtain, by the Delta method,
\begin{equation}
\sqrt n(\hat\omege^{IEO}-\omege^*)\xrightarrow{d} N(0,\nabla_\thete \omege_{\thete_0}\nabla_{\thete\thete}\vale_0(\omege_{\thete_0})^{-1} Var_{P}(\nabla_\thete \cost(\omege_{\thete_0},\ran))\nabla_{\thete\thete}\vale_0(\omege_{\thete_0})^{-1}\nabla_\thete \omege_{\thete_0}^\top).
\end{equation}
We notice that
$$Var_{P}(\nabla_\thete \cost(\omege_{\thete_0},\ran))=Var_{P}(\nabla_\omege \cost(\omege^*,\ran)\nabla_\thete \omege_{\thete_0})=\nabla_\thete \omege_{\thete_0}^\top Var_{P}(\nabla_\omege \cost(\omege^*,\ran)) \nabla_\thete \omege_{\thete_0}$$
since $\nabla_\thete \omege_{\thete_0}$ is deterministic. Hence, we have
\begin{equation} \label{equ:IEOcovergence}
\sqrt n(\hat\omege^{IEO}-\omege^*)\xrightarrow{d} N(0,\nabla_\thete \omege_{\thete_0}\nabla_{\thete\thete}\vale_0(\omege_{\thete_0})^{-1} \nabla_\thete \omege_{\thete_0}^\top Var_{P}(\nabla_\omege \cost(\omege^*,\ran))\nabla_\thete \omege_{\thete_0} \nabla_{\thete\thete}\vale_0(\omege_{\thete_0})^{-1}\nabla_\thete \omege_{\thete_0}^\top).
\end{equation}
Plugging \eqref{equ:IEOcovergence} into \eqref{excess risk}, we have
$$n(\vale_0(\hat\omege^{IEO})-\vale_0(\omege^*))\xrightarrow{d}\mathbb G^{IEO}$$
where $\mathbb G^{IEO}=\frac{1}{2}{\mathcal N_2^{IEO}}^\top\nabla_{\omege\omege}\vale_0(\omege^*)\mathcal N_2^{IEO}$ (noting that $\mathcal N_2^{IEO}$ is different from $\mathcal N_1^{IEO}$) and 
\begin{equation} \label{equ:IEO3}
\mathcal N_2^{IEO}\sim N(0,\nabla_\thete \omege_{\thete_0}\nabla_{\thete\thete}\vale_0(\omege_{\thete_0})^{-1} \nabla_\thete \omege_{\thete_0}^\top Var_{P}(\nabla_\omege \cost(\omege^*,\ran)) \nabla_\thete \omege_{\thete_0}\nabla_{\thete\thete}\vale_0(\omege_{\thete_0})^{-1}\nabla_\thete \omege_{\thete_0}^\top).    
\end{equation}

Now, for SAA, Proposition \ref{RCforEO} gives that
\begin{equation}
\sqrt n(\hat\omege^{SAA}-\omege^*)\xrightarrow{d} N(0,\nabla_{\omege\omege}\vale_0(\omege^*)^{-1}Var_{P}(\nabla_\omege \cost(\omege^*,\ran))\nabla_{\omege\omege}\vale_0(\omege^*)^{-1}).\label{EO decision}
\end{equation} 
Plugging \eqref{EO decision} into \eqref{excess risk}, we have
$$n(\vale_0(\hat\omege^{SAA})-\vale_0(\omege^*))\xrightarrow{d}\mathbb G^{SAA}$$
where $\mathbb G^{SAA}=\frac{1}{2}{\mathcal N^{SAA}}^\top\nabla_{\omege\omege}\vale_0(\omege^*)\mathcal N^{SAA}$ and 
\begin{equation} \label{equ:EO2}
\mathcal N^{SAA}\sim N(0,\nabla_{\omege\omege}\vale_0(\omege^*)^{-1}Var_{P}(\nabla_\omege \cost(\omege^*,\ran))\nabla_{\omege\omege}\vale_0(\omege^*)^{-1}).    
\end{equation}

Comparing IEO \eqref{equ:IEO3} and SAA \eqref{equ:EO2}, note that the difference in the limiting distributions lies in comparing $\nabla_\thete \omege_{\thete_0}\nabla_{\thete\thete}\vale_0(\omege_{\thete_0})^{-1} \nabla_\thete \omege_{\thete_0}^\top$ to $\nabla_{\omege\omege}\vale_0(\omege^*)^{-1}$. 
Consider two cases:

a) Suppose that $\nabla_\thete \omege_{\thete_0}$ is invertible. Then from \eqref{equ:Hessian} we have 
\begin{align*}
\nabla_\thete \omege_{\thete_0}\nabla_{\thete\thete}\vale_0(\omege_{\thete_0})^{-1} \nabla_\thete \omege_{\thete_0}^\top 
=&\nabla_\thete \omege_{\thete_0} \Big(\nabla_\thete \omege_{\thete_0}^\top \nabla_{\omege\omege} \vale_0(\omege^*) \nabla_\thete \omege_{\thete_0}\Big)^{-1} \nabla_\thete \omege_{\thete_0}^\top \quad \\
=& \nabla_\thete \omege_{\thete_0} (\nabla_\thete \omege_{\thete_0})^{-1} \nabla_{\omege\omege} \vale_0(\omege^*)^{-1} (\nabla_\thete \omege_{\thete_0}^\top)^{-1} \nabla_\thete \omege_{\thete_0}^\top\\
=&\nabla_{\omege\omege}\vale_0(\omege^*)^{-1}    
\end{align*}
and thus
$\mathbb G^{IEO}=_{st}\mathbb G^{SAA}.$

b) Suppose that $\nabla_\thete \omege_{\thete_0}$ is not invertible. 
Then we use Lemma \ref{lemma2} by setting
\begin{itemize}
\item $Q_1= \nabla_{\omege\omege} \vale_0(\omege^*)$, which is invertible by Assumption \ref{RCforEO:assm};
\item $Q_2= Var_{P}(\nabla_\omege \cost(\omege^*,\ran))\ge 0$;
\item $Q_3=\nabla_\thete \omege_{\thete_0}$, where $Q_3^\top Q_1 Q_3=\nabla_\thete \omege_{\thete_0}^\top  \nabla_{\omege\omege} \vale_0(\omege^*) \nabla_\thete \omege_{\thete_0}=\nabla_{\thete\thete}\vale_0(\omege_{\thete_0})$ by \eqref{equ:Hessian} and is positive definite by Assumption \ref{RCforIEO:assm}.
\end{itemize}
Then we obtain from Lemma \ref{lemma2} that
\begin{align*}
&\nabla_\thete \omege_{\thete_0}\nabla_{\thete\thete}\vale_0(\omege_{\thete_0})^{-1} \nabla_\thete \omege_{\thete_0}^\top Var_{P}(\nabla_\omege \cost(\omege^*,\ran))\nabla_\thete \omege_{\thete_0} \nabla_{\thete\thete}\vale_0(\omege_{\thete_0})^{-1}\nabla_\thete \omege_{\thete_0}^\top\\
\le & \nabla_{\omege\omege}\vale_0(\omege^*)^{-1}Var_{P}(\nabla_\omege \cost(\omege^*,\ran))\nabla_{\omege\omege}\vale_0(\omege^*)^{-1}.
\end{align*}

Hence, comparing IEO \eqref{equ:IEO3} and SAA \eqref{equ:EO2}  using Lemma \ref{lemma1}, we conclude that 
$\mathbb G^{IEO}\preceq_{st}\mathbb G^{SAA}.$ 

\Halmos
\endproof

\proof{Proof of Corollary \ref{cor:largervariance}}
In Theorem \ref{SD}, we have shown that $nR(\hat\omege^{IEO})\xrightarrow{d}\mathbb G^{IEO}$, $nR(\hat\omege^{SAA})\xrightarrow{d}\mathbb G^{SAA}$, 
and
$\mathbb G^{IEO}\preceq_{st}\mathbb G^{SAA}$.
In fact, we have obtained a even stronger result in the proof of Theorem \ref{SD} and Lemma \ref{lemma1}, showing that 
$$\mathbb G^{IEO} \overset{d}{=} \bm{Y}_0^\top Q^{IEO}\bm{Y}_0, \quad \mathbb G^{SAA} \overset{d}{=} \bm{Y}_0^\top Q^{SAA} \bm{Y}_0$$
for some positive semi-definite matrices $Q^{IEO}$ and $Q^{SAA}$ where $Q^{IEO} \le Q^{SAA}$. Without loss of generality, we assume $Q^{IEO} = \text{diag}\{\lambda_1, \lambda_2, \cdots, \lambda_p\}$. Otherwise, we can write $Q^{IEO} = Q_1^{\top} \text{diag}\{\lambda_1, \lambda_2, \cdots, \lambda_p\} Q_1$ for some orthonormal matrix $Q_1$ and note that 
$$\bm{Y}_0^\top Q^{IEO}\bm{Y}_0 = (Q_1 \bm{Y}_0)^\top \text{diag}\{\lambda_1, \lambda_2, \cdots, \lambda_p\} (Q_1 \bm{Y}_0) \overset{d}{=} \bm{Y}_0^\top \text{diag}\{\lambda_1, \lambda_2, \cdots, \lambda_p\}  \bm{Y}_0$$
which gives the diagonal representation. 
With $Q^{IEO} = \text{diag}\{\lambda_1, \lambda_2, \cdots, \lambda_p\}$, the variance of $\mathbb G^{IEO}$ can be calculated as
\begin{align*}
& Var(\mathbb G^{IEO}) \\
= &  Var(\bm{Y}_0^\top \text{diag}\{\lambda_1, \lambda_2, \cdots, \lambda_p\}  \bm{Y}_0) \\
= &  Var( \sum_{i=1}^p \lambda_i Y_{0,i}^2) \\
= &  \sum_{i=1}^p \lambda_i^2 Var(Y_{0,i}^2) \\
= &  2 \sum_{i=1}^p \lambda_i^2\\
= &  2 \text{trace}((Q^{IEO})^2).
\end{align*}
Therefore, to show $Var(\mathbb G^{IEO}) \le Var(\mathbb G^{SAA})$, we only need to prove $\text{trace}(Q^{IEO})\le \text{trace}(Q^{SAA})$.

As we have $0 \le Q^{IEO} \le Q^{SAA}$, then for any positive semi-definite matrix $Q_2$, 
$$ Q_2^{\frac{1}{2}} Q^{IEO} Q_2^{\frac{1}{2}} \le  Q_2^{\frac{1}{2}} Q^{SAA} Q_2^{\frac{1}{2}}.$$
Hence, 
$$\text{trace}(Q_2 Q^{IEO}) = \text{trace}(Q_2^{\frac{1}{2}} Q^{IEO} Q_2^{\frac{1}{2}}) \le \text{trace}(Q_2^{\frac{1}{2}} Q^{SAA} Q_2^{\frac{1}{2}}) = \text{trace}(Q_2 Q^{SAA})$$
Particularly, letting $Q_2 = Q^{IEO} + Q^{SAA}$, we derive that
$$\text{trace}((Q^{IEO})^2) + \text{trace}(Q^{SAA} Q^{IEO}) \le \text{trace}(Q^{IEO} Q^{SAA}) + \text{trace}((Q^{SAA})^2).$$
Since the trace of a matrix product is commutative,   $\text{trace}(Q^{SAA} Q^{IEO}) = \text{trace}(Q^{IEO} Q^{SAA})$, we obtain that $\text{trace}((Q^{IEO})^2) \le \text{trace}((Q^{SAA})^2)$, which implies that $Var(\mathbb G^{IEO}) \le Var(\mathbb G^{SAA})$, as desired. 

Using a similar argument, we can also derive that $Var(\mathbb G^{ETO}) \le Var(\mathbb G^{IEO})$.\Halmos
\endproof

\proof{Proof of Lemma \ref{lemma1}}

Since $Q_1$, $Q_2$ are positive semi-definite matrices, their square roots $Q_1^{\frac{1}{2}}$, $Q_2^{\frac{1}{2}}$ exist and are also semi-positive definite.
For $i=1,2$, we note that
$$\bm{Y}_i\sim N(0, Q_i^{\frac{1}{2}} Q_i^{\frac{1}{2}} )\sim Q_i^{\frac{1}{2}} \bm{Y}_0$$
where the random vector $\bm{Y}_0$ follows the standard multivariate Gaussian distribution $N(0,I)$, which implies that
$$\bm{Y}_i^\top Q_3 \bm{Y}_i \stackrel{d}{=} (Q_i^{\frac{1}{2}} \bm{Y}_0)^\top Q_3Q_i^{\frac{1}{2}} \bm{Y}_0= \bm{Y}_0^\top Q_i^{\frac{1}{2}}Q_3Q_i^{\frac{1}{2}}\bm{Y}_0.$$

Next, we aim to show that 
\begin{equation}\label{equ:lemma1claim}
Q_1^{\frac{1}{2}}Q_3Q_1^{\frac{1}{2}}\le Q_2^{\frac{1}{2}}Q_3Q_2^{\frac{1}{2}}.    
\end{equation}
Given \eqref{equ:lemma1claim}, we have that $\bm{u}^\top Q_1^{\frac{1}{2}}Q_3Q_1^{\frac{1}{2}} \bm{u}\le \bm{u}^\top Q_2^{\frac{1}{2}}Q_3Q_2^{\frac{1}{2}} \bm{u}$ for any vector $\bm{u}$. By Lemma \ref{prop1}, this implies that $\bm{Y}_1^\top Q_3 \bm{Y}_1 \preceq_{st} \bm{Y}_2^\top Q_3 \bm{Y}_2.$

We now proceed to prove \eqref{equ:lemma1claim}. First, without loss of generality, we can assume that $Q_3$ is a positive definite matrix. Otherwise, we consider the replacement $\tilde{Q}_3=Q_3+ \gamma I$ ($\gamma>0$) which is a positive definite matrix. If we have 
$$Q_1^{\frac{1}{2}}\tilde{Q}_3Q_1^{\frac{1}{2}}\le Q_2^{\frac{1}{2}}\tilde{Q}_3Q_2^{\frac{1}{2}},$$
then taking the limit $\gamma \to 0$, we obtain that 
$$Q_1^{\frac{1}{2}}Q_3Q_1^{\frac{1}{2}}\le Q_2^{\frac{1}{2}}Q_3Q_2^{\frac{1}{2}}$$
by the continuity of the quadratic form (i.e., $\lim_{\gamma \to 0} \bm{u}^\top Q_1^{\frac{1}{2}}\tilde{Q}_3Q_1^{\frac{1}{2}}\bm{u}=\bm{u}^\top Q_1^{\frac{1}{2}}Q_3Q_1^{\frac{1}{2}}\bm{u}$). Hence we only need to prove \eqref{equ:lemma1claim} for any positive definite matrix $Q_3$. Let $Q_3^{\frac{1}{2}}$ be the square root of $Q_3$ which is also positive definite.

Second, without loss of generality, we can also assume that $Q_1$ and $Q_2$ are both positive definite matrices. Otherwise, we consider the replacement $\tilde{Q}_1=Q_1+ \gamma I$ and $\tilde{Q}_2=Q_2+ \gamma I$ (with the same $\gamma>0$ so that $Q_1\le Q_2$ is equivalent to $\tilde{Q}_1\le \tilde{Q}_2$) which are both positive definite matrices, and assume we have shown that
\begin{equation} \label{equ:Q1Q2}
\tilde{Q}_1^{\frac{1}{2}}Q_3\tilde{Q}_1^{\frac{1}{2}}\le \tilde{Q}_2^{\frac{1}{2}}Q_3\tilde{Q}_2^{\frac{1}{2}}. 
\end{equation}
Since $Q_1$ is positive semi-definite, let 
$Q_4DQ_4^{\top}$ be an eigendecomposition of $Q_1$ where $Q_4$ is an orthogonal matrix, and $D=\text{diag}\{d_1, d_2, \cdots, d_r, 0, \cdots, 0\}$ is a diagonal matrix whose diagonal elements are the eigenvalues of $Q_1$. Then it is easy to see that $$Q_1^{\frac{1}{2}}=Q_4D^{\frac{1}{2}}Q_4^{\top}=Q_4\text{diag}\{d_1^{\frac{1}{2}}, d_2^{\frac{1}{2}}, \cdots, d_r^{\frac{1}{2}}, 0, \cdots, 0\}Q_4^{\top}$$ 
and 
$$\tilde{Q}_1^{\frac{1}{2}}=Q_4(D+\gamma I)^{\frac{1}{2}}Q_4^{\top}=Q_4\text{diag}\{(d_1+\gamma)^{\frac{1}{2}}, (d_2+\gamma)^{\frac{1}{2}}, \cdots, (d_r+\gamma)^{\frac{1}{2}}, \gamma^{\frac{1}{2}}, \cdots, \gamma^{\frac{1}{2}}\}Q_4^{\top}.$$
Since 
$(d_j+\gamma)^{\frac{1}{2}}-d_j^{\frac{1}{2}}\le \gamma^{\frac{1}{2}}$ and $\gamma^{\frac{1}{2}}-0\le \gamma^{\frac{1}{2}}$, we have 
$$\|\tilde{Q}_1^{\frac{1}{2}}-Q_1^{\frac{1}{2}}\|_{op}= \|(D+\gamma I)^{\frac{1}{2}}-D^{\frac{1}{2}}\|_{op}\le \gamma^{\frac{1}{2}}.$$
Hence for any vector $\bm{u}$, we have that 
\begin{align*}
&\sqrt{\bm{u}^\top \tilde{Q}_1^{\frac{1}{2}}Q_3\tilde{Q}_1^{\frac{1}{2}}\bm{u}} - \sqrt{\bm{u}^\top Q_1^{\frac{1}{2}}Q_3Q_1^{\frac{1}{2}}\bm{u}}   \\
=& \|Q_3^{\frac{1}{2}}\tilde{Q}_1^{\frac{1}{2}}\bm{u}\|_{2}- \|Q_3^{\frac{1}{2}}Q_1^{\frac{1}{2}}\bm{u}\|_{2}\\
\le& \|Q_3^{\frac{1}{2}}(\tilde{Q}_1^{\frac{1}{2}}-Q_1^{\frac{1}{2}})\bm{u}\|_{2}\\
\le& \|Q_3^{\frac{1}{2}}\|_{op} \|\tilde{Q}_1^{\frac{1}{2}}-Q_1^{\frac{1}{2}}\|_{op} \|\bm{u}\|_2\\
\le& \gamma^{\frac{1}{2}} \|Q_3^{\frac{1}{2}}\|_{op} \|\bm{u}\|_2
\end{align*}
which implies that
$$\lim_{\gamma \to 0} \bm{u}^\top \tilde{Q}_1^{\frac{1}{2}}Q_3\tilde{Q}_1^{\frac{1}{2}}\bm{u} = \bm{u}^\top Q_1^{\frac{1}{2}}Q_3Q_1^{\frac{1}{2}}\bm{u}$$
Similarly, we have
$$\lim_{\gamma \to 0} \bm{u}^\top \tilde{Q}_2^{\frac{1}{2}}Q_3\tilde{Q}_2^{\frac{1}{2}}\bm{u} = \bm{u}^\top Q_2^{\frac{1}{2}}Q_3Q_2^{\frac{1}{2}}\bm{u}$$
Therefore, taking the limit $\gamma \to 0$ on both sides of \eqref{equ:Q1Q2}, we obtain that 
$$Q_1^{\frac{1}{2}}Q_3Q_1^{\frac{1}{2}}\le Q_2^{\frac{1}{2}}Q_3Q_2^{\frac{1}{2}}.$$
Hence we only need to prove \eqref{equ:lemma1claim} for any positive definite matrices $Q_1$ and $Q_2$ (in which case $Q_1^{-\frac{1}{2}}$ and $Q_2^{-\frac{1}{2}}$ exist).

To show \eqref{equ:lemma1claim}, we note that 
$Q_1\le Q_2$ implies that $Q_2^{-\frac{1}{2}}Q_1^{\frac{1}{2}}Q_1^{\frac{1}{2}}Q_2^{-\frac{1}{2}}\le I$ so we have 
$$\|Q_2^{-\frac{1}{2}}Q_1^{\frac{1}{2}}Q_1^{\frac{1}{2}}Q_2^{-\frac{1}{2}}\|_{op}\le 1$$ 
where $\|\cdot\|_{op}$ is the operator norm of the matrix and thus $\|Q_2^{-\frac{1}{2}}Q_1^{\frac{1}{2}}\|^2_{op}= \|Q_2^{-\frac{1}{2}}Q_1^{\frac{1}{2}}(Q_2^{-\frac{1}{2}}Q_1^{\frac{1}{2}})^\top\|_{op}\le 1$. This shows that all eigenvalues of $Q_2^{-\frac{1}{2}}Q_1^{\frac{1}{2}}$ are less than $1$. Since $Q_3^{-\frac{1}{2}}Q_2^{-\frac{1}{2}}Q_1^{\frac{1}{2}}Q_3^{\frac{1}{2}}$ is similar to $Q_2^{-\frac{1}{2}}Q_1^{\frac{1}{2}}$, all the eigenvalues of $Q_3^{-\frac{1}{2}}Q_2^{-\frac{1}{2}}Q_1^{\frac{1}{2}}Q_3^{\frac{1}{2}}$ are the same as $Q_2^{-\frac{1}{2}}Q_1^{\frac{1}{2}}$ (all less than $1$), which implies that
$$\|Q_3^{-\frac{1}{2}}Q_2^{-\frac{1}{2}}Q_1^{\frac{1}{2}}Q_3^{\frac{1}{2}}\|_{op}\le 1.$$ 
Taking the transpose, we also have
$$\|Q_3^{\frac{1}{2}}Q_1^{\frac{1}{2}}Q_2^{-\frac{1}{2}}Q_3^{-\frac{1}{2}}\|_{op}\le 1.$$ 
Hence we have
$$\|Q_3^{-\frac{1}{2}}Q_2^{-\frac{1}{2}}Q_1^{\frac{1}{2}}Q_3^{\frac{1}{2}}Q_3^{\frac{1}{2}}Q_1^{\frac{1}{2}}Q_2^{-\frac{1}{2}}Q_3^{-\frac{1}{2}}\|_{op}\le \|Q_3^{-\frac{1}{2}}Q_2^{-\frac{1}{2}}Q_1^{\frac{1}{2}}Q_3^{\frac{1}{2}}\|_{op}\|Q_3^{\frac{1}{2}}Q_1^{\frac{1}{2}}Q_2^{-\frac{1}{2}}Q_3^{-\frac{1}{2}}\|_{op}\le 1$$ 
which implies that
$$Q_3^{-\frac{1}{2}}Q_2^{-\frac{1}{2}}Q_1^{\frac{1}{2}}Q_3^{\frac{1}{2}}Q_3^{\frac{1}{2}}Q_1^{\frac{1}{2}}Q_2^{-\frac{1}{2}}Q_3^{-\frac{1}{2}}\le I$$
and thus proves \eqref{equ:lemma1claim}. \Halmos

\endproof

\proof{Proof of Lemma \ref{lemma2}}
Without loss of generality, we can assume that $Q_2\in \mathcal{R}^{p\times p}$ is a positive definite matrix. Otherwise, we consider the replacement $\tilde{Q}_2= Q_2+ \gamma I$ ($\gamma>0$) which is a positive definite matrix. If we have 
$$Q_3(Q_3^\top Q_1 Q_3 + \lambda I_q)^{-1}Q_3^\top \tilde{Q}_2 Q_3(Q_3^\top Q_1 Q_3 + \lambda I_q)^{-1}Q_3^\top\le Q_1^{-1} \tilde{Q}_2 Q_1^{-1},$$
then taking the limit $\gamma \to 0$, we obtain that 
$$Q_3(Q_3^\top Q_1 Q_3 + \lambda I_q)^{-1}Q_3^\top Q_2 Q_3(Q_3^\top Q_1 Q_3 + \lambda I_q)^{-1}Q_3^\top\le Q_1^{-1} Q_2 Q_1^{-1}$$
by the continuity of the quadratic form. Hence we only need to prove the result for any positive definite matrix $Q_2$.

Since $Q_2$ is a positive definite matrix, its square root $Q_2^{\frac{1}{2}}$ exists and is also positive definite.
Note that the desired result is equivalent to 
\begin{equation} \label{equ:lemma2}
Q^{-\frac{1}{2}}_2 Q_1 Q_3(Q_3^\top Q_1 Q_3 + \lambda I_q)^{-1}Q_3^\top Q_2 Q_3(Q_3^\top Q_1 Q_3+ \lambda I_q)^{-1}Q_3^\top Q_1 Q^{-\frac{1}{2}}_2\le I   
\end{equation}
as both $Q^{\frac{1}{2}}_2$ and $Q_1$ are invertible.
Next, we claim that
$$\|Q^{-\frac{1}{2}}_2 Q_1 Q_3(Q_3^\top Q_1 Q_3 + \lambda I_q)^{-1}Q_3^\top Q_2^{\frac{1}{2}}\|_{op}\le 1.$$
In fact, we first notice that 
$$\|(Q_3^\top Q_1 Q_3 + \lambda I_q)^{-1}Q_3^\top  Q_1 Q_3\|_{op}\le 1,$$ 
which follows from the classical result in operator theory:
$$\|(Q_3^\top Q_1 Q_3 + \lambda I_q)^{-1}Q_3^\top  Q_1 Q_3\|_{op}\le \|(x+\lambda )^{-1} x\|_\infty=1$$
since $Q_3^\top Q_1 Q_3$ is a positive semi-definite operator on the Hilbert space $L^2(\mathbb{R}^q)$ \citep{kadison1986fundamentals}.
It is known that $Q_1 Q_3(Q_3^\top Q_1 Q_3 + \lambda I_q)^{-1}Q_3^\top$ and $(Q_3^\top Q_1 Q_3 + \lambda I_q)^{-1}Q_3^\top Q_1 Q_3$ (by changing the order of the matrix multiplication) have the same set of eigenvalues except $0$ so we also have
$$\|Q_1 Q_3 (Q_3^\top Q_1 Q_3 + \lambda I_q)^{-1}Q_3^\top \|_{op}\le 1.$$
Note that $Q_1 Q_3 (Q_3^\top Q_1 Q_3 + \lambda I_q)^{-1}Q_3^\top $ and 
$Q^{-\frac{1}{2}}_2 Q_1 Q_3(Q_3^\top Q_1 Q_3 + \lambda I_q)^{-1}Q_3^\top Q_2^{\frac{1}{2}}$ are similar so all the eigenvalues of $Q^{-\frac{1}{2}}_2 Q_1 Q_3(Q_3^\top Q_1 Q_3 + \lambda I_q)^{-1}Q_3^\top Q_2^{\frac{1}{2}}$ are the same as $Q_1 Q_3(Q_3^\top Q_1 Q_3 + \lambda I_q)^{-1}Q_3^\top$ (all less than $1$), which implies that
$$\|Q^{-\frac{1}{2}}_2 Q_1 Q_3(Q_3^\top Q_1 Q_3 + \lambda I_q)^{-1}Q_3^\top Q_2^{\frac{1}{2}}\|_{op}\le 1.$$  
Similarly, we have
$$\|Q^{\frac{1}{2}}_2 Q_3(Q_3^\top Q_1 Q_3 + \lambda I_q)^{-1}Q_3^\top Q_1 Q_2^{-\frac{1}{2}}\|_{op}\le 1.$$ 
Therefore we obtain that
\begin{align*}
&\|Q^{-\frac{1}{2}}_2 Q_1 Q_3(Q_3^\top Q_1 Q_3 + \lambda I_q)^{-1}Q_3^\top Q_2 Q_3(Q_3^\top Q_1 Q_3+ \lambda I_q)^{-1}Q_3^\top Q_1 Q^{-\frac{1}{2}}_2\|_{op}\\
\le &\|Q^{-\frac{1}{2}}_2 Q_1 Q_3(Q_3^\top Q_1 Q_3 + \lambda I_q)^{-1}Q_3^\top Q_2^{\frac{1}{2}}\|_{op} \|Q^{\frac{1}{2}}_2 Q_3(Q_3^\top Q_1 Q_3 + \lambda I_q)^{-1}Q_3^\top Q_1 Q_2^{-\frac{1}{2}}\|_{op}\\
\le &1
\end{align*} 
which implies that
$$Q^{-\frac{1}{2}}_2 Q_1 Q_3(Q_3^\top Q_1 Q_3 + \lambda I_q)^{-1}Q_3^\top Q_2 Q_3(Q_3^\top Q_1 Q_3+ \lambda I_q)^{-1}Q_3^\top Q_1 Q^{-\frac{1}{2}}_2\le I.$$
Hence we conclude that
$$Q_3(Q_3^\top Q_1 Q_3 + \lambda I_q)^{-1}Q_3^\top Q_2 Q_3(Q_3^\top Q_1 Q_3+ \lambda I_q)^{-1}Q_3^\top\le Q_1^{-1} Q_2 Q_1^{-1}.$$
\Halmos
\endproof

\proof{Proof of Proposition \ref{prop: Verifying Assumptions}}
It is sufficient to show the result for $p=1$ since the cost function is an independent summation of the cost for each product. 

\textbf{Step 1.} We first establish the following equality: For any $\theteun \in \mathbb{R}$, 
$$\nabla_{\theteun} \int_{-\infty}^{\omegeun} p_{\theteun}(s) ds = \int_{-\infty}^{\omegeun} \nabla_{\theteun} p_{\theteun}(s) ds$$
where $p_{\theteun}(s)$ is the pdf of $N(t_1\theteun,\sigma_1)$, the demand distribution. 

As this is a point-wise equality, it is sufficient to prove it in an interval $\theteun \in [\theteun_1 - \varepsilon, \theteun_1 + \varepsilon]$ for any fixed $\theteun_1$ and some small $\varepsilon>0$. For this, it is easy to see that there exists a constant $C_0$ such that
$|\nabla_{\theteun} p_{\theteun}(s)| \le C_0 (|\nabla_{\theteun} p_{\theteun_1}(s)|+|p_{\theteun_1}(s)|)$ for any $\theteun \in [\theteun_1 - \varepsilon, \theteun_1 + \varepsilon]$ where $C_0$ may depend on $\varepsilon$. Since 
$$\int_{-\infty}^{\omegeun} C_0 (|\nabla_{\theteun} p_{\theteun_1}(s)|+|p_{\theteun_1}(s)|) ds < +\infty,$$ 
the dominated convergence theorem implies that $$\nabla_{\theteun} \int_{-\infty}^{\omegeun} p_{\theteun_1}(s) ds = \int_{-\infty}^{\omegeun} \nabla_{\theteun} p_{\theteun_1}(s) ds$$
for any $\theteun_1 \in \mathbb{R}$. A similar argument also gives
$$\nabla_{\theteun\theteun} \int_{-\infty}^{\omegeun} p_{\theteun_1}(s) ds = \int_{-\infty}^{\omegeun} \nabla_{\theteun\theteun} p_{\theteun_1}(s) ds.$$

Based on the above fact, we can easily compute the derivative of the expected cost as follows. Note that the expected cost function can be written as 
\begin{align*}
\vale(\omegeun, \theteun) =  &\E_{P_{\theteun}}\left[ \cost(\omegeun,\ranun)\right]   \\
=&\E_{P_{\theteun}}\left[ h(\omegeun-\ranun)^+ + b(\ranun-\omegeun)^+\right]   \\
=& (h+b)\omegeun\int_{-\infty}^{\omegeun} p_{\theteun}(s) ds-(h+b)\int_{-\infty}^{\omegeun} sp_{\theteun}(s) ds-b\omegeun+b\E_{P_{\theteun}}[\ranun]\\
=& (h+b)\omegeun\int_{-\infty}^{\omegeun} p_{\theteun}(s) ds-(h+b)\int_{-\infty}^{\omegeun} sp_{\theteun}(s) ds-b\omegeun+bt_1\theteun
\end{align*}
Hence,
$$\nabla_{\omegeun} \vale(\omegeun, \theteun) = (h+b)\int_{-\infty}^{\omegeun} p_{\theteun}(s) ds -b$$
$$\nabla_{\theteun} \vale(\omegeun, \theteun) = (h+b)\omegeun\int_{-\infty}^{\omegeun} \nabla_{\theteun} p_{\theteun}(s) ds-(h+b)\int_{-\infty}^{\omegeun} s \nabla_{\theteun}p_{\theteun}(s) ds + bt_1$$
$$\nabla_{\omegeun\omegeun} \vale(\omegeun, \theteun) = (h+b)p_{\theteun}(\omegeun) >0$$
$$\nabla_{\omegeun\theteun} \vale(\omegeun, \theteun) = (h+b)\int_{-\infty}^{\omegeun} \nabla_{\theteun} p_{\theteun}(s) ds -b$$
$$\nabla_{\theteun\theteun} \vale(\omegeun, \theteun) = (h+b)\omegeun\int_{-\infty}^{\omegeun} \nabla_{\theteun\theteun} p_{\theteun}(s) ds-(h+b)\int_{-\infty}^{\omegeun} s \nabla_{\theteun\theteun}p_{\theteun}(s) ds.$$

\textbf{Step 2.} We show Assumption \ref{SCforh} holds. 

Based on the above derivatives that we obtained, it is clear that 
$\vale(\omegeun, \theteun)$ is twice continuously differentiable with respect to $(\omegeun, \theteun)$. So Assumption \ref{SCforh} Point 1) is satisfied.

Recall that $\nabla_{\omegeun} \vale(\omegeun, \theteun) = (h+b)\int_{-\infty}^{\omegeun} p_{\theteun}(s) ds -b$ with a positive second derivative. Therefore, the best decision for the oracle problem is 
$$\omegeun_\theteun=t_1\theteun+\sigma_1 \Phi_{normal}^{-1}\left(\frac{b}{b+h}\right) = t_1\theteun +\tilde{\sigma}_1$$ 
where $\tilde{\sigma}_1$ is a constant (See also \citet{turken2012multi}). Therefore
\begin{equation}\label{equ:w_theta_equation}
\nabla_{\theteun} \omegeun_\theteun = t_1, \ \nabla_{\theteun\theteun} \omegeun_\theteun = 0.    
\end{equation}
This shows that the optimal solution $\omegeun_{\theteun}$ to the oracle problem \eqref{equ:oracle} is twice differentiable with respect to $\theteun$. So Assumption \ref{SCforh} Point 2) is satisfied.

Note that $\nabla_{\omegeun} \cost(\omegeun,\ranun) = h$ if $\omegeun>\ranun$ or $-b$ if $\omegeun<\ranun$. So
\begin{equation}\label{equ:exchange1}
\int \nabla_{\omegeun} \cost(\omegeun,\ranun) p_\theteun(\ranun)  d\ranun = h \int_{-\infty}^{\omegeun} p_{\theteun}(s) ds - b \int_{\omegeun}^{+\infty} p_{\theteun}(s) = (h+b)\int_{-\infty}^{\omegeun} p_{\theteun}(s) ds -b.    
\end{equation}
$$\nabla_\theteun \int \nabla_{\omegeun} \cost(\omegeun,\ranun)   p_\theteun(\ranun)  d\ranun = (h+b)\int_{-\infty}^{\omegeun} \nabla_\theteun p_{\theteun}(s) ds.$$
On the other hand, since $p_\theteun(\ranun)$ is a Gaussian distribution, a simple calculation (or using the score function) gives rise to
$\int \nabla_\theteun p_\theteun(\ranun)  d\ranun =0$ and thus 
$$\int \nabla_{\omegeun} \cost(\omegeun,\ranun) \nabla_\theteun p_\theteun(\ranun)  d\ranun = h \int_{-\infty}^{\omegeun} \nabla_\theteun p_{\theteun}(s) ds - b \int_{\omegeun}^{+\infty} \nabla_\theteun p_{\theteun}(s) ds = (h+b)\int_{-\infty}^{\omegeun} \nabla_\theteun p_{\theteun}(s) ds.$$
This shows that
$$ 
\nabla_\theteun \int \nabla_{\omegeun} \cost(\omegeun,\ranun)   p_\theteun(\ranun)  d\ranun =  \int \nabla_{\omegeun} \cost(\omegeun,\ranun) \nabla_\theteun p_\theteun(\ranun)  d\ranun.$$
In addition, note that
$$\nabla_\omegeun \int  \cost(\omegeun,\ranun) p_\theteun(\ranun) d\ranun = \nabla_{\omegeun} \vale(\omegeun, \theteun) = (h+b)\int_{-\infty}^{\omegeun} p_{\theteun}(s) ds -b$$
as computed above. Comparing this with \eqref{equ:exchange1}, we obtain that
$$\int \nabla_{\omegeun} \cost(\omegeun,\ranun) p_\theteun(\ranun)  d\ranun = \nabla_\omegeun \int  \cost(\omegeun,\ranun) p_\theteun(\ranun) d\ranun.$$
So Assumption \ref{SCforh} Point 3) is satisfied.

\textbf{Step 3.} We show Assumptions \ref{IEOconsistency} and \ref{RCforIEO} hold for IEO. 


First, via the chain rule, we have
\begin{align*}
\nabla_{\theteun\theteun} \vale_0(\omegeun_{\theteun}) &= (\nabla_{\theteun} \omegeun_\theteun)^2 \nabla_{\omegeun\omegeun}\vale_0(\omegeun_{\theteun}) + \nabla_{\theteun\theteun} \omegeun_\theteun \nabla_{\omegeun}\vale_0(\omegeun_{\theteun})\\
&= (\nabla_{\theteun} \omegeun_\theteun)^2 \nabla_{\omegeun\omegeun}\vale_0(\omegeun_{\theteun})\\
&= t_1^2(h+b)p_{\theteun_0}(\omegeun_{\theteun}) >0
\end{align*}
where we use the facts \eqref{equ:w_theta_equation}.

By Taylor's theorem and using $\nabla_{\theteun} \vale_0(\omegeun_{\theteun^*})=0$, we have that
$$\vale_0(\omegeun_\theteun) = \vale_0(\omegeun_{\theteun^*}) + \frac{1}{2} \nabla_{\theteun\theteun} \vale_0(\omegeun_{\tilde{\theteun}}) (\theteun-\theteun^*)^2$$
for some $\tilde{\theteun}$ between $\theteun$ and $\theteun^*$. 
Note that $\nabla_{\theteun\theteun} 
\vale_0(\omegeun_{\tilde{\theteun}})>0$, so the above equation implies that for every $\epsilon> 0$,
$\inf_{\theteun\in \Theta: d(\theteun,\theteun^*)\ge \epsilon} \vale_0(\omegeun_\theteun) > \vale_0(\omegeun_{\theteun^*})$. Hence Assumption \ref{IEOconsistency} Part 2 holds. Since $\hat{\Theta}$ is a compact set and $\hat{\theteun}^{IEO} \in \hat{\Theta}$, it is sufficient to verify Assumption \ref{IEOconsistency} Part 1 on the compact set $\hat{\Theta}$. To this end, the uniform law of large numbers directly implies that Assumption \ref{IEOconsistency} Part 1 holds.

The above discussion also shows that the map $\theteun\mapsto \vale_0(\omegeun_{\theteun})$ admits a second-order Taylor expansion at the point of minimum $\theteun^*$ with nonsingular second derivative $\nabla_{\theteun\theteun}\vale_0(\omegeun_{\theteun^*})>0$. (Assumption \ref{RCforIEO:assm}.)

Note that $\omegeun_\theteun=t_1\theteun +\tilde{\sigma}_1$, so we have
$$\cost(\omegeun_\theteun,\ranun) = h(t_1\theteun +\tilde{\sigma}_1-\ranun)^+ + b(\ranun-t_1\theteun -\tilde{\sigma}_1)^+.$$ 
Hence,
$\nabla_{\theteun} \cost(\omegeun_\theteun,\ranun) = ht_1$ if $t_1\theteun +\tilde{\sigma}_1>\ranun$ or $-bt_1$ if $t_1\theteun +\tilde{\sigma}_1<\ranun$. So $\cost(\omegeun_\theteun,\ranun)$ is a measurable function of $\ranun$ such that $\theteun\mapsto \cost(\omegeun_\theteun,\ranun)$ is differentiable at $\theteun^*$ for almost every $\ranun$, actually for all $\ranun\ne t_1\theteun^* +\tilde{\sigma}_1$. (Assumption \ref{RCforIEO:assm}.)

Moreover, since $|\nabla_{\theteun} \cost(\omegeun_\theteun,\ranun)| \le \max(|ht_1|, |bt_1|)$, we have that for any $\theteun_1$ and $\theteun_2$ in a neighborhood of $\theteun^*$, there exists a measurable function $K(\ranun) := \max(|ht_1|, |bt_1|)$ with $\mathbb{E}_{P}[K(\ranun)]<\infty$ such that
$|\cost(\omegeun_{\theteun_1},\ranun)-\cost(\omegeun_{\theteun_2},\ranun)|\le K(\ranun) \|\theteun_1-\theteun_2\|$. (Assumption \ref{RCforIEO:assm}.)

Note that in our experiments, solutions can be obtained precisely, so the assumption on the approximate error $0 = o_{P}(n^{-1})$ in Assumptions \ref{IEOconsistency} and \ref{RCforIEO} holds. 

From the discussions above, we conclude that Assumptions \ref{IEOconsistency} and \ref{RCforIEO} hold for IEO.


\textbf{Step 4.} We show Assumptions \ref{ETOconsistency} and \ref{RCforETO:assm} hold for ETO.  

First, we have
\begin{align*}
\mathbb{E}_{P}[\log p_\theteun(\ranun)] &= \mathbb{E}_{P}\left[-\log \sqrt{2\pi \sigma^2_1} - \frac{1}{2\sigma^2_1} (\ranun-t_1\theteun)^2\right]\\
&= C_0 - \frac{1}{2\sigma^2_1}\mathbb{E}_{P}\left[(\ranun-t_1\theteun)^2\right]\\
&= C_0 - \frac{1}{2\sigma^2_1} (\sigma^2_1+(\theteun-\theteun_0)^2t_1^2)\\
&= C_0 - \frac{t_1^2}{2\sigma^2_1} (\theteun-\theteun_0)^2
\end{align*}
where we use the fact that $P= N(t_1\theteun_0,\sigma_1)$. We have
$$\nabla_{\theteun} \mathbb{E}_{P}[\log p_\theteun(\ranun)] = - \frac{t_1^2}{\sigma^2_1} (\theteun-\theteun_0), \ \nabla_{\theteun\theteun} \mathbb{E}_{P}[\log p_\theteun(\ranun)] = - \frac{t_1^2}{\sigma^2_1}<0.$$
Therefore, the MLE estimator is $\theteun^{KL}=\theteun_0$.

By the formula of $\mathbb{E}_{P}[\log p_\theteun(\ranun)]$, we have that
$$\mathbb{E}_{P}[\log p_\theteun(\ranun)] = \mathbb{E}_{P}[\log p_{\theteun^{KL}}(\ranun)] - \frac{t_1^2}{\sigma^2_1} (\theteun-\theteun^{KL})^2.$$
Note that $\frac{t_1^2}{\sigma^2_1}>0$, so the above equation implies that for every $\epsilon> 0$, $\sup_{\theteun\in \Theta: d(\theteun,\theteun^{KL})\ge \epsilon} \mathbb{E}_{P} [ \log p_\theteun(\ranun)] < \mathbb{E}_{P} [ \log p_{\theteun^{KL}}(\ranun)]$. Hence Assumption \ref{ETOconsistency} Part 2 holds. Since $\hat{\Theta}$ is a compact set and $\hat{\theteun}^{ETO} \in \hat{\Theta}$, it is sufficient to verify Assumption \ref{ETOconsistency} Part 1 on the compact set $\hat{\Theta}$. To this end, the uniform law of large numbers directly implies that Assumption \ref{ETOconsistency} Part 1 holds.

(Assumption \ref{ETOconsistency} Part 2.) In practice, $\Theta$ can be set to a compact set and thus the uniform law of large numbers implies that Assumption \ref{ETOconsistency} Part 1 holds.

The above discussion also shows that the map $\theteun\mapsto \mathbb{E}_{P}[\log p_\theteun(\ranun)]$ admits a second-order Taylor expansion at the point of maximum $\theteun^{KL}$ with nonsingular second derivative $\nabla_{\theteun\theteun} \mathbb{E}_{P}[\log p_\theteun(\ranun)]<0$. (Assumption \ref{RCforETO:assm}.)

Note that we have
$$\log p_\theteun(\ranun) = -\log \sqrt{2\pi \sigma^2_1} - \frac{1}{2\sigma^2_1} (\ranun-t_1\theteun)^2.$$ 
Hence,
$\nabla_{\theteun} \log p_\theteun(\ranun) = - \frac{t_1}{\sigma^2_1} (t_1\theteun-\ranun)$. So $\log p_\theteun(\ranun)$ is a measurable function of $\ranun$ such that $\theteun\mapsto \log p_\theteun(\ranun)$ is differentiable at $\theteun^{KL}$ for all $\ranun$. (Assumption \ref{RCforETO:assm}.)

Moreover, we have for any $\theteun$ such that $|\theteun-\theteun^*|<\varepsilon_1$ (in a neighborhood of $\theteun^*$),  
$$|\nabla_{\theteun}\log p_\theteun(\ranun)| = \left|\frac{t_1}{\sigma^2_1} (t_1\theteun-\ranun)\right| \le \left|\frac{t_1}{\sigma^2_1} (t_1\theteun^{KL}-\ranun)\right| + \left|\frac{t_1}{\sigma^2_1} t_1\varepsilon_1\right|.$$
Hence, we have that for any $\theteun_1$ and $\theteun_2$ in a neighborhood of $\theteun^*$, there exists a measurable function $K(\ranun) := \left|\frac{t_1}{\sigma^2_1} (t_1\theteun^{KL}-\ranun)\right| + \left|\frac{t_1}{\sigma^2_1} t_1\varepsilon_1\right|$ with $\mathbb{E}_{P}[K(\ranun)]<\infty$ such that
$|\log p_{\theteun_1}(\ranun)-\log p_{\theteun_2}(\ranun)|\le K(\ranun) \|\theteun_1-\theteun_2\|$. (Assumption \ref{RCforETO:assm}.)

Note that in our experiments, solutions can be obtained precisely, so the assumption on the approximate error $0 = o_{P}(n^{-1})$ in Assumptions \ref{ETOconsistency} and \ref{RCforETO} holds. 

From the discussions above, we conclude that Assumptions \ref{ETOconsistency} and \ref{RCforETO} hold for ETO.

\textbf{Step 5.} We show Assumptions \ref{EOconsistency} and \ref{RCforEO:assm} hold for SAA. 


First, we recall that
\begin{align*}
\nabla_{\omegeun\omegeun} \vale_0(\omegeun^*)
= (h+b)p_{\theteun_0}(\omegeun^*) >0
\end{align*}

By Taylor's theorem and using $\nabla_{\omegeun} \vale_0(\omegeun^*)=0$, we have that
$$\vale_0(\omegeun) = \vale_0(\omegeun^*) + \frac{1}{2} \nabla_{\omegeun\omegeun} \vale_0(\tilde{\omegeun}) (\omegeun-\omegeun^*)^2$$
for some $\tilde{\omegeun}$ between $\omegeun$ and $\omegeun^*$. 
Note that $\nabla_{\omegeun\omegeun} \vale_0(\omegeun^*)>0$, so the above equation implies that for every $\epsilon> 0$,
$\inf_{\omegeun\in \Omega: d(\omegeun,\omegeun^*)\ge \epsilon} \vale_0(\omegeun) > \vale_0(\omegeun^*)$. Hence Assumption \ref{EOconsistency} Part 2 holds. Since $\hat{\Omega}$ is a compact set and $\hat{\omegeun}^{SAA} \in \hat{\Omega}$, it is sufficient to verify Assumption \ref{EOconsistency} Part 1 on the compact set $\hat{\Omega}$. To this end, the uniform law of large numbers directly implies that Assumption \ref{EOconsistency} Part 1 holds.

The above discussion also shows that the map $\theteun\mapsto \vale_0(\omegeun)$ admits a second-order Taylor expansion at the point of minimum $\omegeun^*$ with nonsingular second derivative $\nabla_{\omegeun\omegeun} \vale_0(\omegeun^*)>0$ (Assumption \ref{RCforEO:assm}).

Note that
$$\cost(\omegeun,\ranun) = h(\omegeun -\ranun)^+ + b(\ranun-\omegeun)^+.$$ 
Hence,
$\nabla_{\omegeun} \cost(\omegeun,\ranun) = h$ if $\omegeun>\ranun$ or $-b$ if $\omegeun<\ranun$. So $\cost(\omegeun,\ranun)$ is a measurable function of $\ranun$ such that $\omegeun\mapsto \cost(\omegeun,\ranun)$ is differentiable at $\omegeun^*$ for almost every $\ranun$, actually for all $\omegeun\ne \omegeun^*$ (Assumption \ref{RCforEO:assm}).

Moreover, since $|\nabla_{\omegeun} \cost(\omegeun,\ranun)| \le \max(|h|, |b|)$, we have that for any $\omegeun_1$ and $\omegeun_2$ in a neighborhood of $\omegeun^*$, there exists a measurable function $K(\ranun) := \max(|h|, |b|)$ with $\mathbb{E}_{P}[K(\ranun)]<\infty$ such that
$|\cost(\omegeun_1,\ranun)-\cost(\omegeun_2,\ranun)|\le K(\ranun) \|\omegeun_1-\omegeun_2\|$ (Assumption \ref{RCforEO:assm}).

Note that in our experiments, solutions can be obtained precisely, so the assumption on the approximate error $0 = o_{P}(n^{-1})$ in Assumptions \ref{EOconsistency} and \ref{RCforEO} holds. 

From the discussions above, we conclude that Assumptions \ref{EOconsistency} and \ref{RCforEO:assm} hold for SAA. 


In conclusion, based on the discussions above, we establish that Assumptions \ref{consistency all} (including Assumptions \ref{EOconsistency:assm}, \ref{ETOconsistency:assm}, \ref{IEOconsistency:assm}), \ref{Regularity conditions for all} (including Assumptions \ref{RCforEO:assm}, \ref{RCforETO:assm}, \ref{RCforIEO:assm}), and \ref{SCforh} hold. Consequently, the result in Theorem \ref{SD} follows.

\Halmos
\endproof

\proof{Proof of Theorem \ref{misspecified}}

For SAA, note that $\hat\omege^{SAA}\xrightarrow{P}\omege^*$ by Proposition \ref{EOconsistency}. Therefore, $R(\hat \omege^{SAA})= \vale_0(\hat \omege^{SAA})-\vale_0(\omege^*) \xrightarrow{P} \vale_0(\omege^*)-\vale_0(\omege^*)=0$ by the continuity of $\vale_0(\omege)$ and the continuous mapping theorem. That is, the regret of SAA is asymptotically 0.

Similarly, for ETO, note that $\hat\thete^{ETO}\xrightarrow{P}\thete^{KL}$ by Proposition \ref{ETOconsistency}. Hence, we have that
$R(\hat \omege^{ETO})=R( \omege_{\hat \thete^{ETO}})=\vale_0(\omege_{\hat \thete^{ETO}})-\vale_0(\omege^*)\xrightarrow{P} \vale_0(\omege_{\thete^{KL}})-\vale_0(\omege^*)$ by the continuity of $\vale_0(\omege)$ and $\omege_\thete$.

Similarly, for IEO, note that $\hat\thete^{IEO}\xrightarrow{P}\thete^*$ by Proposition \ref{IEOconsistency}. Hence, we have that
$R(\hat \omege^{IEO})=R( \omege_{\hat \thete^{IEO}})=\vale_0(\omege_{\hat \thete^{IEO}})-\vale_0(\omege^*)\xrightarrow{P} \vale_0(\omege_{\thete^*})-\vale_0(\omege^*)$ by the continuity of $\vale_0(\omege)$ and $\omege_\thete$.

Comparing ETO and IEO, we must have $\vale_0(\omege_{\thete^{KL}})\geq \vale_0(\omege_{\thete^*})$ by the definition of $\thete^*$ in Assumption \ref{IEOconsistency:assm}. The conclusion of the theorem follows.
\Halmos
\endproof

\subsection{Proofs of Results in Section \ref{sec:cons}}

\proof{Proof of Theorem \ref{RCforEO:cons}}
This proposition is given by Corollary 1 in \citet{duchi2021asymptotic}.
\Halmos
\endproof

\proof{Proof of Theorem \ref{correctlyspecified:cons}}
The proof is similar to Theorem \ref{correctlyspecified}.
\Halmos
\endproof

\proof{Proof of Theorem \ref{SD2}}
In the following proof, we always write $$\bar{\vale}(\omege,\thete)= \vale(\omege, \thete)+ \sum_{j\in J} \alpha_j(\thete) g_j(\omege)$$ 
as a function of $(\omege,\thete)$, and $$\bar{\vale}_0(\omege)= \vale_0(\omege)+ \sum_{j\in J} \alpha^*_j g_j(\omege)$$ as a function of $\omege$. Note that
$$g_j(\omege)-g_j(\omege^*)= \nabla_{\omege}g_j(\omege^*)(\omege-\omege^*)+\frac{1}{2}(\omege-\omege^*)^\top\nabla_{\omege\omege}g_j(\omege^*)(\omege-\omege^*)+ o(\|\omege-\omege^*\|_2^2)$$
or equivalently,
$$- \nabla_{\omege}g_j(\omege^*)(\omege-\omege^*)=\frac{1}{2}(\omege-\omege^*)^\top\nabla_{\omege\omege}g_j(\omege^*)(\omege-\omege^*)+ (g_j(\omege^*)-g_j(\omege)) +o(\|\omege-\omege^*\|_2^2).$$

With the KKT conditions in Assumption \ref{OCforZ:cons}, we have that
\begin{align}
&\vale_0(\omege)-\vale_0(\omege^*)\nonumber\\
=&\nabla_{\omege}\vale_0(\omege^*)(\omege-\omege^*)+\frac{1}{2}(\omege-\omege^*)^\top\nabla_{\omege\omege}\vale_0(\omege^*)(\omege-\omege^*)+ 
o(\|\omege-\omege^*\|_2^2)\nonumber\\
=&-\left(\sum_{j\in J} \alpha^*_j \nabla_{\omege} g_j(\omege^*)\right)(\omege-\omege^*)+\frac{1}{2}(\omege-\omege^*)^\top\nabla_{\omege\omege}\vale_0(\omege^*)(\omege-\omege^*)+ 
o(\|\omege-\omege^*\|_2^2)\nonumber\\
=&\sum_{j\in J} \alpha^*_j \left(\frac{1}{2}(\omege-\omege^*)^\top   \nabla_{\omege\omege}g_j(\omege^*)  (\omege-\omege^*)+(g_j(\omege^*)-g_j(\omege))\right)\nonumber\\
&+\frac{1}{2}(\omege-\omege^*)^\top\nabla_{\omege\omege}\vale_0(\omege^*)(\omege-\omege^*)+ 
o(\|\omege-\omege^*\|_2^2)\nonumber\\
=&\frac{1}{2}(\omege-\omege^*)^\top \nabla_{\omege\omege}\bar{\vale}_0(\omege^*) (\omege-\omege^*)+\sum_{j\in J} \alpha^*_j (g_j(\omege^*)-g_j(\omege)) +
o(\|\omege-\omege^*\|_2^2).
\label{excess risk:cons:old}
\end{align}
In particular, this equation holds for $\omege=\hat\omege^{ETO}$, $\hat\omege^{SAA}$ and $\hat\omege^{IEO}$ (with $o$ replaced by $o_P$). Similarly, as $\omege_\thete$ is a twice differentiable function of $\thete$, we have that under the optimality conditions,
\begin{equation}
\vale_0(\omege_{\thete})-\vale_0(\omege^*)=\frac{1}{2}(\thete-\thete_0)^\top\nabla_{\thete\thete}\vale_0(\omege_{\thete_0})(\thete-\thete_0)+o(\|\thete-\thete_0\|_2^2).\label{excess risk1:cons}
\end{equation}
In particular, this equation holds for $\thete=\hat{\thete}^{ETO}$ and $\hat{\thete}^{IEO}$ (with $o$ replaced by $o_P$). Before going into the comparison of the three approaches, we point out several facts. 

\textbf{Claim 1.} We claim that
\begin{equation}\label{equ:strictequal}
\alpha^*_j (g_j(\omege^*)-g_j(\hat{\omege}))=0, \quad \forall j\in J    
\end{equation}
with probability $p_n$ where $p_n\to 1$ for $\hat{\omege}=\hat\omege^{ETO}$, $\hat\omege^{SAA}$, $\hat\omege^{IEO}$ and for any random sequence $\omege_{\hat{\thete}_n}$ as long as $\hat{\thete}_n\xrightarrow{P} \thete_0$ (noting that $p_n=p_n(\hat{\omege})$ depends on the random sequence $\hat{\omege}$ but we omit it for short).

To see this, we first consider $\hat\omege^{SAA}$. For an index $j$ such that $\alpha^*_j= 0$ or $j\in J_2$ (the set of all equality constraints), $\alpha^*_j (g_j(\omege^*)-g_j(\hat{\omege}^{SAA}))=0$  naturally holds. For an index $j\in J_1$ such that $\alpha^*_j\ne 0$, we have the corresponding $g_j(\omege^*)=0$ by complementary slackness in Assumption \ref{OCforZ:cons}, which implies that $j\in J_1\cap B \cap \{j\in J: \alpha_j^*\ne 0\}$. For this $j$, since $\hat{\alpha}^{SAA}_j\xrightarrow{P}\alpha^*_j\ne 0$ by Assumption \ref{OCforZ:cons}, we must have $\hat{\alpha}^{SAA}_j\ne 0$ with high probability converging to 1, which implies that the corresponding $g_j(\hat{\omege}^{SAA})=0$ by complementary slackness in Assumption \ref{OCforZ:cons}. Therefore we have that $\alpha^*_j (g_j(\omege^*)-g_j(\hat{\omege}^{SAA}))=0$ with high probability converging to 1. We conclude that
$$\alpha^*_j (g_j(\omege^*)-g_j(\hat{\omege}^{SAA}))=0, \quad \forall j\in J$$
with high probability converging to 1. 

We can similarly prove this result for any random sequence $\omege_{\hat{\thete}_n}$ as long as $\hat{\thete}_n\xrightarrow{P} \thete_0$, which obviously includes $\hat\omege^{ETO}$ and $\hat\omege^{IEO}$. It is sufficient to note that by the continuity of $\alpha_j(\thete)$ with respect to $\thete$, 
we have that $\alpha_j(\hat{\thete}_n)\xrightarrow{P}\alpha^*_j$ as $\thete_n\xrightarrow{P} \thete_0$,
$\hat{\alpha}^{ETO}_j=\alpha_j(\hat{\thete}^{ETO})\xrightarrow{P}\alpha^*_j$ as $\hat{\thete}^{ETO}\xrightarrow{P}\thete_0$, and $\hat{\alpha}^{IEO}_j=\alpha_j(\hat{\thete}^{IEO})\xrightarrow{P}\alpha^*_j$ as $\hat{\thete}^{IEO}\xrightarrow{P}\thete_0$.

Finally, we remark that \eqref{equ:strictequal} implies a \textit{strict equality} with high probability, which is stronger than claiming $\alpha^*_j (g_j(\omege^*)-g_j(\hat{\omege}))\xrightarrow{P}0$. 

\textbf{Claim 2.}
We claim that for $\hat{\omege}=\hat\omege^{ETO}$, $\hat\omege^{SAA}$, $\hat\omege^{IEO}$,
\begin{equation}\label{excess risk:cons:final}
\mathbb{P}\left(\vale_0(\hat{\omege})-\vale_0(\omege^*)\le \frac{t}{n}\right)  \to \text{the limit of } \mathbb{P}\left(\frac{1}{2}(\hat{\omege}-\omege^*)^\top \nabla_{\omege\omege}\bar{\vale}_0(\omege^*)(\hat{\omege}-\omege^*) \le \frac{t}{n}\right)    
\end{equation}
as $n\to \infty$ for any $t\in\mathbb{R}$.

Since \eqref{excess risk:cons:old} is a strict equality, it implies that,
for $\hat{\omege}=\hat\omege^{ETO}$, $\hat\omege^{SAA}$, $\hat\omege^{IEO}$, with probability $p_n$ where $p_n\to 1$, 
\begin{equation} \label{excess risk:cons}
\vale_0(\hat{\omege})-\vale_0(\omege^*)=\frac{1}{2}(\hat{\omege}-\omege^*)^\top \nabla_{\omege\omege}\bar{\vale}_0(\omege^*)(\hat{\omege}-\omege^*) +
o_P(\|\hat{\omege}-\omege^*\|^2).    
\end{equation}
By the law of total probability, 
\begin{align*}
&\mathbb{P}(\vale_0(\hat{\omege})-\vale_0(\omege^*)\le \frac{t}{n})\\
= & (1-p_n) \mathbb{P}\left(\vale_0(\hat{\omege})-\vale_0(\omege^*)\le \frac{t}{n}\Big|\sum_{j\in J} \alpha^*_j (g_j(\omege^*)-g_j(\hat{\omege}))\ne 0\right)\\
&+\mathbb{P}\left(\vale_0(\hat{\omege})-\vale_0(\omege^*)-\sum_{j\in J} \alpha^*_j (g_j(\omege^*)-g_j(\hat{\omege}))\le \frac{t}{n}, \sum_{j\in J} \alpha^*_j (g_j(\omege^*)-g_j(\hat{\omege}))= 0\right)\\
\to & \text{the limit of } \mathbb{P}\left(\vale_0(\hat{\omege})-\vale_0(\omege^*)-\sum_{j\in J} \alpha^*_j (g_j(\omege^*)-g_j(\hat{\omege}))\le \frac{t}{n}, \sum_{j\in J} \alpha^*_j (g_j(\omege^*)-g_j(\hat{\omege}))= 0\right)\\
= & \text{the limit of } \mathbb{P}\left(\frac{1}{2}(\hat{\omege}-\omege^*)^\top \nabla_{\omege\omege}\bar{\vale}_0(\omege^*)(\hat{\omege}-\omege^*) \le \frac{t}{n}\right)
\end{align*}
as $n\to \infty$ for any $t\in\mathbb{R}$ where we have use the fact that $\mathbb{P}(A_n\cap B_n)=\mathbb{P}(A_n)+\mathbb{P}(B_n)-\mathbb{P}(A_n\cup B_n)\to \mathbb{P}(A)$ if $\mathbb{P}(B_n)\to 1$ and $\mathbb{P}(A_n)\to \mathbb{P}(A)$.

Therefore, to obtain the limiting distribution of $\vale_0(\hat{\omege})-\vale_0(\omege^*)$, we only need to study the limiting distribution of \eqref{excess risk:cons:final}.

\textbf{Claim 3.} We claim that there exists a $\varepsilon>0$ such that
\begin{equation}\label{secondobservation}
\alpha^*_j (g_j(\omege^*)-g_j(\omege_\thete))=0, \quad \forall j\in J   
\end{equation}
for any $\thete\in \{\thete\in\Theta:\|\thete-\thete_0\|_2\le \varepsilon\}$. The proof is similar to the proof of \eqref{equ:strictequal} by using the continuity of $\alpha_j(\thete)$ and the KKT conditions.

\textbf{Claim 4.} We claim that
\begin{equation}\label{excess risk:cons:new}
\vale_0(\omege_{\thete})-\vale_0(\omege^*)
=\frac{1}{2}(\omege_{\thete}-\omege^*)^\top\nabla_{\omege\omege}\bar{\vale}_0(\omege^*)(\omege_{\thete}-\omege^*)+ 
o(\|\omege_{\thete}-\omege^*\|_2^2)
\end{equation}
for any $\thete\in \{\thete\in\Theta:\|\thete-\thete_0\|_2\le \varepsilon\}$. This follows from plugging \eqref{secondobservation} into \eqref{excess risk:cons:old}.

\textbf{Claim 5.} We claim that
\begin{equation}\label{thirdobservation}
\alpha^*_j \nabla_{\thete} g_j(\omege_{\thete_0})=\alpha^*_j \nabla_{\omege} g_j(\omege^*)\nabla_\thete \omege_{\thete_0}=0, \quad \forall j\in J,  
\end{equation}
\begin{equation}\label{fourthobservation}
 \alpha^*_j \nabla_{\thete\thete} g_j(\omege_{\thete_0})=0, \quad \forall j\in J,   
\end{equation}
\begin{equation}\label{fifthobservation}
\nabla_{\thete} g_j(\omege_{\thete_0})= \nabla_{\omege} g_j(\omege^*)\nabla_\thete \omege_{\thete_0}=0, \quad \forall j\in B.  
\end{equation}

First, note that Equalities \eqref{thirdobservation} and \eqref{fourthobservation} follow from \eqref{secondobservation} because \eqref{secondobservation} holds for any $\thete\in \{\thete\in\Theta:\|\thete-\thete_0\|_2\le \varepsilon\}$ implying that $\nabla_{\thete} \alpha^*_j (g_j(\omege^*)-g_j(\omege_\thete))|_{\thete=\thete_0}=0$ and $\nabla_{\thete\thete} \alpha^*_j (g_j(\omege^*)-g_j(\omege_\thete))|_{\thete=\thete_0}=0$ for all $j\in J$. 

Second, for any $j\in B\cap J_2=J_2$ (the set of all equality constraints), we have that $g_j(\omege_{\thete})=0$ for all $\thete$, which clearly implies that 
$$\nabla_{\thete} g_j(\omege_{\thete_0})= \nabla_{\omege} g_j(\omege^*)\nabla_\thete \omege_{\thete_0}=0, \quad j\in B\cap J_2.$$
In addition, for any $j\in B\cap J_1$ (the set of all active inequality constraints), since $g_j(\omege_{\thete_0})=0$ while $g_j(\omege_{\thete})\le 0$ for all $\thete$, $\thete_0$ (which is an inner point in $\Theta$) is a point of maximum for the function $g_j(\omege_{\thete})$, and thus
$$\nabla_{\thete} g_j(\omege_{\thete_0})= \nabla_{\omege} g_j(\omege^*)\nabla_\thete \omege_{\thete_0}=0, \quad j\in B\cap J_1.$$
Equality \eqref{fifthobservation} then follows. 

\textbf{Claim 6.} We claim that
\begin{equation} \label{equ:strictcomp}
\Phi\nabla_\thete \omege_{\thete_0}=\nabla_\thete \omege_{\thete_0}  
\end{equation}
where $\Phi$ is given in Proposition \ref{RCforEO:cons}.
Note that \eqref{fifthobservation} is equivalent to $A\nabla_\thete \omege_{\thete_0}=0$, and thus
$$\Phi\nabla_\thete \omege_{\thete_0}=(I-A^\top (AA^\top )^{-1}A)\nabla_\thete \omege_{\thete_0}=\nabla_\thete \omege_{\thete_0}.$$

\textbf{Claim 7.} We claim that
\begin{equation}\label{equ:positive}
\Phi\nabla_{\omege\omege}\bar{\vale}_0(\omege^*)\Phi\ge 0, \quad \text{and}  \quad   \text{rank}(\Phi\nabla_{\omege\omege}\bar{\vale}_0(\omege^*)\Phi)=\text{rank}(\Phi).
\end{equation}
In fact, for any $\omege\in \mathbb{R}^p$,
$$A\Phi\omege=A(I-A^\top (AA^\top )^{-1}A)\omege=A\omege-A\omege=0$$
which implies that $\Phi\omege\in \mathcal{T}(\omege^*)$. Hence by the second-order optimality conditions in Assumption \ref{RCforEO:cons:assm}, $\omege^\top\Phi\nabla_{\omege\omege}\bar{\vale}_0(\omege^*)\Phi\omege\ge \mu\|\Phi\omege\|^2\ge 0$. 

In addition, noting that $\Phi$ is the orthogonal projection onto the tangent set $\mathcal{T}(\omege^*)$, we have
$$\text{rank}(\Phi)=\text{dim}(\mathcal{T}(\omege^*))$$
where $\Phi\omege=0$ if and only if $\omege\in \mathcal{T}(\omege^*)^{\perp}$ where $\omege\in \mathcal{T}(\omege^*)^{\perp}$ is the orthogonal complement of $\mathcal{T}(\omege^*)$. Note that whenever $w\in \mathcal{T}(\omege^*)\setminus \{0\} \notin \mathcal{T}(\omege^*)^{\perp}$, we have
$\Phi\omege\ne 0$,  $\omege^\top\Phi\nabla_{\omege\omege}\bar{\vale}_0(\omege^*)\Phi\omege\ge \mu\|\Phi\omege\|^2> 0$ and thus $\Phi\nabla_{\omege\omege}\bar{\vale}_0(\omege^*)\Phi\omege\ne 0$. This implies that
$$\text{ker}(\Phi\nabla_{\omege\omege}\bar{\vale}_0(\omege^*)\Phi) \cap \mathcal{T}(\omege^*)=\{0\}$$
Hence
$$\text{dim}(\text{ker}(\Phi\nabla_{\omege\omege}\bar{\vale}_0(\omege^*)\Phi))+\dim(\mathcal{T}(\omege^*))\le p$$
which gives
$$p-\text{rank}(\Phi\nabla_{\omege\omege}\bar{\vale}_0(\omege^*)\Phi)+\text{rank}(\Phi)\le p$$
and thus
$$\text{rank}(\Phi\nabla_{\omege\omege}\bar{\vale}_0(\omege^*)\Phi)\ge \text{rank}(\Phi)$$
The other side is trivial: 
$$\text{rank}(\Phi\nabla_{\omege\omege}\bar{\vale}_0(\omege^*)\Phi)\le \min(\text{rank}(\Phi),\text{rank}(\nabla_{\omege\omege}\bar{\vale}_0(\omege^*)\Phi))\le \text{rank}(\Phi).$$
Hence,
$$\text{rank}(\Phi\nabla_{\omege\omege}\bar{\vale}_0(\omege^*)\Phi)=\text{rank}(\Phi).$$

\textbf{Claim 8.} We have that
\begin{equation}
\nabla_{\omege\omege}\bar{\vale}(\omege^*,\thete_0)\nabla_{\thete} \omege_{\thete_0}+
\nabla_{\omege\thete}\vale(\omege^*,\thete_0)+ \sum_{j\in J} \nabla_{\omege} g_j(\omege^*)^\top \nabla_{\thete} \alpha_j(\thete_0)=0,
\label{interim:cons}
\end{equation} 
and
\begin{equation}\label{interim:consFact2}
\nabla_{\thete} \omege_{\thete_0}^\top \nabla_{\omege\omege}\bar{\vale}(\omege^*,\thete_0)\nabla_{\thete} \omege_{\thete_0}+\nabla_{\thete} \omege_{\thete_0}^\top
\nabla_{\omege\thete}\vale(\omege^*,\thete_0)=0.
\end{equation}

For \eqref{interim:cons}, the reason is similar to \eqref{interim}: We use the chain rule on the first-order gradient of the Lagrangian function in the KKT conditions in Assumption \ref{OCforZ:cons} to get
\begin{align*}
0=&\nabla_{\thete}\left(\nabla_{\omege} \vale(\omege_\thete, \thete)+ \sum_{j\in J} \alpha_j(\thete) \nabla_{\omege} g_j(\omege_\thete)\right)|_{\thete=\thete_0}\\
=&
\left(\nabla_{\omege\omege} \vale(\omege_\thete, \thete)+ \sum_{j\in J} \alpha_j(\thete) \nabla_{\omege\omege} g_j(\omege_\thete)\right)\nabla_\thete \omege_{\thete}+
\left(\nabla_{\omege\thete}\vale(\omege_\thete,\thete)+ \sum_{j\in J} \nabla_{\omege} g_j(\omege_\thete)^\top \nabla_{\thete}\alpha_j(\thete) \right)|_{\thete=\thete_0}
\end{align*}

For \eqref{interim:consFact2}, note that \eqref{interim:cons} implies that
\begin{equation}\label{interim:consFact1}
\nabla_{\thete} \omege_{\thete_0}^\top \nabla_{\omege\omege}\bar{\vale}(\omege^*,\thete_0)\nabla_{\thete} \omege_{\thete_0}+\nabla_{\thete} \omege_{\thete_0}^\top
\left(\nabla_{\omege\thete}\vale(\omege^*,\thete_0)+ \sum_{j\in J} \nabla_{\omege} g_j(\omege^*)^\top \nabla_{\thete} \alpha_j(\thete_0) \right)=0.    
\end{equation}
When $j\in B$, \eqref{fifthobservation} shows that
$$\nabla_{\thete}\omege_{\thete_0}^\top \nabla_{\omege} g_j(\omege^*)^\top = (\nabla_{\omege} g_j(\omege^*)\nabla_{\thete}\omege_{\thete_0})^\top=0$$
When $j\notin B$, that is, $g_j(\omege^*)=g_j(\omege_{\thete_0})<0$, the continuity implies that there exists a $\varepsilon_j>0$ such that $g_j(\omege_\thete)<0$
for any $\thete\in \{\thete\in\Theta:\|\thete-\thete_0\|_2\le \varepsilon_j\}$. Hence complementary slackness in Assumption \ref{OCforZ:cons} implies that $\alpha_j(\thete)=0$ for any $\thete\in \{\thete\in\Theta:\|\thete-\thete_0\|_2\le \varepsilon_j\}$, which shows that $\nabla_{\thete} \alpha_j(\thete_0)=0$. 

Hence we conclude that for any $j\in J$,
$$\nabla_{\thete}\omege_{\thete_0}^\top \nabla_{\omege} g_j(\omege^*)^\top\nabla_{\thete} \alpha_j(\thete_0)=0.$$
\eqref{interim:consFact2} then follows from \eqref{interim:consFact1}.

\textbf{Claim 9.} We have that 
\begin{equation} \label{equ:Hessian:cons}
\nabla_{\thete\thete}\vale_0(\omege_{\thete_0})=\nabla_\thete \omege_{\thete_0}^\top \nabla_{\omege\omege}\bar{\vale}_0(\omege^*)\nabla_\thete \omege_{\thete_0}.    
\end{equation}
In fact, we have that
\begin{align*}
&\nabla_{\thete\thete}\vale_0(\omege_{\thete})|_{\thete=\thete_0}\\
=& \nabla_{\thete} (\nabla_{\omege} \vale_0(\omege_\thete) \nabla_\thete \omege_{\thete})|_{\thete=\thete_0}\\
=&\nabla_\thete \omege_{\thete}^\top \nabla_{\omege\omege} \vale_0(\omege_\thete) \nabla_\thete \omege_{\thete}|_{\thete=\thete_0}+ \nabla_{\omege} \vale_0(\omege_\thete)\nabla_{\thete} ( \nabla_\thete \omege_{\thete})|_{\thete=\thete_0} \\
=&\nabla_\thete \omege_{\thete_0}^\top \nabla_{\omege\omege} \vale_0(\omege^*) \nabla_\thete \omege_{\thete_0}- \sum_{j\in J} \alpha^*_j \nabla_{\omege} g_j(\omege_\thete) \nabla_{\thete} ( \nabla_\thete \omege_{\thete})|_{\thete=\thete_0} \quad \text{ by Assumption } \ref{OCforZ:cons}\\
=& \nabla_\thete \omege_{\thete_0}^\top \nabla_{\omege\omege} \vale_0(\omege^*) \nabla_\thete \omege_{\thete_0}+ \sum_{j\in J} \alpha^*_j \nabla_\thete \omege_{\thete}^\top \nabla_{\omege\omege} g_j(\omege_\thete) \nabla_\thete \omege_{\thete}|_{\thete=\thete_0} \quad \text{ by } \eqref{fourthobservation} \\
=& \nabla_\thete \omege_{\thete_0}^\top \nabla_{\omege\omege}\bar{\vale}_0(\omege^*)\nabla_\thete \omege_{\thete_0}.
\end{align*} 
Another way to show \eqref{equ:Hessian:cons} is to notice that both \eqref{excess risk:cons:new} and \eqref{excess risk1:cons} hold for any $\thete$ close to $\thete_0$:
\begin{align*}
\vale_0(\omege_{\thete})-\vale_0(\omege^*)&=\frac{1}{2}(\thete-\thete_0)^\top\nabla_{\thete\thete}\vale_0(\omege_{\thete_0})(\thete-\thete_0)+o(\|\thete-\thete_0\|_2^2) \\
& =\frac{1}{2}(\omege_{\thete}-\omege^*)^\top\nabla_{\omege\omege}\bar{\vale}_0(\omege^*)(\omege_{\thete}-\omege^*)+ 
o(\|\omege_{\thete}-\omege^*\|_2^2)\\
& =\frac{1}{2}(\thete-\thete_0)^\top \nabla_\thete \omege_{\thete_0}^\top \nabla_{\omege\omege}\bar{\vale}_0(\omege^*)\nabla_\thete \omege_{\thete_0}(\thete-\thete_0)+ 
o(\|\thete-\thete_0\|_2^2)
\end{align*}
which implies that $\nabla_{\thete\thete}\vale_0(\omege_{\thete_0})=\nabla_\thete \omege_{\thete_0}^\top \nabla_{\omege\omege}\bar{\vale}_0(\omege^*)\nabla_\thete \omege_{\thete_0}.$

\textbf{Step 1:} We show that $$\mathbb G^{ETO}\preceq_{st}\mathbb G^{IEO}.$$

To show this, we compare the performance of ETO and IEO at the level of $\thete$ and then leverage Equation \eqref{excess risk1}. 
Note that ETO can be equivalently written as
$$\hat\omege^{ETO} =\min_{\omege\in\Omega} \vale(\omege,\hat{\thete}^{ETO})=\omege_{\hat{\thete}^{ETO}}$$
using the oracle problem \eqref{equ:oracle} by plugging in the MLE $\hat{\thete}^{ETO}$.

For ETO, Proposition \ref{RCforETO} gives that  
\begin{equation} \label{equ:MLE:cons}
\sqrt n(\hat{\thete}^{ETO}-\thete_0)\xrightarrow{d} N(0,\mathcal{I}_{\thete_0}^{-1}).
\end{equation}

Plugging \eqref{equ:MLE:cons} into \eqref{excess risk1:cons}, we have
$$n(\vale_0(\hat\omege^{ETO})-\vale_0(\omege^*))\xrightarrow{d}\mathbb G^{ETO}$$
where $\mathbb G^{ETO}= 
\frac{1}{2}{\mathcal N_1^{ETO}}^\top\nabla_{\thete\thete}\vale_0(\omege_{\thete_0})\mathcal N_1^{ETO}$ and 
\begin{equation} \label{equ:ETO1:cons}
\mathcal N_1^{ETO}\sim N(0,\mathcal{I}_{\thete_0}^{-1})    
\end{equation}


For IEO, Proposition \ref{RCforIEO} gives that
\begin{equation} \label{Mestimator:cons}
\sqrt{n}(\hat{\thete}^{IEO}-\thete_0)\xrightarrow{d} N(0,\nabla_{\thete\thete}\vale_0(\omege_{\thete_0})^{-1} Var_{P}(\nabla_\thete \cost(\omege_{\thete_0},\ran))\nabla_{\thete\thete}\vale_0(\omege_{\thete_0})^{-1}).    
\end{equation}

Plugging \eqref{Mestimator:cons} into \eqref{excess risk1:cons}, we have
$$n(\vale_0(\hat\omege^{IEO})-\vale_0(\omege^*))\xrightarrow{d}\mathbb G^{IEO}$$
where $\mathbb G^{IEO}=\frac{1}{2}{\mathcal N_1^{IEO}}^\top\nabla_{\thete\thete}\vale_0(\omege_{\thete_0})\mathcal N_1^{IEO}$ and 
\begin{equation} \label{equ:IEO1:cons}
\mathcal N_1^{IEO}\sim N(0,\nabla_{\thete\thete}\vale_0(\omege_{\thete_0})^{-1} Var_{P}(\nabla_\thete \cost(\omege_{\thete_0},\ran))\nabla_{\thete\thete}\vale_0(\omege_{\thete_0})^{-1}).    
\end{equation}

From Lemma \ref{lemma1}, we can complete Step 1 by showing that
\begin{equation}\label{CR2:cons}
\nabla_{\thete\thete}\vale_0(\omege_{\thete_0})^{-1} Var_{P}(\nabla_\thete \cost(\omege_{\thete_0},\ran))\nabla_{\thete\thete}\vale_0(\omege_{\thete_0})^{-1}\ge \mathcal{I}_{\thete_0}^{-1}.    
\end{equation}
We shall use the multivariate Cramer-Rao bound \citep{cramer1946mathematical,rao1945information,bickel2015mathematical}, which is also stated in Lemma \ref{lemma:Cramer-Rao} in Section \ref{sec:multivariateCramerRao}.
Recall that $P=P_{\thete_0}$ and $\omege^*=\omege_{\thete_0}$.
Consider a random vector $\ran\sim P_\thete$ and an estimator with the following form $\nabla_\omege \cost(\omege^*,\ran)$, which has the expectation $\mathbb{E}_\thete[\nabla_\omege \cost(\omege^*,\ran)]$. It follows from Assumption \ref{SCforh} that
$\mathbb{E}_\thete[\nabla_\omege \cost(\omege^*,\ran)] = \nabla_\omege \mathbb{E}_\thete [\cost(\omege,\ran)]|_{\omege=\omege^*} = \nabla_\omege \vale(\omege^*, \thete)$. Therefore we have that 
$$\nabla_\thete (\mathbb{E}_\thete[\nabla_\omege \cost(\omege^*,\ran)])^\top |_{\thete=\thete_0} = \nabla_\thete (\nabla_\omege \vale(\omege^*, \thete))^\top |_{\thete=\thete_0}= \nabla_{\omege\thete} \vale(\omege^*, \thete_0).$$
Applying Cramer-Rao bound on the estimator $ \nabla_\omege \cost(\omege^*,\ran)$ (assured by Assumption \ref{SCforh}), we have that
\begin{equation}\label{CR:cons}
Var_{P}\Big(\nabla_\omege \cost(\omege^*,\ran)\Big) \geq  \nabla_{\omege\thete}\vale(\omege^*, \thete_0) \mathcal{I}_{\thete_0}^{-1}\nabla_{\thete\omege}\vale(\omege^*, \thete_0).    
\end{equation}
We can then show that
\begin{align*}
&Var_{P}(\nabla_\thete \cost(\omege_{\thete_0},\ran)) \\
=&Var_{P}(\nabla_\omege \cost(\omege^*,\ran)\nabla_\thete \omege_{\thete_0}) \\
=&\nabla_\thete \omege_{\thete_0}^\top Var_{P}(\nabla_\omege \cost(\omege^*,\ran)) \nabla_\thete \omege_{\thete_0} \quad\text{ since $\nabla_\thete \omege_{\thete_0}$ is deterministic}\\
\geq & \nabla_\thete \omege_{\thete_0}^\top\nabla_{\omege\thete}\vale(\omege^*, \thete_0) \mathcal{I}_{\thete_0}^{-1}\nabla_{\thete\omege}\vale(\omege^*, \thete_0)\nabla_\thete \omege_{\thete_0} \quad\text{ by } \eqref{CR:cons} \\
=  &\nabla_\thete \omege_{\thete_0}^\top\nabla_{\omege\omege}\bar{\vale}(\omege^*,\thete_0)\nabla_\thete \omege_{\thete_0} \mathcal{I}_{\thete_0}^{-1} \nabla_\thete \omege_{\thete_0}^\top\nabla_{\omege\omege}\bar{\vale}(\omege^*,\thete_0)\nabla_\thete \omege_{\thete_0}  \quad\text{ by } \eqref{interim:consFact2} \\
= & \nabla_{\thete\thete}\vale_0(\omege_{\thete_0}) \mathcal{I}_{\thete_0}^{-1} \nabla_{\thete\thete}\vale_0(\omege_{\thete_0}) \quad\text{ by } \eqref{equ:Hessian:cons}.
\end{align*}
Since $\nabla_{\thete\thete}\vale_0(\omege_{\thete_0})^{-1}$ exists due to Assumption \ref{RCforIEO:assm},  then multiplying by the inverse twice gives \eqref{CR2:cons}.


Hence, Comparing ETO \eqref{equ:ETO1:cons} and IEO \eqref{equ:IEO1:cons} by using Lemma \ref{lemma1}, we conclude that 
$$\mathbb G^{ETO}\preceq_{st}\mathbb G^{IEO}.$$

\textbf{Step 2:} We show that $$\mathbb G^{IEO}\preceq_{st}\mathbb G^{SAA}.$$


To show this, we compare the performance of IEO and SAA at the level of $\omege$ and then leverage Equation \eqref{excess risk:cons}. 
For IEO, we reuse the fact \eqref{Mestimator:cons} to obtain, by the Delta method,
\begin{equation}
\sqrt n(\hat\omege^{IEO}-\omege^*)\xrightarrow{d} N(0,\nabla_\thete \omege_{\thete_0}\nabla_{\thete\thete}\vale_0(\omege_{\thete_0})^{-1} Var_{P}(\nabla_\thete \cost(\omege_{\thete_0},\ran))\nabla_{\thete\thete}\vale_0(\omege_{\thete_0})^{-1}\nabla_\thete \omege_{\thete_0}^\top).
\end{equation}


Next we notice that
$$Var_{P}(\nabla_\thete \cost(\omege_{\thete_0},\ran))=Var_{P}(\nabla_\omege \cost(\omege^*,\ran)\nabla_\thete \omege_{\thete_0})=\nabla_\thete \omege_{\thete_0}^\top Var_{P}(\nabla_\omege \cost(\omege^*,\ran)) \nabla_\thete \omege_{\thete_0}$$
since $\nabla_\thete \omege_{\thete_0}$ is deterministic. 

Hence, we have
\begin{equation*}
\sqrt n(\hat\omege^{IEO}-\omege^*)\xrightarrow{d} N(0,\nabla_\thete \omege_{\thete_0}\nabla_{\thete\thete}\vale_0(\omege_{\thete_0})^{-1} \nabla_\thete \omege_{\thete_0}^\top Var_{P}(\nabla_\omege \cost(\omege^*,\ran))\nabla_\thete \omege_{\thete_0} \nabla_{\thete\thete}\vale_0(\omege_{\thete_0})^{-1}\nabla_\thete \omege_{\thete_0}^\top).
\end{equation*}
which is equivalent to 
\begin{align}
&\sqrt n(\hat\omege^{IEO}-\omege^*)\xrightarrow{d} \nonumber\\
& N(0,\Phi^2\nabla_\thete \omege_{\thete_0}\nabla_{\thete\thete}\vale_0(\omege_{\thete_0})^{-1} \nabla_\thete \omege_{\thete_0}^\top \Phi Var_{P}(\nabla_\omege \cost(\omege^*,\ran))\Phi \nabla_\thete \omege_{\thete_0} \nabla_{\thete\thete}\vale_0(\omege_{\thete_0})^{-1}\nabla_\thete \omege_{\thete_0}^\top\Phi^2)\label{equ:IEOcovergence:cons}
\end{align}
since $\Phi^2\nabla_\thete \omege_{\thete_0}= \Phi\nabla_\thete \omege_{\thete_0}=\nabla_\thete \omege_{\thete_0}$ by \eqref{equ:strictcomp}.

Plugging \eqref{equ:IEOcovergence:cons} into \eqref{excess risk:cons}, we have
$$n(\vale_0(\hat\omege^{IEO})-\vale_0(\omege^*)) \xrightarrow{d}\mathbb G^{IEO}$$
where $\mathbb G^{IEO}=\frac{1}{2}{\mathcal N_2^{IEO}}^\top\Phi\nabla_{\omege\omege}\bar{\vale}_0(\omege^*)\Phi\mathcal N_2^{IEO}$ and 
\begin{equation} \label{equ:IEO3:cons}
\mathcal N_2^{IEO}\sim N(0,\Phi\nabla_\thete \omege_{\thete_0}\nabla_{\thete\thete}\vale_0(\omege_{\thete_0})^{-1} \nabla_\thete \omege_{\thete_0}^\top \Phi Var_{P}(\nabla_\omege \cost(\omege^*,\ran)) \Phi \nabla_\thete \omege_{\thete_0}\nabla_{\thete\thete}\vale_0(\omege_{\thete_0})^{-1}\nabla_\thete \omege_{\thete_0}^\top \Phi).    
\end{equation}
Note that we use $\Phi\nabla_{\omege\omege}\bar{\vale}_0(\omege^*)\Phi$ instead of $\nabla_{\omege\omege}\bar{\vale}_0(\omege^*)$ in the formula of $\mathbb G^{IEO}$ since $\Phi\nabla_{\omege\omege}\bar{\vale}_0(\omege^*)\Phi\ge 0$ by \eqref{equ:positive}, which then can be applied in Lemma \ref{lemma1}. However, $\nabla_{\omege\omege}\bar{\vale}_0(\omege^*)$ may not be positive definite and cannot be used in Lemma \ref{lemma1}.

Now, for SAA, Proposition \ref{RCforEO:cons} gives that
\begin{equation}
\sqrt n(\hat\omege^{SAA}-\omege^*)\xrightarrow{d} N(0,\Phi^2(\Phi\nabla_{\omege\omege}\bar{\vale}_0(\omege^*)\Phi)^{\dagger}\Phi Var_{P}(\nabla_\omege \cost(\omege^*,\ran))\Phi(\Phi\nabla_{\omege\omege}\bar{\vale}_0(\omege^*)\Phi)^{\dagger}\Phi^2).\label{EO decision:cons}
\end{equation}
where we use $\Phi^2=\Phi$.

Plugging \eqref{EO decision:cons} into \eqref{excess risk:cons}, we have
$$n(\vale_0(\hat\omege^{SAA})-\vale_0(\omege^*)) \xrightarrow{d}\mathbb G^{SAA}$$
where $\mathbb G^{SAA}=\frac{1}{2}{\mathcal N^{SAA}}^\top \Phi \nabla_{\omege\omege}\bar{\vale}_0(\omege^*)\Phi\mathcal N^{SAA}$ and 
\begin{equation} \label{equ:EO2:cons}
\mathcal N^{SAA}\sim N(0,\Phi(\Phi\nabla_{\omege\omege}\bar{\vale}_0(\omege^*)\Phi)^{\dagger}\Phi Var_{P}(\nabla_\omege \cost(\omege^*,\ran))\Phi(\Phi\nabla_{\omege\omege}\bar{\vale}_0(\omege^*)\Phi)^{\dagger}\Phi).    
\end{equation}

Comparing IEO \eqref{equ:IEO3:cons} and SAA \eqref{equ:EO2:cons}, note that the difference in the limiting distributions lies in $$\nabla_\thete \omege_{\thete_0}\nabla_{\thete\thete}\vale_0(\omege_{\thete_0})^{-1} \nabla_\thete \omege_{\thete_0}^\top$$ versus $(\Phi\nabla_{\omege\omege}\bar{\vale}_0(\omege^*)\Phi)^{\dagger}$. 

In this regard, we first notice that $\nabla_{\thete\thete}\vale_0(\omege_{\thete_0})=\nabla_\thete \omege_{\thete_0}^\top  \nabla_{\omege\omege} \bar{\vale}_0(\omege^*) \nabla_\thete \omege_{\thete_0}=\nabla_\thete \omege_{\thete_0}^\top \Phi  \nabla_{\omege\omege} \bar{\vale}_0(\omege^*) \Phi \nabla_\thete \omege_{\thete_0}$ by the facts \eqref{equ:Hessian:cons} and \eqref{equ:strictcomp}. Note that in general $\nabla_\thete \omege_{\thete_0}=\Phi \nabla_\thete \omege_{\thete_0}$ is not invertible as $\Phi$ is an orthogonal projection matrix. Next, we use Lemma \ref{lemma3} by setting
\begin{itemize}
\item $Q_0=\Phi$, which is an orthogonal projection matrix;
\item $Q_1= \nabla_{\omege\omege} \bar{\vale}_0(\omege^*)$, where $Q_0 Q_1 Q_0=\Phi\nabla_{\omege\omege} \bar{\vale}_0(\omege^*)\Phi$ is positive semi-definite and $\text{rank}(Q_0 Q_1 Q_0)=\text{rank}(Q_0)$ by \eqref{equ:positive};
\item $Q_2=\Phi Var_{P}(\nabla_\omege \cost(\omege^*,\ran))\Phi\ge 0$ since $Var_{P}(\nabla_\omege \cost(\omege^*,\ran))\ge 0$;
\item $Q_3=\nabla_\thete \omege_{\thete_0}$, where $Q_3^\top Q_0 Q_1 Q_0 Q_3=\nabla_\thete \omege_{\thete_0}^\top \Phi  \nabla_{\omege\omege} \bar{\vale}_0(\omege^*) \Phi \nabla_\thete \omege_{\thete_0}=\nabla_{\thete\thete}\vale_0(\omege_{\thete_0})$ is positive definite by Assumption \ref{RCforIEO:assm}.
\end{itemize}
Then we obtain from Lemma \ref{lemma3} that
\begin{align*}
& \Phi \nabla_\thete \omege_{\thete_0}\nabla_{\thete\thete}\vale_0(\omege_{\thete_0})^{-1} \nabla_\thete \omege_{\thete_0}^\top \Phi Var_{P}(\nabla_\omege \cost(\omege^*,\ran)) \Phi \nabla_\thete \omege_{\thete_0} \nabla_{\thete\thete}\vale_0(\omege_{\thete_0})^{-1}\nabla_\thete \omege_{\thete_0}^\top \Phi\\
\le & \Phi(\Phi\nabla_{\omege\omege}\bar{\vale}_0(\omege^*)\Phi)^{\dagger}\Phi Var_{P}(\nabla_\omege \cost(\omege^*,\ran))\Phi(\Phi\nabla_{\omege\omege}\bar{\vale}_0(\omege^*)\Phi)^{\dagger}\Phi
\end{align*}

Hence, Comparing IEO \eqref{equ:IEO3:cons} and SAA \eqref{equ:EO2:cons} by using Lemma \ref{lemma1}, we conclude that 
$$\mathbb G^{IEO}\preceq_{st}\mathbb G^{SAA}.$$
\Halmos
\endproof

\proof{Proof of Lemma \ref{lemma3}}
Since $Q_0Q_1Q_0$ is positive semi-definite, $Q_0Q_1Q_0+\gamma I_p>0$ for any $\gamma>0$ and thus we apply Lemma \ref{lemma2} to obtain
\begin{align*}
&Q_3(Q_3^\top (Q_0Q_1Q_0+\gamma I_p) Q_3 + \lambda I_q)^{-1}Q_3^\top Q_2 Q_3(Q_3^\top (Q_0Q_1Q_0+\gamma I_p) Q_3 + \lambda I_q)^{-1}Q_3^\top\\
\le &(Q_0Q_1Q_0+\gamma I_p)^{-1} Q_2 (Q_0Q_1Q_0+\gamma I_p)^{-1} 
\end{align*}
Obviously, it implies that
\begin{align}
&Q_0Q_3(Q_3^\top (Q_0Q_1Q_0+\gamma I_p) Q_3 + \lambda I_q)^{-1}Q_3^\top Q_2 Q_3(Q_3^\top (Q_0Q_1Q_0+\gamma I_p) Q_3 + \lambda I_q)^{-1}Q_3^\top Q_0\nonumber\\
\le &Q_0(Q_0Q_1Q_0+\gamma I_p)^{-1} Q_2 (Q_0Q_1Q_0+\gamma I_p)^{-1} Q_0 \label{equ:lemma3_0}
\end{align}

\textbf{Step 1:} We claim that for any vector $\bm{u}$,
\begin{align}
&\lim_{\gamma\to 0}\bm{u}^\top Q_0Q_3(Q_3^\top (Q_0Q_1Q_0+\gamma I_p) Q_3 + \lambda I_q)^{-1}Q_3^\top Q_2 Q_3(Q_3^\top (Q_0Q_1Q_0+\gamma I_p) Q_3 + \lambda I_q)^{-1}Q_3^\top Q_0 \bm{u} \nonumber\\
= & \bm{u}^\top Q_0Q_3(Q_3^\top Q_0Q_1Q_0 Q_3 + \lambda I_q)^{-1}Q_3^\top Q_2 Q_3(Q_3^\top Q_0Q_1Q_0 Q_3 + \lambda I_q)^{-1}Q_3^\top Q_0 \bm{u}    \label{equ:lemma3_1_1}
\end{align}
To prove this, we first notice that for any invertible matrix $Q_4$ and any matrix $Q_5$, we have
$$\lim_{\gamma\to 0}\|Q_4^{-1}-(Q_4 + \gamma Q_5)^{-1}\|_{op}=0$$
This follows from the local Lipschitz continuity of the matrix inversion, but for completeness, we provide proof here. Let $\|Q_4^{-1}\|_{op}=\frac{1}{\delta}>0$ since $Q_4$ is invertible. Then for any vector $\bm{u}$,
$$\|\bm{u}\|_2=\|Q_4^{-1}Q_4\bm{u}\|_2\le \|Q_4^{-1}\|_{op}\|Q_4\bm{u}\|_2$$
we have $\|Q_4\bm{u}\|_2\ge \delta\|\bm{u}\|_2$.
We consider all $\gamma>0$ such that $\gamma\le \frac{\delta}{2 \|Q_5\|_{op}}$. In this case,
$$\|(Q_4+\gamma Q_5)\bm{u}\|_2\ge \|Q_4\bm{u}\|_2-\|\gamma Q_5\bm{u}\|_2\ge \delta \|\bm{u}\|_2-\frac{\delta}{2} \|\bm{u}\|_2=\frac{\delta}{2} \|\bm{u}\|_2$$
showing that $Q_4+\gamma Q_5$ is invertible and has an inverse $(Q_4+\gamma Q_5)^{-1}$ of norm $\le \frac{2}{\delta}$. Note that
$$\|Q_4^{-1}-(Q_4 + \gamma Q_5)^{-1}\|_{op}=\|Q_4^{-1}(\gamma Q_5)(Q_4 + \gamma Q_5)^{-1}\|_{op}\le \frac{2}{\delta^2} \gamma \|Q_5\|_{op}$$
and thus
$$\lim_{\gamma\to 0}\|Q_4^{-1}-(Q_4 + \gamma Q_5)^{-1}\|_{op}=0.$$
Let $Q_4=Q_3^\top Q_0 Q_1 Q_0 Q_3 + \lambda I_q>0$ which is an invertible matrix and $Q_5= Q_3^\top Q_3$. Then we have that
$$\lim_{\gamma\to 0}\|(Q_3^\top (Q_0Q_1Q_0+\gamma I_p) Q_3 + \lambda I_q)^{-1}-(Q_3^\top Q_0Q_1Q_0 Q_3 + \lambda I_q)^{-1}\|_{op}=0$$
which implies that for any vector $\bm{u}$,
$$\lim_{\gamma\to 0}\|(Q_3^\top (Q_0Q_1Q_0+\gamma I_p) Q_3 + \lambda I_q)^{-1}\bm{u}-(Q_3^\top Q_0Q_1Q_0 Q_3 + \lambda I_q)^{-1}\bm{u}\|_2=0$$
Hence we conclude our claim \eqref{equ:lemma3_1_1} using the sub-multiplicative property of the norm.

\textbf{Step 2:} We claim that for any vector $\bm{u}$,
\begin{align}
&\lim_{\gamma\to 0}\bm{u}^\top Q_0(Q_0 Q_1 Q_0+\gamma I_p)^{-1} Q_2 (Q_0 Q_1 Q_0+\gamma I_p)^{-1}Q_0\bm{u} \nonumber\\
= & \bm{u}^\top Q_0(Q_0 Q_1 Q_0)^{\dagger} Q_2 (Q_0 Q_1 Q_0)^{\dagger}Q_0\bm{u}    \label{equ:lemma3_2_1}
\end{align}

Since $Q_0$ is an orthogonal projection, let 
$Q_6DQ_6^{\top}$ be an eigendecomposition of $Q_0$ where $Q_6$ is the orthogonal matrix whose column is the eigenvector $\bm{q}_{0,j}$ of $Q_0$, and $D=\text{diag}\{1, 1, \cdots, 1, 0, \cdots, 0\}$ is the diagonal matrix whose diagonal elements are the corresponding eigenvalues. 
We write
$$D=\begin{bmatrix} I_r & 0\\ 0 & 0 \end{bmatrix}, \quad Q_6^\top Q_1 Q_6= \begin{bmatrix} Q_{7,1} & Q_{7,2}\\ Q_{7,3} & Q_{7,4} \end{bmatrix}, \quad Q_6^\top Q_2 Q_6= \begin{bmatrix} Q_{8,1} & Q_{8,2}\\ Q_{8,3} & Q_{8,4} \end{bmatrix},$$
where $r=\text{rank}(Q_0)$.
Then we have
$$Q_0Q_1Q_0=Q_6 \begin{bmatrix} I_r & 0\\ 0 & 0 \end{bmatrix} \begin{bmatrix} Q_{7,1} & Q_{7,2}\\ Q_{7,3} & Q_{7,4} \end{bmatrix} \begin{bmatrix} I_r & 0\\ 0 & 0 \end{bmatrix} Q_6^\top =Q_6 \begin{bmatrix} Q_{7,1} & 0\\ 0 & 0 \end{bmatrix} Q_6^\top.$$
Note that by our assumptions: 
$$r=\text{rank}(Q_0)=\text{rank}(Q_0Q_1Q_0)=\text{rank}\left(Q_6 \begin{bmatrix} Q_{7,1} & 0\\ 0 & 0 \end{bmatrix} Q_6^\top\right)=\text{rank}(Q_{7,1})$$
since $Q_6$ is an orthogonal matrix. This shows that
$Q_{7,1}$ must be invertible since $Q_{7,1}$ is an $r\times r$ matrix. Hence we have $Q_{7,1}^\dagger=Q_{7,1}^{-1}$, which implies that 
$$(Q_0 Q_1 Q_0)^{\dagger}=Q_6 \begin{bmatrix} Q^{-1}_{7,1} & 0\\ 0 & 0 \end{bmatrix} Q_6^\top.$$

Now we consider
$$(Q_0 Q_1 Q_0+\gamma I_p)^{-1}=Q_6 \begin{bmatrix} (Q_{7,1}+\gamma I_r)^{-1} & 0\\ 0 & \gamma^{-1} I_{p-r} \end{bmatrix} Q_6^\top$$
and thus
\begin{align*}
&\bm{u}^\top Q_0(Q_0 Q_1 Q_0+\gamma I_p)^{-1} Q_2 (Q_0 Q_1 Q_0+\gamma I_p)^{-1}Q_0\bm{u}\\
=&\bm{u}^\top Q_0Q_6\begin{bmatrix} (Q_{7,1}+\gamma I_r)^{-1} & 0\\ 0 & \gamma^{-1} I_{p-r} \end{bmatrix} \begin{bmatrix} Q_{8,1} & Q_{8,2}\\ Q_{8,3} & Q_{8,4} \end{bmatrix} \begin{bmatrix} (Q_{7,1}+\gamma I_r)^{-1} & 0\\ 0 & \gamma^{-1} I_{p-r} \end{bmatrix} Q_6^{\top} Q_0\bm{u}\\
=&\bm{u}^\top Q_0 Q_6\begin{bmatrix} (Q_{7,1}+\gamma I_r)^{-1} Q_{8,1} (Q_{7,1}+\gamma I_r)^{-1} & * \\ * & * \end{bmatrix} Q_6^{\top} Q_0 \bm{u}
\end{align*}
where $*$ represents the term that is not of interest, as we can show the last $p-r$ elements in $\bm{u}^\top  Q_0 Q_6$ must be 0, i.e., $\bm{u}^\top  Q_0 Q_6=(u^{(1)}, \cdots, u^{(r)}, 0, \cdots, 0)=(\tilde{\bm{u}}, 0)$, as follows: Recall that $Q_6=(\bm{q}_{0,1},\cdots,\bm{q}_{0,p})$ where $\bm{q}_{0,j}$ is the eigenvector of $Q_0$, and $Q_0\bm{q}_{0,j}=\bm{0}$ for all $j\ge r+1$ since its corresponding eigenvalue is $0$. Hence we conclude that
\begin{equation}\label{equ:lemma3_2_2}
\bm{u}^\top Q_0(Q_0 Q_1 Q_0+\gamma I_p)^{-1} Q_2 (Q_0 Q_1 Q_0+\gamma I_p)^{-1}Q_0\bm{u}=\tilde{\bm{u}}^\top (Q_{7,1}+\gamma I_r)^{-1} Q_{8,1} (Q_{7,1}+\gamma I_r)^{-1} \tilde{\bm{u}}    
\end{equation}
On the other hand, consider 
$$(Q_0 Q_1 Q_0)^{\dagger}=Q_6 \begin{bmatrix} Q^{-1}_{7,1} & 0\\ 0 & 0 \end{bmatrix} Q_6^\top.$$
Hence,
\begin{align*}
=&\bm{u}^\top Q_0(Q_0 Q_1 Q_0)^{\dagger} Q_2 (Q_0 Q_1 Q_0)^{\dagger}Q_0\bm{u}\\
=&\bm{u}^\top Q_0Q_6\begin{bmatrix} Q^{-1}_{7,1} & 0\\ 0 & 0 \end{bmatrix} \begin{bmatrix} Q_{8,1} & Q_{8,2}\\ Q_{8,3} & Q_{8,4} \end{bmatrix} \begin{bmatrix} Q^{-1}_{7,1} & 0\\ 0 & 0 \end{bmatrix} Q_6^{\top} Q_0\bm{u}\\
=&\bm{u}^\top Q_0Q_6\begin{bmatrix} Q^{-1}_{7,1} Q_{8,1} Q^{-1}_{7,1} & 0\\ 0 & 0 \end{bmatrix} Q_6^{\top} Q_0\bm{u}\\
=& \tilde{\bm{u}}^\top Q^{-1}_{7,1} Q_{8,1} Q^{-1}_{7,1} \tilde{\bm{u}}.
\end{align*}
Using again the notation $\bm{u}^\top  Q_0 Q_6=(u^{(1)}, \cdots, u^{(r)}, 0, \cdots, 0)=(\tilde{\bm{u}}, 0)$, we have
\begin{equation}\label{equ:lemma3_2_3}
\bm{u}^\top Q_0(Q_0 Q_1 Q_0)^{\dagger} Q_2 (Q_0 Q_1 Q_0)^{\dagger}Q_0\bm{u}=\tilde{\bm{u}}^\top Q^{-1}_{7,1} Q_{8,1} Q^{-1}_{7,1} \tilde{\bm{u}}.
\end{equation}
Comparing \eqref{equ:lemma3_2_2} and \eqref{equ:lemma3_2_3}, note that the inverses in \eqref{equ:lemma3_2_2} and \eqref{equ:lemma3_2_3} are both the standard inverse instead of the Moore-Penrose pseudoinverse. Therefore, by using the local Lipschitz continuity of the matrix inversion as we did in Step 1, we have
$$\lim_{\gamma\to 0}\tilde{\bm{u}}^\top (Q_{7,1}+\gamma I_r)^{-1} Q_{8,1} (Q_{7,1}+\gamma I_r)^{-1} \tilde{\bm{u}}=\tilde{\bm{u}}^\top Q^{-1}_{7,1} Q_{8,1} Q^{-1}_{7,1} \tilde{\bm{u}},$$
which gives our claim \eqref{equ:lemma3_2_1}. 

\textbf{Step 3:} Finally, we take $\gamma\to 0$ in both sides of \eqref{equ:lemma3_0} and use the above two claims \eqref{equ:lemma3_1_1} and \eqref{equ:lemma3_2_1}. We conclude that
\begin{align*}
& Q_0Q_3(Q_3^\top Q_0Q_1Q_0 Q_3 + \lambda I_q)^{-1}Q_3^\top Q_2 Q_3(Q_3^\top Q_0Q_1Q_0 Q_3 + \lambda I_q)^{-1}Q_3^\top Q_0   \\
& \le  Q_0(Q_0 Q_1 Q_0)^{\dagger} Q_2 (Q_0 Q_1 Q_0)^{\dagger}Q_0.
\end{align*}
\Halmos
\endproof

\proof{Proof of Theroem \ref{misspecified:cons}}
The proof is similar to Theorem \ref{misspecified}.
\Halmos
\endproof

\subsection{Proofs of Results in Section \ref{sec:cont}}

\proof{Proof of Theorem \ref{correctlyspecified:cont}}
In the well-specified case, it is easy to see that $\thete^*=\thete_0$ since $\thete_0$ indeed minimizes $\vale_0(\omege_{\thete})$. Therefore 
$\omege^*=\omege_{\thete_0}=\omege_{\thete^*}$. In addition,
$\thete^{KL}=\thete_0$ since $\thete_0$ indeed minimizes $KL(P, P_\thete(\ctext,\ran))$. Therefore 
$\omege^*=\omege_{\thete_0}=\omege_{\thete^{KL}}$. 

For ETO, we have $\hat\thete^{ETO}\xrightarrow{P}\thete^{KL}$ by Proposition \ref{ETOconsistency:cont} where $\thete^{KL}$ is defined in Assumption \ref{ETOconsistency:cont:assm}. Hence, we have that
$R(\hat \omege^{ETO})=R( \omege_{\hat \thete^{ETO}})=\vale_0(\omege_{\hat \thete^{ETO}})-\vale_0(\omege^*)\xrightarrow{P} \vale_0(\omege_{\thete^{KL}})-\vale_0(\omege^*)$ by the continuity of $\vale_0(\omege_\thete)$  and the continuous mapping theorem.

For IEO, note that $\hat\thete^{IEO}\xrightarrow{P}\thete^*$ by Proposition \ref{IEOconsistency:cont} where $\thete^*$ is defined in Assumption \ref{IEOconsistency:cont:assm}. Hence, we have that
$R(\hat \omege^{IEO})=R( \omege_{\hat \thete^{IEO}})=\vale_0(\omege_{\hat \thete^{IEO}})-\vale_0(\omege^*)\xrightarrow{P} \vale_0(\omege_{\thete^*})-\vale_0(\omege^*)$ by the continuity of $\vale_0(\omege_\thete)$ and the continuous mapping theorem.

Since $\omege^*=\omege_{\thete_0}=\omege_{\thete^*}=\omege_{\thete^{KL}}$, the conclusion of the theorem follows.
\Halmos
\endproof

\proof{Proof of Theorem \ref{SD3}}
In the following proof, we always write 
$$\bar{\vale}(\omege(\ctext),\thete|\ctext)= \vale(\omege(\ctext), \thete|\ctext)+ \sum_{j\in J} \alpha_j(\thete,\ctext) g_j(\omege(\ctext)),$$
$$\bar{\vale}_0(\omege(\ctext)|\ctext)= \vale_0(\omege(\ctext))+ \sum_{j\in J} \alpha^*_j(\ctext) g_j(\omege(\ctext)).$$ 
With the optimality of $\thete_0$ (Assumption \ref{RCforIEO:cont:assm}), we have that $\nabla_{\thete}\vale_0(\omege_{\thete_0})=0$ and thus
\begin{equation}
\vale_0(\omege_{\thete})-\vale_0(\omege^*)=\frac{1}{2}(\thete-\thete_0)^\top\nabla_{\thete\thete}\vale_0(\omege_{\thete_0})(\thete-\thete_0)+o(\|\thete-\thete_0\|_2^2).\label{excess risk1:cont}
\end{equation}
In particular, this equation holds for $\thete=\hat{\thete}^{ETO}$ and $\hat{\thete}^{IEO}$ (with $o$ replaced by $o_P$).

Before going into the comparison of the two approaches, we point out several facts. 

\textbf{Claim 1.} We claim that for any fixed $\ctext\in\mathcal{X}$, there exists a $\varepsilon(\ctext)>0$ such that
\begin{equation}\label{secondobservation:cont}
\alpha^*_j(\ctext) (g_j(\omege^*(\ctext))-g_j(\omege_\thete(\ctext)))=0, \quad \forall j\in J   
\end{equation}
for any $\thete\in \{\thete\in\Theta:\|\thete-\thete_0\|_2\le \varepsilon(\ctext)\}$. The proof is similar to the proof of Claim 1 in Theorem \ref{SD2} by using the continuity of $\alpha_j(\thete,\ctext)$ with respect to $\thete$ and the KKT conditions in Assumption \ref{OCforZ:cont}. 

\textbf{Claim 2.} We claim that for any fixed $\ctext\in\mathcal{X}$, 
\begin{equation}\label{thirdobservation:cont}
\alpha^*_j(\ctext) \nabla_{\thete} g_j(\omege_{\thete_0}(\ctext))=\alpha^*_j(\ctext) \nabla_{\omege} g_j(\omege^*(\ctext))\nabla_\thete \omege_{\thete_0}(\ctext)=0, \quad \forall j\in J,  
\end{equation}
\begin{equation}\label{fourthobservation:cont}
 \alpha^*_j(\ctext) \nabla_{\thete\thete} g_j(\omege_{\thete_0}(\ctext))=0, \quad \forall j\in J,   
\end{equation}
\begin{equation}\label{fifthobservation:cont}
\nabla_{\thete} g_j(\omege_{\thete_0}(\ctext))= \nabla_{\omege} g_j(\omege^*(\ctext))\nabla_\thete \omege_{\thete_0}(\ctext)=0, \quad \forall j\in B(\ctext).  
\end{equation}
where $B(\ctext):=\{j \in J: g_j(\omege^*(\ctext)) = 0\}$.

First, Equalities \eqref{thirdobservation:cont} and \eqref{fourthobservation:cont} follow from \eqref{secondobservation:cont} because \eqref{secondobservation:cont} holds for any $\thete\in \{\thete\in\Theta:\|\thete-\thete_0\|_2\le \varepsilon(\ctext)\}$ implying that $\nabla_{\thete} \alpha^*_j(\ctext) (g_j(\omege^*(\ctext))-g_j(\omege_\thete(\ctext)))|_{\thete=\thete_0}=0$ and $\nabla_{\thete\thete} \alpha^*_j(\ctext) (g_j(\omege^*(\ctext))-g_j(\omege_\thete(\ctext)))|_{\thete=\thete_0}=0$ for all $j\in J$. 

Second,
for any $j\in B(\ctext)\cap J_2=J_2$ (the set of all equality constraints), we have that $g_j(\omege_{\thete}(\ctext))=0$ for all $\thete$, which clearly implies that 
$$\nabla_{\thete} g_j(\omege_{\thete_0}(\ctext))= \nabla_{\omege} g_j(\omege^*(\ctext))\nabla_\thete \omege_{\thete_0}(\ctext)=0, \quad j\in B(\ctext)\cap J_2.$$
In addition, for any $j\in B(\ctext)\cap J_1$ (the set of all active inequality constraints), since $g_j(\omege_{\thete_0}(\ctext))=0$ while $g_j(\omege_{\thete}(\ctext))\le 0$ for all $\thete$, $\thete_0$ (which is an inner point in $\Theta$) is a point of maximum for the function $g_j(\omege_{\thete}(\ctext))$, and thus
$$\nabla_{\thete} g_j(\omege_{\thete_0}(\ctext))= \nabla_{\omege} g_j(\omege^*(\ctext))\nabla_\thete \omege_{\thete_0}(\ctext)=0, \quad j\in B(\ctext)\cap J_1.$$
Equality \eqref{fifthobservation:cont} then follows.


\textbf{Claim 3.} 
We claim that for any fixed $\ctext\in\mathcal{X}$,
\begin{equation}
\nabla_{\omege\omege}\bar{\vale}(\omege_\thete(\ctext),\thete|\ctext)\nabla_{\thete} \omege_{\thete}(\ctext)+
\left(\nabla_{\omege\thete}\vale(\omege_\thete(\ctext),\thete|\ctext)+ \sum_{j\in J} \nabla_{\omege} g_j(\omege_\thete(\ctext))^\top \nabla_{\thete} \alpha_j(\thete,\ctext) \right)|_{\thete=\thete_0}=0.
\label{interim:cont}
\end{equation} 
and
\begin{equation}\label{interim:consFact2:cont}
\nabla_{\thete} \omege_{\thete_0}(\ctext)^\top \nabla_{\omege\omege}\bar{\vale}(\omege^*(\ctext),\thete_0|\ctext)\nabla_{\thete} \omege_{\thete_0}(\ctext)+\nabla_{\thete} \omege_{\thete_0}(\ctext)^\top
\nabla_{\omege\thete}\vale(\omege^*(\ctext),\thete_0|\ctext)=0.
\end{equation} 

For \eqref{interim:cont}, the reason is similar to \eqref{interim:cons}. For \eqref{interim:consFact2:cont},
first, 
\eqref{interim:cont} implies that
\begin{align}
&\nabla_{\thete} \omege_{\thete_0}(\ctext)^\top \nabla_{\omege\omege}\bar{\vale}(\omege^*(\ctext),\thete_0|\ctext)\nabla_{\thete} \omege_{\thete_0}(\ctext)\nonumber\\
&+\nabla_{\thete} \omege_{\thete_0}(\ctext)^\top
\left(\nabla_{\omege\thete}\vale(\omege^*(\ctext),\thete_0|\ctext)+ \sum_{j\in J} \nabla_{\omege} g_j(\omege^*(\ctext))^\top \nabla_{\thete} \alpha_j(\thete_0,\ctext) \right)=0.   \label{interim:consFact1:cont}  
\end{align}
When $j\in B(\ctext)$, \eqref{fifthobservation} shows that
$$\nabla_{\thete}\omege_{\thete_0}(\ctext)^\top \nabla_{\omege} g_j(\omege^*(\ctext))^\top = (\nabla_{\omege} g_j(\omege^*(\ctext))\nabla_{\thete}\omege_{\thete_0}(\ctext))^\top=0$$
When $j\notin B(\ctext)$, that is, $g_j(\omege^*(\ctext))=g_j(\omege_{\thete_0}(\ctext))<0$, the continuity implies that there exists a $\varepsilon_j(\ctext)>0$ such that $g_j(\omege_\thete(\ctext))<0$
for any $\thete\in \{\thete\in\Theta:\|\thete-\thete_0\|_2\le \varepsilon_j(\ctext)\}$. Hence complementary slackness in Assumption \ref{OCforZ:cont} implies that $\alpha_j(\thete,\ctext)=0$ for any $\thete\in \{\thete\in\Theta:\|\thete-\thete_0\|_2\le \varepsilon_j(\ctext)\}$, which shows that $\nabla_{\thete} \alpha_j(\thete_0,\ctext)=0$. 

Hence we conclude that for any $j\in J$,
$$\nabla_{\thete}\omege_{\thete_0}(\ctext)^\top \nabla_{\omege} g_j(\omege^*(\ctext))^\top\nabla_{\thete} \alpha_j(\thete_0,\ctext)=0.$$
Therefore \eqref{interim:consFact1:cont} implies \eqref{interim:consFact2:cont}.

\textbf{Claim 4.} We have that for any fixed $\ctext\in\mathcal{X}$,
\begin{equation} \label{equ:Hessian:cont}
\nabla_{\thete\thete}\vale_0(\omege_{\thete_0}(\ctext)|\ctext)=\nabla_\thete \omege_{\thete_0}(\ctext)^\top \nabla_{\omege\omege}\bar{\vale}_0(\omege^*(\ctext)|\ctext)\nabla_\thete \omege_{\thete_0}(\ctext).    
\end{equation}
In fact, we have that
\begin{align*}
&\nabla_{\thete\thete}\vale_0(\omege_{\thete}(\ctext)|\ctext)|_{\thete=\thete_0}\\
=& \nabla_{\thete} (\nabla_{\omege} \vale_0(\omege_\thete(\ctext)) \nabla_\thete \omege_{\thete}(\ctext))|_{\thete=\thete_0}\\
=&\nabla_\thete \omege_{\thete}(\ctext)^\top \nabla_{\omege\omege} \vale_0(\omege_\thete(\ctext)|\ctext) \nabla_\thete \omege_{\thete}(\ctext)+ \nabla_{\omege} \vale_0(\omege_\thete(\ctext)|\ctext) \nabla_{\thete} ( \nabla_\thete \omege_{\thete}(\ctext))|_{\thete=\thete_0} \\
=&\nabla_\thete \omege_{\thete}(\ctext)^\top \nabla_{\omege\omege} \vale_0(\omege_\thete(\ctext)|\ctext) \nabla_\thete \omege_{\thete}(\ctext)- \sum_{j\in J} \alpha^*_j(\ctext) \nabla_{\omege} g_j(\omege_\thete(\ctext)) \nabla_{\thete} ( \nabla_\thete \omege_{\thete}(\ctext))|_{\thete=\thete_0} \text{ by Assumption } \ref{OCforZ:cont}\\
=& \nabla_\thete \omege_{\thete}(\ctext)^\top \nabla_{\omege\omege} \vale_0(\omege_\thete(\ctext)|\ctext) \nabla_\thete \omege_{\thete}(\ctext)+ \sum_{j\in J} \alpha^*_j(\ctext) \nabla_\thete \omege_{\thete}(\ctext)^\top \nabla_{\omege\omege} g_j(\omege_\thete(\ctext)) \nabla_\thete \omege_{\thete}(\ctext)|_{\thete=\thete_0} \text{ by } \eqref{fourthobservation:cont} \\
=& \nabla_\thete \omege_{\thete_0}(\ctext)^\top \nabla_{\omege\omege}\bar{\vale}_0(\omege^*(\ctext)|\ctext)\nabla_\thete \omege_{\thete_0}(\ctext).
\end{align*}

\textbf{Step 1:} We show that $$\mathbb G^{ETO}\preceq_{st}\mathbb G^{IEO}.$$

To show this, we compare the performance of ETO and IEO at the level of $\thete$ and then leverage Equation \eqref{excess risk1:cont}. 
Note that ETO can be equivalently written as
$$\hat\omege^{ETO} =\min_{\omege\in\Omega} \vale(\omege,\hat{\thete}^{ETO})=\omege_{\hat{\thete}^{ETO}}$$
using the oracle problem \eqref{equ:oracle:cont} by plugging in the MLE $\hat{\thete}^{ETO}$.

For ETO, Proposition \ref{RCforETO:cont} gives that  
\begin{equation} \label{equ:MLE:cont}
\sqrt n(\hat{\thete}^{ETO}-\thete_0)\xrightarrow{d} N(0,\mathcal{I}_{\thete_0}^{-1}).
\end{equation}

Plugging \eqref{equ:MLE:cont} into \eqref{excess risk1:cont}, we have
$$n(\vale_0(\hat\omege^{ETO})-\vale_0(\omege^*))\xrightarrow{d}\mathbb G^{ETO}$$
where $\mathbb G^{ETO}=\frac{1}{2}{\mathcal N_1^{ETO}}^\top\nabla_{\thete\thete}\vale_0(\omege_{\thete_0})\mathcal N_1^{ETO}$ and 
\begin{equation} \label{equ:ETO1:cont}
\mathcal N_1^{ETO}\sim N(0,\mathcal{I}_{\thete_0}^{-1}).    
\end{equation}


For IEO, Proposition \ref{RCforIEO:cont} gives that
\begin{equation} \label{Mestimator:cont}
\sqrt{n}(\hat{\thete}^{IEO}-\thete_0)\xrightarrow{d} N(0,\nabla_{\thete\thete}\vale_0(\omege_{\thete_0})^{-1} Var_{P}(\nabla_\thete \cost(\omege_{\thete_0}(\ctext),\ran))\nabla_{\thete\thete}\vale_0(\omege_{\thete_0})^{-1}).    
\end{equation}

Plugging \eqref{Mestimator:cont} into \eqref{excess risk1:cont}, we have
$$n(\vale_0(\hat\omege^{IEO})-\vale_0(\omege^*)) \xrightarrow{d}\mathbb G^{IEO}$$
where $\mathbb G^{IEO}=\frac{1}{2}{\mathcal N_1^{IEO}}^\top\nabla_{\thete\thete}\vale_0(\omege_{\thete_0})\mathcal N_1^{IEO}$ and 
\begin{equation} \label{equ:IEO1:cont}
\mathcal N_1^{IEO}\sim N(0,\nabla_{\thete\thete}\vale_0(\omege_{\thete_0})^{-1} Var_{P}(\nabla_\thete \cost(\omege_{\thete_0}(\ctext),\ran))\nabla_{\thete\thete}\vale_0(\omege_{\thete_0})^{-1}).    
\end{equation}

We assert that 
\begin{equation}\label{CR2:cont}
\nabla_{\thete\thete}\vale_0(\omege_{\thete_0})^{-1} Var_{P}(\nabla_\thete \cost(\omege_{\thete_0}(\ctext),\ran))\nabla_{\thete\thete}\vale_0(\omege_{\thete_0})^{-1}\ge \mathcal{I}_{\thete_0}^{-1}.    
\end{equation}
where $\mathcal{I}_{\thete_0}$ is given in Proposition \ref{RCforETO:cont}.
We shall use the multivariate Cramer-Rao bound at each $\ctext$ and then aggregate them to prove the above inequality. 

Recall that $P(\ran|\ctext)=P_{\thete_0}(\ran|\ctext)$ and $\omege^*(\ctext)=\omege_{\thete_0}(\ctext)$ for any fixed $\ctext\in\mathcal{X}$.
For any fixed $\ctext\in\mathcal{X}$, consider a random vector $\ran\sim P_\thete(\ran|\ctext)$ and an estimator with the following form  $\nabla_\omege \cost(\omege^*(\ctext),\ran)$, which has the expectation $\mathbb{E}_\thete[\nabla_\omege \cost(\omege^*(\ctext),\ran)|\ctext]$. It follows from Assumption \ref{SCforh:cont} that
$\mathbb{E}_\thete[\nabla_\omege \cost(\omege^*(\ctext),\ran)|\ctext] = \nabla_\omege \mathbb{E}_\thete [\cost(\omege,\ran)|\ctext]|_{\omege=\omege^*(\ctext)} = \nabla_\omege \vale(\omege^*(\ctext), \thete|\ctext)$. 
Therefore we have that 
$$\nabla_\thete (\mathbb{E}_\thete[\nabla_\omege \cost(\omege^*(\ctext),\ran)|\ctext])^\top |_{\thete=\thete_0} = \nabla_\thete (\nabla_\omege \vale(\omege^*(\ctext), \thete|\ctext))^\top |_{\thete=\thete_0}= \nabla_{\omege\thete} \vale(\omege^*(\ctext), \thete_0|\ctext).$$
Applying Cramer-Rao bound on the estimator $ \nabla_\omege \cost(\omege^*(\ctext),\ran)$ (assured by Assumption \ref{SCforh:cont}), we have that
\begin{equation}\label{CR:cont}
Var_{P(\ran|\ctext)}\Big(\nabla_\omege \cost(\omege^*(\ctext),\ran)\Big) \geq  \nabla_{\omege\thete}\vale_0(\omege^*(\ctext), \thete_0|\ctext) \mathcal{I}_{\thete_0}(\ctext)^{-1}\nabla_{\thete\omege}\vale_0(\omege^*(\ctext), \thete_0|\ctext).    
\end{equation}
where
$$\mathcal{I}_{\thete_0}(\ctext)=\mathbb{E}_{P(\ran|\ctext)}[(\nabla_\thete \log p_{\thete^{KL}}(\ran|\ctext))^\top \nabla_\thete \log p_{\thete^{KL}}(\ran|\ctext) ].$$
Note that $\mathbb{E}_{P(\ctext)}[\mathcal{I}_{\thete_0}(\ctext)]=\mathcal{I}_{\thete_0}$ in Proposition \ref{RCforETO:cont}.

We can then show that
\begin{align*}
&Var_{P(\ran|\ctext)}(\nabla_\thete \cost(\omege_{\thete_0}(\ctext),\ran)) \\
=&Var_{P(\ran|\ctext)}(\nabla_\omege \cost(\omege^*(\ctext),\ran)\nabla_\thete \omege_{\thete_0}(\ctext)) \\
=&\nabla_\thete \omege_{\thete_0}(\ctext)^\top Var_{P(\ran|\ctext)}(\nabla_\omege \cost(\omege^*(\ctext),\ran)) \nabla_\thete \omege_{\thete_0}(\ctext) \quad \text{ since $\nabla_\thete \omege_{\thete_0}(\ctext)$ is deterministic}\\
\geq & \nabla_\thete \omege_{\thete_0}(\ctext)^\top \nabla_{\omege\thete}\vale_0(\omege^*(\ctext), \thete_0|\ctext) \mathcal{I}_{\thete_0}(\ctext)^{-1}\nabla_{\thete\omege}\vale_0(\omege^*(\ctext), \thete_0|\ctext) \nabla_\thete \omege_{\thete_0}(\ctext) \quad\text{ by } \eqref{CR:cont} \\
=  &\nabla_\thete \omege_{\thete_0}(\ctext)^\top\nabla_{\omege\omege} \bar{\vale}_0(\omege^*(\ctext)|\ctext)\nabla_\thete \omege_{\thete_0}(\ctext) \mathcal{I}_{\thete_0}(\ctext)^{-1}  \nabla_\thete \omege_{\thete_0}(\ctext)^\top\nabla_{\omege\omege} \bar{\vale}_0(\omege^*(\ctext)|\ctext)\nabla_\thete \omege_{\thete_0}(\ctext)  \quad\text{ by } \eqref{interim:consFact2:cont} \\
= & \nabla_{\thete\thete}\vale_0(\omege_{\thete_0}(\ctext)|\ctext) \mathcal{I}_{\thete_0}(\ctext)^{-1} \nabla_{\thete\thete}\vale_0(\omege_{\thete_0}(\ctext)|\ctext) \quad\text{ by } \eqref{equ:Hessian:cont}.
\end{align*}
By taking the expectation over $P(\ctext)$, we have that
\begin{align*}
&Var_{P}(\nabla_\thete \cost(\omege_{\thete_0}(\ctext),\ran))\\ =& \mathbb{E}_{P(\ctext)}[Var_{P(\ran|\ctext)}(\nabla_\thete \cost(\omege_{\thete_0}(\ctext),\ran))]+Var_{P(\ctext)}(\mathbb{E}_{P(\ran|\ctext)}[\nabla_\thete \cost(\omege_{\thete_0}(\ctext),\ran)])\\  
\geq  &\mathbb{E}_{P(\ctext)}[Var_{P(\ran|\ctext)}(\nabla_\thete \cost(\omege_{\thete_0}(\ctext),\ran))]\\
\geq  &\mathbb{E}_{P(\ctext)}[\nabla_{\thete\thete}\vale_0(\omege_{\thete_0}(\ctext)|\ctext) \mathcal{I}_{\thete_0}(\ctext)^{-1}\nabla_{\thete\thete}\vale_0(\omege_{\thete_0}(\ctext)|\ctext)]\\
\geq  &\mathbb{E}_{P(\ctext)}[\nabla_{\thete\thete}\vale_0(\omege_{\thete_0}(\ctext)|\ctext)] \mathbb{E}_{P(\ctext)}[\mathcal{I}_{\thete_0}(\ctext)]^{-1}\mathbb{E}_{P(\ctext)}[\nabla_{\thete\thete}\vale_0(\omege_{\thete_0}(\ctext)|\ctext)]\\
=  &\nabla_{\thete\thete}\vale_0(\omege_{\thete_0}) \mathcal{I}_{\thete_0}^{-1}\nabla_{\thete\thete}\vale_0(\omege_{\thete_0})
\end{align*}
where the last inequality follows from the matrix extension of the Cauchy-Schwarz inequality (Lemma 2 in \citet{lavergne2008cauchy}, which is also stated in Lemma \ref{lemma:Cauchy-Schwarz} in Section \ref{sec:multivariateCramerRao}): Noting that $\mathcal{I}_{\thete_0}(\ctext)$ is a positive definite matrix, its square root $\mathcal{I}_{\thete_0}(\ctext)^\frac{1}{2}$ exists and thus we use Lemma \ref{lemma:Cauchy-Schwarz} to obtain that
\begin{align*}
&\mathbb{E}_{P(\ctext)}[\left(\nabla_{\thete\thete}\vale_0(\omege_{\thete_0}(\ctext)|\ctext) \mathcal{I}_{\thete_0}(\ctext)^{-\frac{1}{2}}\right) 
\left(\mathcal{I}_{\thete_0}(\ctext)^{-\frac{1}{2}}\nabla_{\thete\thete}\vale_0(\omege_{\thete_0}(\ctext)|\ctext)\right)]\\
\geq  &\mathbb{E}_{P(\ctext)}\left[\left(\nabla_{\thete\thete}\vale_0(\omege_{\thete_0}(\ctext)|\ctext) \mathcal{I}_{\thete_0}(\ctext)^{-\frac{1}{2}}\right) \mathcal{I}_{\thete_0}(\ctext)^{\frac{1}{2}} \right] \mathbb{E}_{P(\ctext)}\left[ \mathcal{I}_{\thete_0}(\ctext)^{\frac{1}{2}}  \mathcal{I}_{\thete_0}(\ctext)^{\frac{1}{2}} \right]^{-1}\\
&\mathbb{E}_{P(\ctext)}\left[ \mathcal{I}_{\thete_0}(\ctext)^{\frac{1}{2}} \left(\mathcal{I}_{\thete_0}(\ctext)^{-\frac{1}{2}}\nabla_{\thete\thete}\vale_0(\omege_{\thete_0}(\ctext)|\ctext) \right)\right]\\
= &\mathbb{E}_{P(\ctext)}[\nabla_{\thete\thete}\vale_0(\omege_{\thete_0}(\ctext)|\ctext)] \mathbb{E}_{P(\ctext)}[\mathcal{I}_{\thete_0}(\ctext)]^{-1}\mathbb{E}_{P(\ctext)}[\nabla_{\thete\thete}\vale_0(\omege_{\thete_0}(\ctext)|\ctext)]
\end{align*}

Therefore our target \eqref{CR2:cont} follows.
Hence, Comparing ETO \eqref{equ:ETO1:cont} and IEO \eqref{equ:IEO1:cont} by using Lemma \ref{lemma1}, we conclude that 
$$\mathbb G^{ETO}\preceq_{st}\mathbb G^{IEO}.$$
\Halmos
\endproof

\proof{Proof of Theorem \ref{misspecified:cont}}



For ETO, we have $\hat\thete^{ETO}\xrightarrow{P}\thete^{KL}$ by Proposition \ref{ETOconsistency:cont} where $\thete^{KL}$ is defined in Assumption \ref{ETOconsistency:cont:assm}. Hence, we have that
$R(\hat \omege^{ETO})=R( \omege_{\hat \thete^{ETO}})=\vale_0(\omege_{\hat \thete^{ETO}})-\vale_0(\omege^*)\xrightarrow{P} \vale_0(\omege_{\thete^{KL}})-\vale_0(\omege^*)$ by the continuity of $\vale_0(\omege_\thete)$  and the continuous mapping theorem.

For IEO, note that $\hat\thete^{IEO}\xrightarrow{P}\thete^*$ by Proposition \ref{IEOconsistency:cont} where $\thete^*$ is defined in Assumption \ref{IEOconsistency:cont:assm}. Hence, we have that
$R(\hat \omege^{IEO})=R( \omege_{\hat \thete^{IEO}})=\vale_0(\omege_{\hat \thete^{IEO}})-\vale_0(\omege^*)\xrightarrow{P} \vale_0(\omege_{\thete^*})-\vale_0(\omege^*)$ by the continuity of $\vale_0(\omege_\thete)$ and the continuous mapping theorem.

Comparing ETO and IEO, we must have $\vale_0(\omege_{\thete^{KL}})\geq \vale_0(\omege_{\thete^*})$ by the definition of $\thete^*$ in Assumption \ref{IEOconsistency:cont:assm}. Thus the conclusion of the theorem follows.
\Halmos
\endproof

%% file: appendix_experiments.tex
\section{Additional Experiment Details}
\label{sec:appendix_experiment}

\subsection{Details for the Newsvendor Problems}
\subsubsection{Algorithms for the multi-product newsvendor problem.}\label{sec:appendix_newsvendor_unconstrained}

Recall the multi-product newsvendor objective is 
\begin{align*}
\min_{\omege}
\E_{P}\left[ \bm h^\top (\omege-\ran)^+ + \bm b^\top (\ran-\omege)^+\right].
\end{align*}
Recall we assume that demand for each product $j$ is independent and has distribution $\mathcal{N}(j\theteun,\sigma_j^2)$ with known $\sigma_j$ and an unknown parameter $\theteun\in\mathbb{R}$ that we want to learn. It is well known that the best decision to make is $\omege_\theteun=(\omegeun_\theteun^{(1)},\cdots,\omegeun_\theteun^{(p)})^\top$ where $$\omegeun_\theteun^{(j)}=j\theteun+\sigma_j \Phi_{normal}^{-1}\left(\frac{b^{(j)}}{b^{(j)}+h^{(j)}}\right)$$
for each product $j$ \citep{turken2012multi}. 

{\bf SAA.} For SAA, we solve the following linear optimization problem.
\begin{align*}
\min_{\omege , \bm u, \bm v}  &\sum_{i=1}^n \bm h^\top \bm u_i + \bm b^\top \bm v_i\\
\text{s.t.} &\ \bm u_i\geq \vec{0}, \ \bm u_{i} \geq \omege - \ran_{i}, \quad\forall i\\
&\ \bm v_i\geq \vec{0}, \ \bm v_i \geq \ran_i - \omege, \quad\forall i, 
\end{align*}
where $\ge$ between two vectors denotes element-wise greater than or equal to. 

{\bf IEO.} For IEO, we solve the following linear optimization problem.
\begin{align*}
\min_{\thete, \bm u, \bm v}  &\sum_{i=1}^n  \bm h^\top \bm u_i + \bm b^\top \bm v_i\\
\text{s.t.} &\ \bm u_i\geq \vec{0}, \ \bm u_i \geq \omege_\thete - \ran_i,\quad\forall i\\
&\ \bm v_i\geq \vec{0}, \ \bm v_i \geq \ran_i - \omege_\thete,\quad\forall i.
\end{align*}

{\bf ETO.} For ETO, we first compute the maximum likelihood estimator (MLE) $\hat\thete^{ETO}$ for the unknown mean in the Gaussian model. Once we have the MLE estimator, we use the Gaussian model to compute the decision $\hat{\omege}^{ETO}=\omege_{\hat\thete^{ETO}}$. Now we derive the MLE. The joint likelihood function is 
$$\prod_{i=1}^n \prod_{j=1}^p f(\ran_{i}^{(j)}) = \prod_{j=1}^p \left[ \left(\frac{1}{\sqrt{2\pi}}\right)^n e^{-\frac{1}{2}\sum_{i=1}^n (\ran_i^{(j)}-j\theteun)^2}\right].$$
Thus, the log-likelihood is 
$$ -\frac{1}{2} \sum_{i=1}^n \sum_{j=1}^p(\ran_i^{(j)}-j\theteun)^2 +\text{constant}.$$
Consequently, the MLE is 
$$\hat\theteun^{ETO} = \frac{\sum_{j=1}^p\sum_{i=1}^n \ran_{i}^{(j)}}{n\sum_{j=1}^p j}.$$



\subsubsection{Algorithms for the multi-product newsvendor problem with a single capacity constraint.}

Recall that the multi-product newsvendor problem with a capacity constraint is
\begin{align*}
\min_{\omege}
\E_{P}\left[ \bm h^\top(\omege-\ran)^+ + \bm b^\top(\ran-\omege)^+\right], 
\quad\text{s.t. } \sum_{j=1}^p \omegeun^{(j)} \le C.
\end{align*}

Recall we assume that demand for each product $j$ is independent and has distribution $\mathcal{N}(j\theteun,\sigma_j^2)$ with known $\sigma_j$ and an unknown parameter $\theteun\in\mathbb{R}$ that we want to learn.

The best decision to make is $w_\theteun$, which is given by Algorithm~\ref{alg:binarySearch}. This algorithm is modified from \citet{zhang2009binary}, and the correctness of the algorithm follows from Lemma 1 and Proposition 1 in \citet{zhang2009binary}.

\begin{algorithm}[!htb]
\caption{Binary Search Algorithm}
\begin{algorithmic}[1]
\State Input tolerance $\epsilon$, backing order cost $b$, holding cost $h$, and cdf $F_j(\cdot)$ for each product $j\in[p]$. Input the capacity parameter $C$. 
\State Solve $\omege^*$ for the unconstrained problem. 
\State \textbf{if} Constraints is satisfied at $\omege^*$ \textbf{then}
\State \quad Output solution $\omege^*$. 
\State \textbf{end if}
\State Set $r_L = -b$ and $r_U=0$.
\State \textbf{while} $r_U - r_L > \epsilon$ \textbf{do}
\State \quad Set $r = \frac{r_U+r_L}{2}$.
\State \quad \textbf{for} {j$\in$[p]} \textbf{do}
\State \quad \quad	\textbf{if} $\frac{r+b}{h+b} > F_j(0)$ \textbf{then}
\State \quad \quad \quad Set $\omege^{(j)}=F_j^{-1}\left(\frac{r+b}{h+b}\right)$.
\State \quad \quad	\textbf{else} 
\State \quad \quad	\quad Set $\omege^{(j)}=0$.
\State \quad \quad	\textbf{end if} 
\State \quad	\textbf{end for}
\State \quad 	\textbf{if} $\sum_{j=1}^p \omege^{(j)} < C$ \textbf{then}
\State \quad  \quad Let $r_L=r$.
\State \quad 	\textbf{elseif} $\sum_{j=1}^p \omege^{(j)} > C$ \textbf{then}
\State \quad  \quad Let $r_U=r$.
\State \quad 	\textbf{else} 
\State \quad  \quad Output solution $\omege$.
\State \quad 	\textbf{end if } 
\State 	\textbf{end while}
\end{algorithmic}
\label{alg:binarySearch}
\end{algorithm}


{\bf SAA.} For SAA, we solve the following linear optimization problem. 

\begin{align*}
\min_{\omege, \bm u, \bm v}  &\sum_{i=1}^n \bm h^\top \bm u_i + \bm b^\top \bm v_i\\
\text{s.t.} &\ \bm u_i\geq \vec{0}, \ \bm u_i \geq \omege - \ran_i, \quad\forall i\\
&\ \bm v_i\geq \vec{0}, \ \bm v_i \geq \ran_i - \omege, \quad\forall i \\
& \sum_{j=1}^d \omegeun^{(j)} \le C
\end{align*}

{\bf IEO.} For IEO, we solve the following problem, where $\omege_{\thete}$ is computed using Algorithm~\ref{alg:binarySearch}. 


\begin{align*}
\min_{\thete} & 
\sum_{i=1}^n  \left[ \bm h^\top(\omege_\theteun-\ran_i)^+ + \bm b^\top(\ran_i - \omege_\theteun)^+\right] 
\end{align*}
Notice we only have one variable to optimize over. We use grid search to find an approximate optimal solution. We search from $\theteun \in [2,4]$ in increments of 0.01.

{\bf ETO.} 
For ETO, we first compute the maximum likelihood estimator (MLE) $\hat\theteun^{ETO}$ for the unknown mean in the Gaussian model. Recall from Section~\ref{sec:appendix_newsvendor_unconstrained} that the MLE is 
$$\hat\theteun^{ETO} = \frac{\sum_{j=1}^p\sum_{i=1}^n \ran_{i}^{(j)}}{n\sum_{j=1}^p j}.$$
Once we have the MLE estimator, we compute the decision $\omege_{\thete}$ using Algorithm~\ref{alg:binarySearch}


\subsubsection{Algorithms for the feature-based newsvendor problem.}



Recall that the feature-based multi-product newsvendor problem is
$$
\min_{\omege(\cdot)} \E_P \left[(\bm h(\omege(\ctext)-\ran)^+ + \bm b(\ran-\omege(\ctext))^+\right],
$$
where $\omege(\cdot)$ maps a feature $\ctext$ to a decision (order quantity). 

We assume the demand distribution is $\mathcal{N}((1,\ctext^\top)\thete,1)$, where $\thete\in\mathbb{R}^3$ are unknown parameters that we want to learn. The best decision to make is $\omege_\thete(\ctext) = (1,\ctext^\top)\thete + \sigma \Phi_{normal}^{-1}\left(\frac{b}{b + h}\right)$.

{\bf IEO.} For IEO, we solve the following optimization problem. 
\begin{align*}
\min_{\thete} \frac{1}{n}\sum_{i=1}^n \left(h(\omege_\thete(\ctext_i)-\ran_i)^+ + b(\ran_i-\omege_\thete(\ctext_i))^+\right),
\end{align*} 

When the model is Gaussian, the problem above is equivalent to 

\begin{align*}
\min_{\bm \thete, u,v}  &\sum_{i=1}^n  h u_{i}+b v_{i}\\
\text{s.t.} &\ u_{i}\geq 0, \ u_{i} \geq \left((1,\ctext_i^\top)\thete +\sigma\Phi^{-1}\left(\frac{b}{b+h}\right)\right) - \ran_{i}, \quad\forall i\\
&\ v_{i}\geq 0, \ v_{i} \geq \ran_{i} - \left((1,\ctext_i^\top)\thete + \sigma \Phi^{-1}\left(\frac{b}{b + h}\right)\right), \quad\forall i.
\end{align*}

{\bf ETO.} For ETO, we first compute the maximum likelihood estimator (MLE) $\hat\thete^{ETO}$. Once we have the MLE estimator, we use the Gaussian model to compute the decision $\hat{\omege}^{ETO}=\omege_{\hat\thete^{ETO}}$. Now we derive the MLE. The joint likelihood function is 
$$\prod_{i=1}^n f(\ran_{i}) =\left[ \left(\frac{1}{\sqrt{2\pi}}\right)^n e^{-\frac{1}{2}\sum_{i=1}^n (\ran_i-(1,\ctext_i))^\top\thete)^2}\right].$$
Thus, the log-likelihood is 
$$ -\frac{1}{2}\sum_{i=1}^n (\ran_i-(1,\ctext_i)^\top\thete)^2 +\text{constant}.$$
Consequently, we obtain the MLE by solving 
$$\hat\thete^{ETO} \in \argmax_{\thete} -\frac{1}{2}\sum_{i=1}^n (\ran_i-(1,\ctext_i)^\top\thete)^2.$$

\subsubsection{Runtime Analysis.}\label{sec:runtime}

In Table~\ref{table:newsvendor_runtime}, we provide the runtimes of the three compared methods. The runtimes are measured in the unconstrained case with varying sample sizes, on an Intel I5-13500 CPU. We observe that ETO is substantially faster than IEO and SAA. This is because the MLE can be computed in closed form. In contrast, both SAA and IEO require solving a linear optimization problem. While this is only one of the settings that we consider in our numerical investigation, it showcases the general expected phenomenon that ETO is likely faster than IEO, and in the case that a fast closed-form formula is present to compute the model-based decision oracle, it is also faster than SAA.

\begin{table}[h!]
\centering
    \begin{tabular}{@{}lcccc@{}}
        \toprule
        $n$ & 50 & 100 & 200 & 400 \\ \midrule
        SAA       & 0.172 &  0.305 & 0.634 & 1.575\\
        IEO   & 0.146 & 0.280 &  0.561 & 1.344\\
        ETO & 0.001 & 0.002 & 0.004 & 0.010 \\
        \midrule
    \end{tabular}
    \caption{Runtime in seconds, averaged over 10 runs. The runtime is measured in the unconstrained case.} 
    \label{table:newsvendor_runtime}
\end{table}

\subsubsection{Further Results Regarding Dimensionality of Decisions and Model Parameters.}\label{sec:more dim}
\begin{figure}[ht]
\centering
        \includegraphics[width=0.38\textwidth]{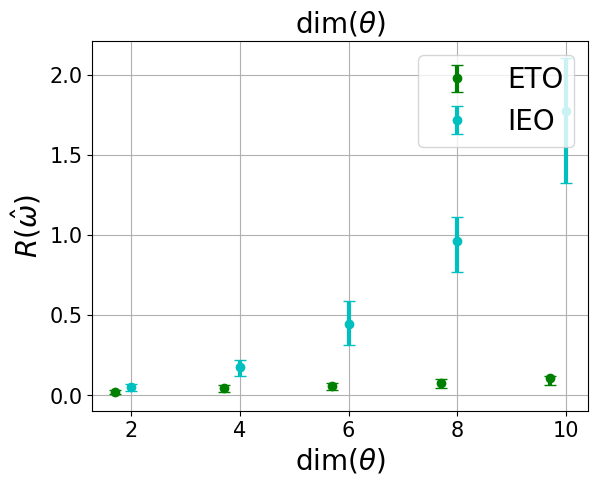}
\caption{Results for varying dimensions of $\theta$ and $\omege$. The regret plots show median, 25$^{th}$ quantile, and 75$^{th}$ quantile over 50 random seeds. Results are for the contextual case, and the decision dimension is fixed. Sample size is $n=100$.}
\label{fig:appendix_dimension_omege_theta}
\end{figure}

In Section \ref{sec:dim} we presented experimental results on the relative dimensions of decisions and model parameters, where we fix the latter to be 1 while varying the decision dimension. That was for the unconstrained case, where our theory imposes that the model parameter dimension is no greater than the decision dimension. On the other hand, in the contextual case, our theory allows the model parameter dimension to be greater than the decision dimension, as the decision now becomes a map from the feature that has an enlarged flexibility. In Figure~\ref{fig:appendix_dimension_omege_theta}, we present results for the well-specified, contextual case studied in Section \ref{sec:all cases}, where we now increase the model parameter dimension while keeping the decision dimension fixed at 1. We see that, coinciding with our Theorem \ref{SD3}, ETO outperforms IEO across the considered configurations. 

\subsection{Details for the Portfolio Optimization Problems}
\subsubsection{Algorithms for the portfolio optimization problem.}
Recall we consider the following objective:
\begin{align*}
c(\omege, \ran) := \alpha \left(\omege^\top( \ran, -1)\right)^2 - \omege^\top (\ran, 0).
\end{align*}

Recall the expectation of the first term is $\E_P\left[\alpha \left(\omege^\top(\ran, -1)\right)^2\right] = 
\alpha\text{Var}\left(\omege^\top (\ran,0)\right)$, when $\omegeun^{(p)}$ is chosen optimally as $\mathbb{E}\left[\sum_{j=1}^{p-1}\omegeun^{(j)}\ranun^{(j)}\right]$ \citep{kallus2022stochastic,grigas2021integrated}. We assume each asset $j$ is independent and has distribution $\mathcal{N}(\theteun^{(j)},\sigma_j^2)$ with known $\sigma_j$ and unknown $\theteun^{(j)}$. We have
\begin{align*}
\mathbb{E}\left[c(\omege,\ran)\right] = \alpha \sum_{j=1}^{p-1} (\omegeun^{(j)}\sigma_j)^2 - \sum_{j=1}^{p-1} \omegeun^{(j)} \theteun^{(j)}
\end{align*}
The best decision to make $\omege_\thete$ is the solution to
\begin{align*}
\min_{\omege} & \ 
\alpha \sum_{j=1}^{p-1} (\omegeun^{(j)} \sigma_j)^2 - \sum_{j=1}^{p-1} \omegeun^{(j)} \theteun^{(j)}\\
\text{s.t. }& (\omegeun^{(1)},...,\omegeun^{(p-1)})\in \Delta^{p-1}\\
& \omegeun^{(p)} = \sum_{j=1}^{p-1}\omegeun^{(j)}\theteun^{(j)}.
\end{align*}


{\bf ETO.} For ETO, we first compute the maximum likelihood estimator (MLE) $\hat\thete^{ETO}=\frac{1}{n}\sum_{i=1}^n \ran_{i}$ for the unknown mean in the Gaussian model. We then compute the decision $\hat\omege^{ETO} = \omege_{\hat\thete^{ETO}}$.

{\bf IEO.} We solve the following optimization problem.
\begin{align*}
\min_{\bm\thete} & 
\sum_{i=1}^n \alpha\left(\omege_\thete^\top (\ran_i, -1)\right)^2 -  \sum_{i=1}^n \omege_\thete^\top (\ran_i,0)
\end{align*}
We use grid search to find an approximate optimal solution $\hat\thete^{IEO}$. We then compute the decision $\hat\omege^{IEO} = \omege_{\hat\thete^{IEO}}$.



{\bf SAA.} We solve the following constrained optimization problem
\begin{align*}
\min_{\omege} & \ \sum_{i=1}^n \alpha\left(\omegeun^\top(\ran_i-1)\right)^2 -  \sum_{i=1}^n \omegeun^\top(\ran_i,0)\\
\text{s.t. }& (\omegeun^{(1)},...,\omegeun^{(p-1)})\in \Delta^{p-1}\\
& \omega^{(p)} \geq 0
\end{align*}

%% file: bib.bib
@book{van2000asymptotic,
  title={Asymptotic statistics},
  author={Van der Vaart, Aad W},
  volume={3},
  year={2000},
  publisher={Cambridge university press}
}

@article{lam2021impossibility,
  title={On the impossibility of statistically improving empirical optimization: A second-order stochastic dominance perspective},
  author={Lam, Henry},
  journal={arXiv preprint arXiv:2105.13419},
  year={2021}
}

@book{kadison1986fundamentals,
  title={Fundamentals of the theory of operator algebras. Volume II: Advanced theory},
  author={Kadison, Richard V and Ringrose, John R},
  year={1986},
  publisher={Academic press New York}
}

@article{zhang2009binary,
  title={A binary solution method for the multi-product newsboy problem with budget constraint},
  author={Zhang, Bin and Xu, Xiaoyan and Hua, Zhongsheng},
  journal={International Journal of Production Economics},
  volume={117},
  number={1},
  pages={136--141},
  year={2009},
  publisher={Elsevier}
}

@article{ban2019big,
  title={The big data newsvendor: Practical insights from machine learning},
  author={Ban, Gah-Yi and Rudin, Cynthia},
  journal={Operations Research},
  volume={67},
  number={1},
  pages={90--108},
  year={2019},
  publisher={INFORMS}
}

@incollection{qi2022integrating,
  title={Integrating Prediction/Estimation and Optimization with Applications in Operations Management},
  author={Qi, Meng and Shen, Zuo-Jun},
  booktitle={Tutorials in Operations Research: Emerging and Impactful Topics in Operations},
  pages={36--58},
  year={2022},
  publisher={INFORMS}
}

@article{elmachtoub2022smart,
  title={Smart “predict, then optimize”},
  author={Elmachtoub, Adam N and Grigas, Paul},
  journal={Management Science},
  volume={68},
  number={1},
  pages={9--26},
  year={2022},
  publisher={INFORMS}
}

@article{bertsimas2020predictive,
  title={From predictive to prescriptive analytics},
  author={Bertsimas, Dimitris and Kallus, Nathan},
  journal={Management Science},
  volume={66},
  number={3},
  pages={1025--1044},
  year={2020},
  publisher={INFORMS}
}

@article{kallus2022stochastic,
  title={Stochastic optimization forests},
  author={Kallus, Nathan and Mao, Xiaojie},
  journal={Management Science},
  year={2022},
  publisher={INFORMS}
}

@article{grigas2021integrated,
  title={Integrated conditional estimation-optimization},
  author={Grigas, Paul and Qi, Meng and Shen, Zuo-Jun},
  journal={arXiv preprint arXiv:2110.12351},
  year={2021}
}

@article{hu2022fast,
  title={Fast rates for contextual linear optimization},
  author={Hu, Yichun and Kallus, Nathan and Mao, Xiaojie},
  journal={Management Science},
  year={2022},
  publisher={INFORMS}
}

@article{duchi2021asymptotic,
  title={Asymptotic optimality in stochastic optimization},
  author={Duchi, John C and Ruan, Feng},
  journal={The Annals of Statistics},
  volume={49},
  number={1},
  pages={21--48},
  year={2021},
  publisher={Institute of Mathematical Statistics}
}

@article{tripathi1999matrix,
  title={A matrix extension of the Cauchy-Schwarz inequality},
  author={Tripathi, Gautam},
  journal={Economics Letters},
  volume={63},
  number={1},
  pages={1--3},
  year={1999},
  publisher={Elsevier}
}

@article{shapiro1989asymptotic,
  title={Asymptotic properties of statistical estimators in stochastic programming},
  author={Shapiro, Alexander},
  journal={The Annals of Statistics},
  volume={17},
  number={2},
  pages={841--858},
  year={1989},
  publisher={Institute of Mathematical Statistics}
}

@article{chen2022statistical,
  title={A statistical learning approach to personalization in revenue management},
  author={Chen, Xi and Owen, Zachary and Pixton, Clark and Simchi-Levi, David},
  journal={Management Science},
  volume={68},
  number={3},
  pages={1923--1937},
  year={2022},
  publisher={INFORMS}
}

@article{donti2017task,
  title={Task-based end-to-end model learning in stochastic optimization},
  author={Donti, Priya and Amos, Brandon and Kolter, J Zico},
  journal={Advances in neural information processing systems},
  volume={30},
  year={2017}
}

@book{shapiro2021lectures,
  title={Lectures on stochastic programming: modeling and theory},
  author={Shapiro, Alexander and Dentcheva, Darinka and Ruszczynski, Andrzej},
  year={2021},
  publisher={SIAM}
}

@article{wright1993identifiable,
  title={Identifiable surfaces in constrained optimization},
  author={Wright, Stephen J},
  journal={SIAM Journal on Control and Optimization},
  volume={31},
  number={4},
  pages={1063--1079},
  year={1993},
  publisher={SIAM}
}

@article{quirk1962admissibility,
  title={Admissibility and measurable utility functions},
  author={Quirk, James P and Saposnik, Rubin},
  journal={The Review of Economic Studies},
  volume={29},
  number={2},
  pages={140--146},
  year={1962},
  publisher={JSTOR}
}

@book{mas1995microeconomic,
  title={Microeconomic theory},
  author={Mas-Colell, Andreu and Whinston, Michael Dennis and Green, Jerry R and others},
  volume={1},
  year={1995},
  publisher={Oxford university press New York}
}

@book{hastie2009elements,
  title={The elements of statistical learning: data mining, inference, and prediction},
  author={Hastie, Trevor and Tibshirani, Robert and Friedman, Jerome H and Friedman, Jerome H},
  volume={2},
  year={2009},
  publisher={Springer}
}

@article{namkoong2017variance,
  title={Variance-based regularization with convex objectives},
  author={Namkoong, Hongseok and Duchi, John C},
  journal={Advances in neural information processing systems},
  volume={30},
  year={2017}
}

@article{goh2010distributionally,
  title={Distributionally robust optimization and its tractable approximations},
  author={Goh, Joel and Sim, Melvyn},
  journal={Operations research},
  volume={58},
  number={4-part-1},
  pages={902--917},
  year={2010},
  publisher={INFORMS}
}

@article{delage2010distributionally,
  title={Distributionally robust optimization under moment uncertainty with application to data-driven problems},
  author={Delage, Erick and Ye, Yinyu},
  journal={Operations research},
  volume={58},
  number={3},
  pages={595--612},
  year={2010},
  publisher={INFORMS}
}

@article{lam2016robust,
  title={Robust sensitivity analysis for stochastic systems},
  author={Lam, Henry},
  journal={Mathematics of Operations Research},
  volume={41},
  number={4},
  pages={1248--1275},
  year={2016},
  publisher={INFORMS}
}

@article{ben2013robust,
  title={Robust solutions of optimization problems affected by uncertain probabilities},
  author={Ben-Tal, Aharon and Den Hertog, Dick and De Waegenaere, Anja and Melenberg, Bertrand and Rennen, Gijs},
  journal={Management Science},
  volume={59},
  number={2},
  pages={341--357},
  year={2013},
  publisher={INFORMS}
}

@article{gupta2022data,
  title={Data pooling in stochastic optimization},
  author={Gupta, Vishal and Kallus, Nathan},
  journal={Management Science},
  volume={68},
  number={3},
  pages={1595--1615},
  year={2022},
  publisher={INFORMS}
}

@article{mohajerin2018data,
  title={Data-driven distributionally robust optimization using the Wasserstein metric: Performance guarantees and tractable reformulations},
  author={Mohajerin Esfahani, Peyman and Kuhn, Daniel},
  journal={Mathematical Programming},
  volume={171},
  number={1},
  pages={115--166},
  year={2018},
  publisher={Springer}
}

@article{turken2012multi,
  title={The multi-product newsvendor problem: Review, extensions, and directions for future research},
  author={Turken, Nazli and Tan, Yinliang and Vakharia, Asoo J and Wang, Lan and Wang, Ruoxuan and Yenipazarli, Arda},
  journal={Handbook of newsvendor problems},
  pages={3--39},
  year={2012},
  publisher={Springer}
}

@book{bickel2015mathematical,
  title={Mathematical statistics: basic ideas and selected topics, volumes I-II package},
  author={Bickel, Peter J and Doksum, Kjell A},
  year={2015},
  publisher={Chapman and Hall/CRC}
}

@book{cramer1946mathematical,
  title={Mathematical methods of statistics},
  author={Cram{\'e}r, Harald},
  year={1946},
  publisher={Princeton university press}
}

@article{rao1945information,
  title={Information and the accuracy attainable in the estimation of statistical parameters},
  author={Rao, C Radhakrishna},
  journal={Reson. J. Sci. Educ},
  volume={20},
  pages={78--90},
  year={1945}
}

@inproceedings{wilder2019melding,
  title={Melding the data-decisions pipeline: Decision-focused learning for combinatorial optimization},
  author={Wilder, Bryan and Dilkina, Bistra and Tambe, Milind},
  booktitle={Proceedings of the AAAI Conference on Artificial Intelligence},
  volume={33},
  pages={1658--1665},
  year={2019}
}

@inproceedings{elmachtoub2020decision,
  title={Decision trees for decision-making under the predict-then-optimize framework},
  author={Elmachtoub, Adam N and Liang, Jason Cheuk Nam and McNellis, Ryan},
  booktitle={International Conference on Machine Learning},
  pages={2858--2867},
  year={2020},
  organization={PMLR}
}

@inproceedings{mandi2020smart,
  title={Smart Predict--and--Optimize for Hard Combinatorial Optimization Problems},
  author={Mandi, Jayanta and Demirovi{\'c}, Emir and Stuckey, Peter J and Guns, Tias},
  booktitle={Thirty-Fourth AAAI Conference on Artificial Intelligence},
  pages={1603--1610},
  year={2020},
  organization={AAAI Press}
}

@article{wilder2019end,
  title={End to end learning and optimization on graphs},
  author={Wilder, Bryan and Ewing, Eric and Dilkina, Bistra and Tambe, Milind},
  journal={Advances in Neural Information Processing Systems},
  volume={32},
  year={2019}
}

@article{yan2020semi,
  title={A semi-“smart predict then optimize”(semi-SPO) method for efficient ship inspection},
  author={Yan, Ran and Wang, Shuaian and Fagerholt, Kjetil},
  journal={Transportation Research Part B: Methodological},
  volume={142},
  pages={100--125},
  year={2020},
  publisher={Elsevier}
}

@article{qi2022practical,
  title={A practical end-to-end inventory management model with deep learning},
  author={Qi, Meng and Shi, Yuanyuan and Qi, Yongzhi and Ma, Chenxin and Yuan, Rong and Wu, Di and Shen, Zuo-Jun},
  journal={Management Science},
  year={2022},
  publisher={INFORMS}
}

@article{el2019generalization,
  title={Generalization bounds in the predict-then-optimize framework},
  author={El Balghiti, Othman and Elmachtoub, Adam N and Grigas, Paul and Tewari, Ambuj},
  journal={Advances in neural information processing systems},
  volume={32},
  year={2019}
}

@article{liu2021risk,
  title={Risk bounds and calibration for a smart predict-then-optimize method},
  author={Liu, Heyuan and Grigas, Paul},
  journal={Advances in Neural Information Processing Systems},
  volume={34},
  pages={22083--22094},
  year={2021}
}

@inproceedings{poganvcic2020differentiation,
  title={Differentiation of blackbox combinatorial solvers},
  author={Pogan{\v{c}}i{\'c}, Marin Vlastelica and Paulus, Anselm and Musil, Vit and Martius, Georg and Rolinek, Michal},
  booktitle={International Conference on Learning Representations},
  year={2020}
}

@article{gupta2022debiasing,
  title={Debiasing in-sample policy performance for small-data, large-scale optimization},
  author={Gupta, Vishal and Huang, Michael and Rusmevichientong, Paat},
  journal={Operations Research},
  year={2022},
  publisher={INFORMS}
}

@article{besbes2021big,
  title={How big should your data really be? data-driven newsvendor: Learning one sample at a time},
  author={Besbes, Omar and Mouchtaki, Omar},
  journal={arXiv preprint arXiv:2107.02742},
  year={2021}
}

@book{bazaraa2013nonlinear,
  title={Nonlinear programming: theory and algorithms},
  author={Bazaraa, Mokhtar S and Sherali, Hanif D and Shetty, Chitharanjan M},
  year={2013},
  publisher={John Wiley \& Sons}
}

@article{ho2022risk,
  title={Risk guarantees for end-to-end prediction and optimization processes},
  author={Ho-Nguyen, Nam and K{\i}l{\i}n{\c{c}}-Karzan, Fatma},
  journal={Management Science},
  volume={68},
  number={12},
  pages={8680--8698},
  year={2022},
  publisher={INFORMS}
}

@article{estes2021slow,
  title={Slow rates of convergence in optimization with side information},
  author={Estes, Alexander},
  journal={Available at SSRN 3803427},
  year={2021}
}

@article{wiesemann2014distributionally,
  title={Distributionally robust convex optimization},
  author={Wiesemann, Wolfram and Kuhn, Daniel and Sim, Melvyn},
  journal={Operations Research},
  volume={62},
  number={6},
  pages={1358--1376},
  year={2014},
  publisher={INFORMS}
}

@article{blanchet2019robust,
  title={Robust Wasserstein profile inference and applications to machine learning},
  author={Blanchet, Jose and Kang, Yang and Murthy, Karthyek},
  journal={Journal of Applied Probability},
  volume={56},
  number={3},
  pages={830--857},
  year={2019},
  publisher={Cambridge University Press}
}

@article{gotoh2018robust,
  title={Robust empirical optimization is almost the same as mean--variance optimization},
  author={Gotoh, Jun-ya and Kim, Michael Jong and Lim, Andrew EB},
  journal={Operations research letters},
  volume={46},
  number={4},
  pages={448--452},
  year={2018},
  publisher={Elsevier}
}

@article{gupta2021small,
  title={Small-data, large-scale linear optimization with uncertain objectives},
  author={Gupta, Vishal and Rusmevichientong, Paat},
  journal={Management Science},
  volume={67},
  number={1},
  pages={220--241},
  year={2021},
  publisher={INFORMS}
}

@book{shaked2007stochastic,
  title={Stochastic orders},
  author={Shaked, Moshe and Shanthikumar, J George},
  year={2007},
  publisher={Springer}
}

@article{johnson2013accelerating,
  title={Accelerating stochastic gradient descent using predictive variance reduction},
  author={Johnson, Rie and Zhang, Tong},
  journal={Advances in neural information processing systems},
  volume={26},
  year={2013}
}

@article{defazio2014saga,
  title={SAGA: A fast incremental gradient method with support for non-strongly convex composite objectives},
  author={Defazio, Aaron and Bach, Francis and Lacoste-Julien, Simon},
  journal={Advances in neural information processing systems},
  volume={27},
  year={2014}
}

@article{butler2023integrating,
  title={Integrating prediction in mean-variance portfolio optimization},
  author={Butler, Andrew and Kwon, Roy H},
  journal={Quantitative Finance},
  pages={1--24},
  year={2023},
  publisher={Taylor \& Francis}
}

@article{chung2022decision,
  title={Decision-Aware Learning for Optimizing Health Supply Chains},
  author={Chung, Tsai-Hsuan and Rostami, Vahid and Bastani, Hamsa and Bastani, Osbert},
  journal={arXiv preprint arXiv:2211.08507},
  year={2022}
}

@inproceedings{kotary2022end,
  title={End-to-end learning for fair ranking systems},
  author={Kotary, James and Fioretto, Ferdinando and Van Hentenryck, Pascal and Zhu, Ziwei},
  booktitle={Proceedings of the ACM Web Conference 2022},
  pages={3520--3530},
  year={2022}
}

@article{stanimirovic2017minimization,
  title={Minimization of quadratic forms and generalized inverses},
  author={Stanimirovic, Predrag and Pappas, Dimitrios and Katsikis, Vasilios N},
  journal={Advances in Linear Algebra Research},
  pages={1},
  year={2017}
}

@article{tang2022pyepo,
  title={PyEPO: A PyTorch-based End-to-End Predict-then-Optimize Library for Linear and Integer Programming},
  author={Tang, Bo and Khalil, Elias B},
  journal={arXiv preprint arXiv:2206.14234},
  year={2022}
}

@article{kannan2022data,
  title={Data-driven sample average approximation with covariate information},
  author={Kannan, Rohit and Bayraksan, G{\"u}zin and Luedtke, James R},
  journal={arXiv preprint arXiv:2207.13554},
  year={2022}
}

@article{srivastava2021data,
  title={On data-driven prescriptive analytics with side information: A regularized nadaraya-watson approach},
  author={Srivastava, Prateek R and Wang, Yijie and Hanasusanto, Grani A and Ho, Chin Pang},
  journal={arXiv preprint arXiv:2110.04855},
  year={2021}
}

@article{kannan2020residuals,
  title={Residuals-based distributionally robust optimization with covariate information},
  author={Kannan, Rohit and Bayraksan, G{\"u}zin and Luedtke, James R},
  journal={arXiv preprint arXiv:2012.01088},
  year={2020}
}

@article{sadana2023survey,
  title={A Survey of Contextual Optimization Methods for Decision Making under Uncertainty},
  author={Sadana, Utsav and Chenreddy, Abhilash and Delage, Erick and Forel, Alexandre and Frejinger, Emma and Vidal, Thibaut},
  journal={arXiv preprint arXiv:2306.10374},
  year={2023}
}

@article{l1990unified,
  title={A unified view of the IPA, SF, and LR gradient estimation techniques},
  author={L'Ecuyer, Pierre},
  journal={Management Science},
  volume={36},
  number={11},
  pages={1364--1383},
  year={1990},
  publisher={INFORMS}
}

@book{asmussen2007stochastic,
  title={Stochastic simulation: algorithms and analysis},
  author={Asmussen, S{\o}ren and Glynn, Peter W},
  volume={57},
  year={2007},
  publisher={Springer}
}

@book{glasserman2004monte,
  title={Monte Carlo methods in financial engineering},
  author={Glasserman, Paul},
  volume={53},
  year={2004},
  publisher={Springer}
}

@article{lavergne2008cauchy,
  title={A Cauchy-Schwarz inequality for expectation of matrices},
  author={Lavergne, Pascal and others},
  journal={Simon Fraser University, Tech. Rep},
  year={2008}
}

@article{kannan2021heteroscedasticity,
  title={Heteroscedasticity-aware residuals-based contextual stochastic optimization},
  author={Kannan, Rohit and Bayraksan, G{\"u}zin and Luedtke, James},
  journal={arXiv preprint arXiv:2101.03139},
  year={2021}
}

@article{bertsimas2022data,
  title={Data-driven optimization: A reproducing kernel Hilbert space approach},
  author={Bertsimas, Dimitris and Koduri, Nihal},
  journal={Operations Research},
  volume={70},
  number={1},
  pages={454--471},
  year={2022},
  publisher={INFORMS}
}

@article{bertsimas2019predictions,
  title={From predictions to prescriptions in multistage optimization problems},
  author={Bertsimas, Dimitris and McCord, Christopher},
  journal={arXiv preprint arXiv:1904.11637},
  year={2019}
}

@book{nocedal1999numerical,
  title={Numerical optimization},
  author={Nocedal, Jorge and Wright, Stephen J},
  year={1999},
  publisher={Springer}
}

@article{feng2023framework,
  title={The framework of parametric and non-parametric operational data analytics ({ODA})},
  author={Feng, Qi and Shanthikumar, J George},
  journal={Available at SSRN 4400555},
  year={2023}
}

@article{esteban2022distributionally,
  title={Distributionally robust stochastic programs with side information based on trimmings},
  author={Esteban-P{\'e}rez, Adri{\'a}n and Morales, Juan M},
  journal={Mathematical Programming},
  volume={195},
  number={1-2},
  pages={1069--1105},
  year={2022},
  publisher={Springer}
}

@article{munoz2022bilevel,
  title={A bilevel framework for decision-making under uncertainty with contextual information},
  author={Mu{\~n}oz, Miguel Angel and Pineda, Salvador and Morales, Juan Miguel},
  journal={Omega},
  volume={108},
  pages={102575},
  year={2022},
  publisher={Elsevier}
}

@article{liyanage2005practical,
  title={A practical inventory control policy using operational statistics},
  author={Liyanage, Liwan H and Shanthikumar, J George},
  journal={Operations Research Letters},
  volume={33},
  number={4},
  pages={341--348},
  year={2005},
  publisher={Elsevier}
}

@incollection{lim2006model,
  title={Model uncertainty, robust optimization, and learning},
  author={Lim, Andrew EB and Shanthikumar, J George and Shen, ZJ Max},
  booktitle={Models, Methods, and Applications for Innovative Decision Making},
  pages={66--94},
  year={2006},
  publisher={INFORMS}
}

@article{gao2022wasserstein,
  title={Wasserstein distributionally robust optimization and variation regularization},
  author={Gao, Rui and Chen, Xi and Kleywegt, Anton J},
  journal={Operations Research},
  year={2022},
  publisher={INFORMS}
}

@article{bertsimas2018robust,
  title={Robust sample average approximation},
  author={Bertsimas, Dimitris and Gupta, Vishal and Kallus, Nathan},
  journal={Mathematical Programming},
  volume={171},
  pages={217--282},
  year={2018},
  publisher={Springer}
}

@article{lam2018sensitivity,
  title={Sensitivity to serial dependency of input processes: A robust approach},
  author={Lam, Henry},
  journal={Management Science},
  volume={64},
  number={3},
  pages={1311--1327},
  year={2018},
  publisher={INFORMS}
}

@article{gupta2019near,
  title={Near-optimal Bayesian ambiguity sets for distributionally robust optimization},
  author={Gupta, Vishal},
  journal={Management Science},
  volume={65},
  number={9},
  pages={4242--4260},
  year={2019},
  publisher={INFORMS}
}

@inproceedings{zhang2017assessing,
  title={Assessing the performance of deep learning algorithms for newsvendor problem},
  author={Zhang, Yanfei and Gao, Junbin},
  booktitle={Neural Information Processing: 24th International Conference, ICONIP 2017, Guangzhou, China, November 14-18, 2017, Proceedings, Part I 24},
  pages={912--921},
  year={2017},
  organization={Springer}
}

@article{oroojlooyjadid2020applying,
  title={Applying deep learning to the newsvendor problem},
  author={Oroojlooyjadid, Afshin and Snyder, Lawrence V and Tak{\'a}{\v{c}}, Martin},
  journal={IISE Transactions},
  volume={52},
  number={4},
  pages={444--463},
  year={2020},
  publisher={Taylor \& Francis}
}

@article{elmachtoub2025dissecting,
  title={Dissecting the Impact of Model Misspecification in Data-Driven Optimization},
  author={Elmachtoub, Adam N and Lam, Henry and Lan, Haixiang and Zhang, Haofeng},
  journal={arXiv preprint arXiv:2503.00626},
  year={2025}
}

@article{iyengar2024cross,
  title={Is Cross-validation the Gold Standard to Estimate Out-of-sample Model Performance?},
  author={Iyengar, Garud and Lam, Henry and Wang, Tianyu},
  journal={Advances in Neural Information Processing Systems},
  volume={37},
  pages={94736--94775},
  year={2024}
}

@article{mandi2024decision,
  title={Decision-focused learning: Foundations, state of the art, benchmark and future opportunities},
  author={Mandi, Jayanta and Kotary, James and Berden, Senne and Mulamba, Maxime and Bucarey, Victor and Guns, Tias and Fioretto, Ferdinando},
  journal={Journal of Artificial Intelligence Research},
  volume={80},
  pages={1623--1701},
  year={2024}
}
